%% file: policy_grad_optimality.tex
\documentclass[10pt]{article}

\input{macros}
	
\begin{document}
	\title{\vspace{-2cm}{\Large{\bfseries{Global Optimality Guarantees For Policy Gradient Methods}} }}
	\author{Jalaj Bhandari}
	\author{Daniel Russo}
	\affil{Columbia University}
	\date{}
	\renewcommand\Authands{ and }
	\maketitle
	\input{abstract}
	\input{intro}

\input{literature}

\input{formulation_abstract}

\input{background}

\input{lq_body}

	\input{stationary_points}

	\input{examples}

	\input{initial_distribution}

\input{approx_closure}

	\input{conclusion}

	{\footnotesize 
		\setlength{\bibsep}{2pt plus 0.4ex}
		\bibliographystyle{plainnat}
		\bibliography{bibfile}
	}

	\appendix
	\input{notation}
	\input{new_app_ag_discussion}
\input{app_initial_distribution}
	\input{omitted_proofs}

	\input{app_example_details}
\end{document}

%% file: macros.tex
\PassOptionsToPackage{compress, sort}{natbib}
\usepackage[margin = 1in]{geometry}
\usepackage{tikz}
\usetikzlibrary{arrows,automata}
\usetikzlibrary{shapes.geometric,positioning}
\usepackage{hyperref}       
\hypersetup{colorlinks,linkcolor=blue,
            citecolor=blue,
            urlcolor=magenta,
            linktocpage,
            plainpages=false}
\usepackage{url}            
\usepackage{amsfonts}       
\usepackage{nicefrac}       
\usepackage{microtype} 
\usepackage{amsmath, amsthm, amssymb}
\usepackage{thm-restate}
\usepackage[ruled,algo2e]{algorithm2e}
\usepackage{algpseudocode}
\usepackage{bbm}
\usepackage{graphicx}
\usepackage{mathtools}
\usepackage{enumitem}
\usepackage{physics}
\usepackage{subcaption}
\usepackage{natbib}
\usepackage{authblk} 
\usepackage{times}
\usepackage{tabulary, booktabs}
\usepackage{thmtools}
\usepackage{thm-restate}
\usepackage{titling}

\makeatletter
\newenvironment{subtheorem}[1]{%
	\def\subtheoremcounter{#1}%
	\refstepcounter{#1}%
	\protected@edef\theparentnumber{\csname the#1\endcsname}%
	\setcounter{parentnumber}{\value{#1}}%
	\setcounter{#1}{0}%
	\expandafter\def\csname the#1\endcsname{\theparentnumber.\Alph{#1}}%
	\expandafter\def\csname theH#1\endcsname{thm.\theparentnumber.\Alph{#1}}%
	\unskip\ignorespaces
}{%
	\setcounter{\subtheoremcounter}{\value{parentnumber}}%
	\ignorespacesafterend
}
\makeatother
\newcounter{parentnumber}

\newtheorem{thm}{Theorem}

\newtheorem{definition}{Definition}
\newtheorem{assumption}{Assumption}
\newtheorem{lem}{Lemma}
\newtheorem{cor}{Corollary}
\newtheorem{rem}{Remark}
\newtheorem{cond}{Condition}
\newtheorem{example}{Example}

\newcommand{\thetabar}{{\overline{\theta}}}
\newcommand{\pibar}{{\bar{\pi}}}

\newcommand{\Mc}{\mathcal{M}}
\newcommand{\Bc}{\mathcal{B}}
\newcommand{\X}{\mathcal{X}}

\newcommand{\Sc}{\mathcal{S}}

\newcommand{\Kc}{\mathcal{K}}
\newcommand{\Xc}{\mathcal{X}}
\newcommand{\Yc}{\mathcal{Y}}
\newcommand{\Ac}{\mathcal{A}}
\newcommand{\Jc}{\mathcal{J}}
\newcommand{\R}{\mathbb{R}}
\newcommand{\E}{\mathbb{E}}

\newcommand{\Prob}{\mathbb{P}}

\DeclareMathOperator*{\argmax}{arg\,max}
\DeclareMathOperator*{\argmin}{arg\,min}

\def\jb#1{{\color{black}#1}}

\renewenvironment{abstract}
  {{\centering\large\bfseries Abstract\par}\vspace{0.7ex}%
    \bgroup
       \leftskip 20pt\rightskip 20pt\small\noindent\ignorespaces}%
  {\par\egroup\vskip 0.25ex}

\newenvironment{keywords}
{\bgroup\leftskip 20pt\rightskip 20pt \small\noindent{\bfseries
Keywords:} \ignorespaces}%
{\par\egroup\vskip 0.25ex}

\usepackage{setspace}

%% file: abstract.tex
\begin{abstract}
\noindent 

Policy gradients methods apply to complex, poorly understood, control problems by performing stochastic gradient descent over a parameterized class of polices. Unfortunately, even for simple control problems solvable by standard dynamic programming techniques, policy gradient algorithms face non-convex optimization problems and are widely understood to converge only to a stationary point. This work identifies structural properties -- shared by several classic control problems -- that ensure the policy gradient objective function has no suboptimal stationary points despite being non-convex. When these conditions  are strengthened, this objective satisfies a Polyak-lojasiewicz (gradient dominance) condition that yields convergence rates. We also provide bounds on the optimality gap of any stationary point when some of these conditions are relaxed.

\end{abstract}
\vspace{1pt}

%% file: intro.tex

\section{Introduction}
Many recent successes in reinforcement learning are driven by a class of algorithms called policy gradient methods. These methods search over a parameterized class of polices by performing stochastic gradient descent on a cost function capturing the cumulative expected cost incurred. Specifically, for discounted or episodic problems, they treat the scalar cost function $\ell(\pi) = \intop J_{\pi}(s) d\rho(s)$, which averages the total cost-to-go function $J_{\pi}$ over a random initial state distribution $\rho$.  Policy gradient methods aim to optimize over a smooth, and often stochastic, class of parameterized policies $\{\pi_{\theta}\}_{\theta \in \Theta}$ by performing stochastic gradient descent on $\ell(\cdot)$, as in the iteration
\[
\theta_{k+1} = \theta_k - \alpha_k \left(\grad_{\theta}  {\ell(\pi_{\theta_k})} + {\rm noise} \right).
\]
Stochastic gradients can be generated by monte-carlo simulation, even in complex environments and with policies represented by deep neural networks \citep{schulman2015trust, schulman2015high, schulman2017proximal}. This approach is especially appealing when one has an inductive bias about the form of an effective policy. For example, \cite{glasserman1995sensitivity} use gradient descent to optimize over simple structured policies in a realistic simulator of a multi-echelon inventory control problem. Direct policy search has been used in control problems in robotics \citep{peters2006policy}, manufacturing \citep{caramanis1992perturbation}, arcade games \citep{schulman2015trust}, revenue management \citep[Section~3.5.1]{talluri2006theory}, ambulance redeployment \citep{maxwell2013tuning},  scheduling in queues \citep{l1994stochastic1,l1994stochastic2}, and many other areas. 



Unfortunately, while policy gradient methods can be applied to a very broad class of problems, it is not clear whether they adequately address even simple control problems solvable by classical methods. A key challenge is that the total cost $\ell(\cdot)$ is a non-convex function of the chosen policy. Typical of results concerning black-box optimization of non-convex functions, policy gradient methods are widely understood to converge asymptotically to a stationary point or a local minimum. Important theory guarantees this under technical conditions \citep{marbach2001simulation,sutton2000policy, baxter2001infinite} and it is widely repeated in textbooks and surveys \citep{peters2006policy, grondman2012survey, sutton2018reinforcement}. But the literature seems to provide almost no guarantees into the \emph{quality} of these stationary points. Worse yet, Example \ref{ex: counterexample} shows that policy gradient methods can get stuck in a bad local minimum in very simple examples even when the policy class contains the optimal policy. 

\begin{example}\label{ex: counterexample}
Consider the MDP depicted in Figure \ref{subfig:MDP}. There are two states, left $(s_L)$ and right $(s_R)$, and two possible actions, $L$ and $R$, which move the agent to the desired state in the next period. Staying in the state $L$ incurs a cost $g(s_L,L)=1$ per period, whereas staying in the right state is costless with $g(s_R,R)=0$. Moving between states incurs a per-period cost of $2$. When the discount factor exceeds $1/2$, it is easy to show that the optimal policy chooses action $R$ in either state. In that case, it is reasonable to search in a constrained policy class, $\{\pi_{\theta}: \theta \in [0,1]\}$ that plays action $R$ with probability $\theta\in [0,1]$ regardless of the current state. Setting $\theta=1$ yields an optimal policy. Unfortunately, as shown in Figure \ref{fig:PG_counterexample}, the total discounted cost incurred, $\ell(\pi_{\theta})$, is a nonconvex function of $\theta$. When initialized with small value of $\theta$, cost is locally increasing as a function of $\theta$, and so a gradient method updates the policy toward a bad local minimum at $\theta=0$. Once there, any local policy search approach gets stuck as there are no descent directions that reduce cost. It is worth noting here that in general, policy gradient methods face many additional challenges, for instance due to unsophisticated exploration or policy parameterization. This example instead highlights the risk of bad local minima due to the non-convexity of the infinite horizon cost function $\ell(\cdot)$.


\begin{figure}
	\centering
	\begin{subfigure}{.55\textwidth}
	\centering
	\begin{tikzpicture}[state/.style={circle, draw, minimum size=1.2cm}, node distance=1.9cm]
	\node[state] (left) {\small $s_L$};
	\node[state, right=of left] (right) {\small $s_R$};
	\draw[
	>=latex,
	auto=right,                      
	loop above/.style={min distance=10mm,out=75,in=1055,loop,},
	every loop,
	]	
	(left)   edge[loop left] node[above left,pos=1.05] {\scriptsize $g(s_L,L)=1$}   (left)
	edge[bend right] node {\scriptsize $g(s_L,R)=2$}   (right)
	(right)  edge[loop right] node[above right,pos=-0.04] {\scriptsize $g(s_R,R)=0$}   (right)
	edge[bend right] node {\scriptsize $g(s_R,L)=2$}   (left);
	\end{tikzpicture}
	\caption{Two state, two action MDP}
	\label{subfig:MDP}
	\end{subfigure}%
	\begin{subfigure}{.45\textwidth}
	\centering
	\includegraphics[width=0.8\textwidth]{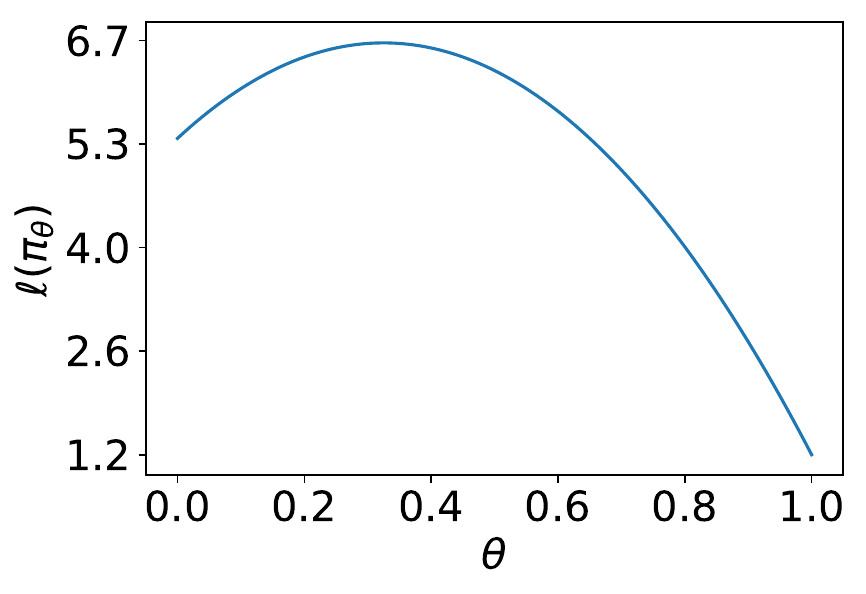}
	\caption{$\ell(\pi_{\theta})$ when $\gamma=0.8$ and $\rho = [0.6, 0.4]$}
	\label{subfig:ell}
	\end{subfigure}
	\caption{Presence of bad local minima with a constrained policy class.} 
	\label{fig:PG_counterexample}
\end{figure}
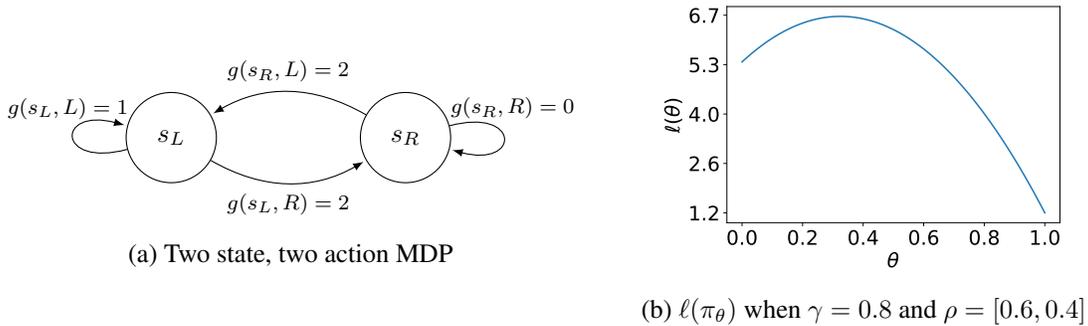
\end{example}

In marked contrast to the example above, important recent work of \citet{fazel2018global} showed that  for the deterministic linear quadratic control problem, policy gradient with the class of linear policies converges to the global optimum, despite non-convexity of the objective. Here, the authors provided an intricate analysis, leveraging a variety of closed form expressions available for linear-quadratic problems. Separately, in the operations literature,  \citet{kunnumkal2008using} propose a stochastic approximation method for setting base-stock levels in inventory control. In this example too, the objective is non-convex, but the authors  establish convergence to the globally optimal solution using an intricate analysis quite different than that of \citet{fazel2018global}. How do we reconcile these success stories with the simple counterexample given in Example \ref{ex: counterexample}?

\subsection{Our Contribution}
\label{subsec:contribution}
Policy gradient methods aim to directly minimize the \textit{multi-period} total discounted cost by applying first-order optimization methods. By contrast, classical dynamic programming methods, like value and policy iteration, indirectly minimize the total cost by solving a sequence of simpler \textit{single period} problems. We uncover an \emph{indirect} analysis of policy gradient methods, deducing global convergence properties from conditions on the single period problems solved by policy iteration. 

As a consequence of our general framework, we show that for several classic control problems, policy gradient methods performed with respect to natural structured policy classes face no suboptimal local minima. More precisely, despite its non-convexity, any stationary point\footnote{For unconstrained problems, stationary points of a function $f$ satisfy $\nabla f(x)=0$. More generally, for constrained optimization over some set $\Xc$, any stationary point $x$ satisfies the first order necessary conditions for optimality, $\nabla f(x)^{\top} (x'-x) \geq 0 \,\, \forall x' \in \Xc$.} of the policy gradient cost function is a global optimum. The examples we treat include:
\begin{itemize}
	\item Finite state and action MDPs with the class of all stochastic policies. 
	\item Linear quadratic (LQ) control problems with the class of linear policies. 
	\item An optimal stopping problem with the class of threshold policies. 
	\item A finite horizon inventory control problems with the class of non-stationary base-stock policies.  
\end{itemize}

These canonical control problems provide an important benchmark and sanity check. But why do policy gradient methods avoid suboptimal local minima in these cases as opposed to the simple case in Example \ref{ex: counterexample}? Interestingly, the examples above share some important structural properties. Consider the following properties of the LQ control example: 
\begin{enumerate}
	\item The policy class is \textit{closed} under policy improvement. That is, starting with a linear policy and performing a policy iteration step yields another linear policy.
	\item  The single period optimization problem defining a policy iteration update has no suboptimal stationary points. In particular, the objective is a simple convex quadratic  that can be easily optimized using first-order methods. 
\end{enumerate}
These same properties hold for finite MDPs as well as the optimal stopping problem. Our result in Theorem \ref{thm: no bad stationary points general} shows that these two properties, together with mild regularity conditions, imply that any stationary point of the policy gradient loss function is globally optimal.

We remark that the closure condition is much weaker than requiring the policy class to contain all possible policies. This is useful, for example, in problems where simple structured policy classes may be naturally aligned with the problem objective. However, the closure property is stronger than only requiring the policy class to contain (near) optimal policies.
The presence of bad local minima in Example \ref{ex: counterexample} illustrates why a stronger condition is necessary. In that case, the policy class contains the optimal policy, but it is not closed under policy improvement (see Section \ref{subsec:conditions}). 



\jb{We extend our result in Theorem \ref{thm: no bad stationary points general} in several ways.}
 Theorem \ref{thm: convergence_rates} studies a strengthening of the second condition above that leads to fast converge rates. In particular, when the single period objective defined by a (weighted) policy iteration problem satisfies a Polyak-lojasiewicz (PL) condition (also popularly known as ``gradient dominance''), this property is inherited by the policy gradient objective $\ell(\cdot)$. The PL conditions are relaxations of (strong) convexity that guarantee fast global convergence of first-order methods even for non-convex objectives \citep{polyak1963gradient, nesterov2006cubic}. 

Next, in Theorem \ref{thm: finite_horizon}, we show that for finite horizon problems with non-stationary policy classes, like the finite horizon inventory control problem, we only require that the policy class contains an optimal policy instead of the closure condition. In addition, we only require a weaker version of condition 2 above. See Section \ref{sec:stationary_points} for more details. Finally, Theorem \ref{thm: approximate closure}, studies a weakening of the closure condition more generally. It assumes that the policy class is \textit{approximately} closed under policy improvement -- meaning that the policy iteration update can be solved in the given policy class up to a small error. In this case, we show how bound the optimality gap of any stationary point. Many of our intermediate results may also be of interest, especially our approach to concentrability coefficients in Section \ref{sec:concentrability} and the corresponding bounds in Theorem \ref{thm: concentrability general}.

\subsection{Limited scope of this work}\label{subsec: scope}
This paper is focused on understanding the optimization landscape of the non-convex policy gradient objective $\ell(\cdot)$, which is a fundamental challenge for local policy search methods. Such an investigation is simultaneously relevant to many strategies for searching locally over the policy space, including policy gradient methods \citep{sutton2000policy}, natural gradient methods \citep{kakade2002natural}, finite difference schemes \citep{riedmiller2007evaluation}, evolutionary strategies \citep{salimans2017evolution} etc. 

However, we do sidestep many other issues that could lead to poor performance for specific local policy search methods. One such notable issue is that of \emph{exploration}. It is well known that the naive random exploration provided by stochastic policies may be insufficient to guarantee convergence to an optimal policy. See \cite{kakade2002approximately} or our discussion in Appendix \ref{sec: app_iniital_distribution} for examples. We imagine that we have access to a simulator where a restart distribution ensures that important regions of the state space are visited under all policies (see Assumption \ref{ass: exploratory rho}). 

Another issue involves the choice of \emph{policy parameterization}. In particular, for the popular softmax parameterization, the Jacobian matrix becomes ill conditioned near corners of the probability simplex, making a policy nearly insensitive to changes in the parameter space. Successful policy gradient methods must perform a local change of variables, like the popular natural policy gradient methods proposed by \cite{kakade2002natural}. See Section \ref{sec:examples} for a detailed discussion on softmax policies. In addition, we 
do not address the choice of a \emph{gradient estimator}, which is the subject of a large literature on stochastic simulation and reinforcement learning. See \cite{sutton2018reinforcement}, \cite{mohamed2020monte}, \cite{glasserman1991gradient} or \cite{fu2006gradient} for reviews from different perspectives \jb{as well as recent work by \citet{Masatoshi-OPPG,Masatoshi_doubly-robust} on policy gradient estimation using limited off-policy data}.

For the rest of this paper, we will consider an idealized policy gradient update with access to exact gradient evaluations. Generalizations to treat stochastic noise in gradient evaluations are possible using classical results from stochastic approximation literature, which show that under regularity conditions and appropriately decaying step-sizes most noisy iterative algorithms converge to the same limit as their deterministic counterparts \cite[see e.g.][]{borkar2009stochastic, bertsekas1996neuro}. We do not pursue such extensions for brevity.

%% file: literature.tex
\subsection{Related Literature}\label{sec: lit}

\paragraph{Realizability and closure}
\jb{Closure conditions play an important role in the study of methods which fit parametric approximations to value functions. Such conditions appear implicitly already in the classic textbook of \cite{bertsekas1996neuro} and refined notions were introduced in
 \cite{munos2003error, munos2005error,munos2007performance} and \cite{munos2008finite}. Closure conditions are sometimes called \emph{completeness} assumptions in the literature. They are stronger than \emph{realizability} conditions, which assume that the parametric class of value functions (nearly) contains the true value function. There is some evidence that conditions stronger than realizability are necessary for any statistically efficient RL algorithm to exist.  Recent papers of \cite{chen2019information, Du2020Is, Weisz21ALT} and \cite{wang2021what} construct examples where realizability holds but no algorithm which uses a polynomial number of samples from the environment can produce a near optimal policy. The insufficiency of realizability does not imply that closure conditions are necessary. For instance, in an off-policy estimation problem, \cite{uehara2021finite} identify a relaxed condition.}
 
 \jb{Our paper is focused on studying a seemingly distinct issue -- landscape of the policy gradient loss function -- so it is intruiging that many of the same issues arise. A realizability assumption holds in Example \ref{ex: counterexample}, but the optimization landspace is poor and policy gradient could get stuck in a bad local minimum. The optimization landscape improves when the policy class is closed under policy improvement.}

\paragraph{Prior work on analysis of policy gradient methods.}
Apart from the aforementioned works of \cite{kunnumkal2008using} and \cite{fazel2018global}, there has been limited prior work in theoretical guarantees for policy gradient methods, especially beyond tabular MDPs. See \cite{agarwal2020optimality} for a detailed literature review of results for the tabular setting. 

	One notable exception is the work by \cite{scherrer2014local} which provides guarantees on the quality of (approximate) stationary points obtained by local policy search algorithms with a \textit{convex policy class}, such as Conservative Policy Iteration \citep{kakade2002approximately}. \jb{While Conservative Policy Iteration does not explicitly place convexity conditions on the policy class, 
	 \cite{scherrer2014local}  argue that it can be viewed as a special case of local policy search over an expanded policy class which is the convex hull of the original policy class (i.e. the set of all mixture distributions over policies).} Relative to these works, our result in Theorem \ref{thm: no bad stationary points general} is more general as it applies to problem settings with deterministic policies, infinite action spaces, structured cost functions, and policy classes that are not convex, such as the class of threshold or base-stock policies.


 \paragraph{Concurrent work.}
 Concurrently with this work, \cite{agarwal2020optimality} provide a detailed study of policy gradient methods, primarily focusing on convergence rates in the tabular case for specific algorithms and with different policy parameterizations. They also extend their insights to the function approximation setting, focusing on natural gradient descent with a restricted class of log-linear policies in finite action spaces. \jb{We provide a detailed comparison with their concurrent work in Appendix \ref{sec: app_agrawal_discussion}. There we also establish a precise link between their conditions on the quality of parametric approximations to the value function and our conditions on closure of the policy class.} Along similar lines, to \cite{agarwal2020optimality}, \cite{shani2020adaptive} also provide convergence rates for trust-region based policy optimization methods for regularized tabular MDPs using ideas from analysis of the classic mirror descent method \citep{beck2003mirror}.
Although we do not analyze specific algorithms, intellectually we view regularized MDPs as satisfying a second-order gradient dominance condition, for which basic results in optimization theory imply fast convergence rates with first-order methods. See Section \ref{sec:examples} for details. 
 Finally, a recent paper by \cite{wang2019neural} studies global convergence properties of actor-critic methods in the compatible function approximation setting of \cite{sutton2000policy} with over-parameterized two layer neural networks, a regime in which the linearization error is bounded and neural networks essentially behave like kernel functions \citep{jacot2018neural}. Complimentary to these works, we focus primarily on novel insights into when, why, and how local policy search methods succeed in finding a near-optimal policy. Our results apply to classic control problems, including those with continuous state and action spaces, and structured classes of deterministic policies. \jb{A unique feature of the work is that it applies in a unified way not only to a problem like the linear MDPs (Example \ref{ex:linear_mdp}) which are popular among reinforcement learning theorists, but also to problems like optimal stopping, linear quadratic control, and inventory control, where it is especially natural to restrict policy search to a particular structured class of policies.} 
%

\paragraph{On non-convex optimization in machine learning.}
Beyond reinforcement learning, our work connects to an emerging body of literature on first-order methods for non-convex optimization problems, giving rates of convergence to stationary-points for first-order methods \citep{ghadimi2013stochastic,xiao2014proximal, defazio2014saga, reddi2016stochastic, ghadimi2016accelerated,reddi2016proximal, reddi2016stochasticfrank,  carmon2018accelerated, davis2019proximally, davis2020stochastic} and \jb{ensuring convergence to approximate local minima rather than saddle points}  \citep{lee2016gradient, agarwal2017finding, jin2017escape}. A complementary line of research studies the optimization landscape of specific problems to essentially ensure that any local minima is globally optimal \citep{ge2015escaping, 7755794, bhojanapalli2016global, ge2016matrix, kawaguchi2016deep}. Taken together, these results show how interesting non-convex optimization problems can be efficiently solved using first-order methods. Our work contributes to the second line of research, offering insight into landscape of the optimization objective $\ell(\cdot)$. \jb{It is worth noting that, unlike some of the aforementioned work, we do not focus on ensuring convergence to local minima rather than saddle points. Example \ref{ex: counterexample} shows that local minima could be far from optimal in simple examples. The conditions we identify ensure that there are no suboptimal stationary points -- including suboptimal saddle points.}

%% file: formulation_abstract.tex
\section{Problem formulation}
\label{section:problem_formulation}
We defer measurability assumptions required for a rigorous presentation until the end of this section, keeping most of the formulation less formal but more accessible. A Markov decision process (MDP) is a six-tuple $\left(\Sc, (\Ac_s)_{s\in \Sc}, g, P, \gamma, \rho \right)$, consisting of a state space $\Sc \subset \mathbb{R}^n$, action spaces $(\Ac_s)_{s\in \Sc}$, cost function $g$, transition kernel $P$, discount factor $\gamma \in (0,1)$ and initial distribution $\rho$. For each state $s\in \Sc$,  $\Ac_s\subset \mathbb{R}^k$ is the set of feasible actions. We take $\Ac=\cup_{s} \Ac_s$. The transition kernel $P$ specifies a probability distribution $P(\cdot | s, a)$ over $\Sc$ for any given state $s\in \Sc$ and action $a\in \Ac_s$. The cost function $g(\cdot)$ encodes the instantaneous expected cost  $g(s,a)$ incurred when selecting action $a$ in state $s$. We assume that per-period costs are uniformly bounded, meaning $\sup_{s\in \Sc, a\in \Ac_s} |g(s,a)|<\infty$.  

A stationary policy $\pi: \Sc \to \Ac$ is a function that prescribes a feasible action\footnote{ Rather than develop separate notation for randomized and deterministic policies, in specific settings like Example \ref{ex:tabularMDPs}, we accommodate randomized policies by letting $a\in \Ac_s$ denote a choice of some probability vector.} $\pi(s)\in \Ac_s$ for each state $s\in \Sc$. Let $\Pi$ denote the set of all (measurable) stationary polices and $\Mc \subset \Sc$ be any measurable set. For any $\pi \in \Pi$, let $g_{\pi}(s) = g(s, \pi(s))$ denote the per step cost function and $J_{\pi}(s) = \E^{\pi}_{s}\left[ \sum_{t=0}^{\infty} \gamma^t g_{\pi}(s_t) \right]$ to be the corresponding cost-to-go function. Here, the notation $\E^{\pi}_{s}[\cdot]$ indicates that expectation is taken over the Markovian sequence of states $(s_0, s_1, \cdots)$ with $s_0=s$ and the transition kernel $\Prob^{\pi}(s_{t+1} \in \Mc | s_t, \cdots ,s_{0} )  = P(\Mc | s_t, \pi(s_t) )$. More generally, we write $\E_{\nu}^{\pi}[\cdot]$ when the initial state is randomly drawn from some distribution $\nu$ over $\Sc$ and let $\Prob_{\nu}^{\pi}(\cdot)$ denote the corresponding probability measure. 

\paragraph{Loss function and policy gradient methods.}
Under the initial distribution $\rho$, a stationary policy $\pi \in \Pi$ has discounted average cost
\begin{equation}\label{eq:pg loss} 
\ell(\pi) =  (1-\gamma)\E^{\pi}_{\rho}\left[ \sum_{t=0}^{\infty} \gamma^t g_{\pi}(s_t) \right] = (1-\gamma)\intop J_{\pi}(s) \rho (ds), 
\end{equation}
and \textit{discounted state-occupancy measure} given by 
\begin{equation} \label{eq: discounted_state_occupancy_measure}
\eta_{\pi}(\Mc) =  (1-\gamma)\sum_{t=0}^{\infty} \gamma^t \Prob_{\rho}^{\pi}(s_t \in \Mc)   \qquad \Mc \subset \Sc. 
\end{equation}
Since $\Prob_{\rho}^{\pi}(s_t \in \cdot)$ is the distribution of the state at time $t$ under the policy $\pi$ and initial distribution $\rho$,
the discounted state occupancy measure gives the discounted fraction of time the system spends in a given part of the state space. It satisfies the balance equation $\eta_{ \pi}(\Mc) =  \intop \left[(1-\gamma) \rho(\Mc) + \gamma P(\Mc | s, \pi(s))\right]\eta_{ \pi}(ds)$, allowing it to also be interpreted as the steady state distribution of an equivalent average cost problem where in every period the state resets according to $\rho$ with probability $(1-\gamma)$ and otherwise evolves according to the transition kernel. See Lemma \ref{lem: balance equation} in  Appendix \ref{appendix B} for a proof. Note that $\ell(\pi)$ can be equivalently written as $\ell(\pi) = \intop  g_{\pi}(s) \, \eta_{\pi}(ds)$,  allowing it to be interpreted as an average per-period cost.  

Policy gradient methods directly apply first-order optimization methods to minimize $\ell(\cdot)$ over a chosen parameterized class of policies, $\Pi_{\Theta} = \{ \pi_\theta(\cdot): \theta \in \Theta \} \subset \Pi$. For this reason, we often refer to $\ell$ as the policy gradient objective function. We assume  $\Theta \subset \mathbb{R}^d$ is convex and denote $\Jc_{\Theta} = \{ J_{\pi_\theta} : \theta \in \Theta \}$ to be the set of cost-to-go functions corresponding to $\Pi_{\Theta}$. To indicate that we are referring to a policy in the restricted policy class, rather than an arbitrary stationary policy $\pi \in \Pi$, we typically either write $\pi_\theta$ or specify that $\pi \in \Pi_{\Theta}$. We overload notation, writing $\ell(\theta) = \ell(\pi_{\theta})$. We will later impose smoothness conditions to ensure differentiability of $\ell(\theta)$.

The optimal cost-to-go function is defined as $J^*(s) = \inf_{\pi \in \Pi} J_{\pi}(s)$. A stationary policy $\pi$ is said to be optimal if $J_{\pi}(s) = J^*(s)$ for every $s\in \Sc$. Under the technical conditions stated below in Assumption \ref{ass: measurable selection}, at least one optimal policy exists, which we denote by $\pi^*$ throughout the paper. The next lemma establishes that optimal policies are minimizers of the average cost function $\ell$ and that the reverse direction essentially holds when $\rho$ places positive weight on all parts of the state space. For example, if $\Sc$ is discrete and $\rho(s)>0$ for all $s\in \Sc$, then any minimizer of $\ell$ is an optimal policy. The proof can be found in Appendix \ref{appendix B}. 
\begin{restatable}[]{lem}{optimalPolicies}\label{lem:minimizers_of_ell}
	A policy satisfies  $\pi \in \argmin_{\pi' \in \Pi} \ell(\pi')$ if and only if $J_{\pi} = J^*$ $\rho$--almost surely, i.e. \\ $\rho\left(\{ s\in \Sc : J_{\pi}(s) =J^*(s) \} \right)=1$.
\end{restatable}

\paragraph{Exploratory initial distribution.} 
Policy gradient methods have poor convergence properties if applied without an exploratory initial distribution. See Appendix \ref{sec: app_iniital_distribution} for a full discussion. Inspired by \cite{kakade2002approximately}, we assume that the discounted state occupancy measure under an optimal policy is absolutely continuous with respect to the initial distribution, mathematically denoted as $\eta_{ \pi^*} \ll \rho$. Roughly, an assumption of this form ensures that the policy gradient loss function in \eqref{eq:pg loss} is sensitive to policy performance in all important parts of the state space. When $\Sc$ is discrete, it suffices to assume $\rho(s)>0$ for each $s\in \Sc$. When $\rho$ and $\eta_{ \pi}$ both possess probability density functions (PDFs) over $\Sc$, it suffices to assume that $\rho$ is supported over $\Sc$. In Section \ref{sec:concentrability}, we discuss more about the dependence of our results on specific choices of the initial distribution.
\begin{assumption}\label{ass: exploratory rho}  
	We assume that $\eta_{ \pi^*}$ is absolutely continuous with respect to $\rho$. That is, for any $\Mc \subset \Sc$ such that $\rho(\Mc)=0$, we have $\eta_{\pi^*}(\Mc)=0$.  
\end{assumption}
By \eqref{eq: discounted_state_occupancy_measure},  $\eta_{\pi} \succeq (1-\gamma)\rho$ for any stationary policy $\pi \in\Pi$ where $\succeq$ is used to denote an inequality that holds element-wise. Therefore, under Assumption \ref{ass: exploratory rho}, $\eta_{ \pi^*} \ll \eta_{ \pi}$ for any policy $\pi$. Many of our results actually rely on this consequence of Assumption \ref{ass: exploratory rho}, rather than Assumption \ref{ass: exploratory rho} itself.

\paragraph{Bellman operators.}
Let $\Jc$ denote the set of bounded (measurable) functions on the state space. 
Define the Bellman operator $T_{\pi}: \Jc \to \Jc$  and Bellman optimality operator $T: \Jc \to \Jc$ by
\begin{align}
\left(T_{\pi} J\right)(s) := g(s, \pi(s)) +   \intop J(s') P(ds' | s, \pi(s)), \label{eq: bellman update} \\ 
(TJ)(s) := \min_{a\in \Ac_s}  \left[ g(s, a) +   \intop J(s') P(ds' | s, a) \right] \label{eq: optimal bellman update}.
\end{align}
The Bellman optimality operator in \eqref{eq: optimal bellman update} can be equivalently defined as $TJ(s) = \min_{\pi \in \Pi} T_{\pi} J(s)$. Assumption \ref{ass: measurable selection} below ensures that the minimum in \eqref{eq: optimal bellman update} is attained and in particular there exists policy $\pi \in \Pi$ such that $T_{\pi} J = TJ$. It is well known that when the per-period costs are uniformly bounded (as we assumed), $T$ and $T_{\pi}$ are monotone and contraction operators with respect to the maximum norm. Their unique fixed points are $J^*$ and $J_{\pi}$, respectively. We repeatedly use the following element-wise inequalities, which   hold for any policy $\pi$ and $J \in \Jc$ and can be deduced from \eqref{eq: bellman update} and \eqref{eq: optimal bellman update} above:
\begin{align}\label{eq: bellman update reduces cost}  
	TJ \preceq T_{\pi}J    \quad \text{and} \quad TJ_{\pi} \preceq J_{\pi}. 
\end{align}

The state-action cost-to-go function (``Q-function'') corresponding to a policy $\pi\in \Pi$ is given by,
\begin{align}\label{eq: def q function}
Q_{\pi}(s,a)= g(s,a)+ \gamma \intop  J_{\pi}(s') P(ds'\mid s,a).
\end{align}
Define $Q^*(\cdot) = Q_{\pi^*}(\cdot)$. Notice that for any polices $\pi,\pi'\in \Pi$, we have the following relations,
\begin{align}\label{eq: q to bellman}
Q_{\pi}(s,\pi(s)) =  J_{\pi}(s),  \qquad    Q_{\pi}(s,\pi'(s)) =  (T_{\pi'} J_{\pi})(s),  \qquad  \min_{a\in \Ac_s} Q_{\pi}(s,a) = (TJ_{\pi})(s).
\end{align}

\paragraph{Notation.} For any $J \in \Jc$, we define the weighted $p$-norm as, $\| J\|_{p,\nu} = \left(\intop |J(s)|^p  \nu(ds)\right)^{1/p}$ for a given probability distribution $\nu$ over $\Sc$ and $p\geq 1$. Similarly, the maximum norm is given by $\| J\|_{\infty} =  \sup_{s \in \Sc} |J(s)|$. For a matrix $A \in \mathbb{R}^{n\times m}$, we write $\| A \|_p = \max_{x: \| x\|_p = 1 }  \| Ax\|_p$ for the operator norm and, in the case where $m=n$ and $A=A^\top$, let $\lambda_{\min}(A)$ and $\lambda_{\max}(A)$ denote the minimum and maximum eigenvalue. Let $\langle x, y\rangle = x^\top y=\sum_{i} x_i y_i$ denote the standard inner product. We write element-wise inequalities as $\preceq$ or $\succeq$, so $J\preceq J'$ if and only if $J(s)\leq J'(s)$ for each $s\in \Sc$.

\paragraph{Measurability assumptions.} 
Here we state assumptions which avoid pathological measurability issues that can arise in dynamic programming with general state and action spaces \citep{blackwell1965discounted,bertsekas1978stochastic}, i.e. when both state and actions spaces are uncountably infinite. Readers unfamiliar with measure theory can safely skip this while still understanding the paper's main insights. The key condition is the existence of a measurable selection rule, which is a measurable rule that associates each state with  an action attaining the minimum in a Bellman update. All of our examples satisfy appropriate smoothness and compactness conditions that ensure such a condition holds. We refer the readers to \cite[Section 3.3]{hernandez2012discrete} for a detailed account. A brief introduction is also given in \cite[Section 6.2.5]{puterman2014markov}. We use the term measurable to refer to a Borel measurable set or function. 
\begin{assumption}[Measurable selection] \label{ass: measurable selection}
	Assume the sets $\Sc$, $\Ac$ and  $\Kc := \{ (s,a): s\in \Sc, a\in \Ac_s  \}$ as well as the function $g: \Kc\to \mathbb{R}$ are measurable. The transition kernel $P$ is a stochastic kernel on $\Sc$ given $\Kc$, meaning that for each $(s,a)\in \Kc$, $P(\cdot | s,a)$ is a probability measure on $\Sc$ and for each measurable set $\Mc \subset \Sc$, $P(\Mc | \cdot )$ is a measurable function. Assume that for each bounded measurable function $J$ on $\Sc$, there is a measurable policy $\pi\in \Pi$ such that 
	\[ 
	g(s, \pi(s)) + \gamma \intop J(s') P(ds' | s, \pi(s) ) = \inf_{a\in \Ac_s} \left[ g(s, a) + \gamma \intop J(s') P(ds' | s, a ) \right]. 
	\]
\end{assumption}

%% file: background.tex
\section{Background on smooth nonconvex optimization}\label{sec:optimization_background}
\paragraph{Convergence to stationary points.} Given that the policy gradient objective is almost always non-convex, optimization algorithms generally will not converge to a global minimum. Instead, classical theory suggests that under appropriate smoothness conditions many algorithms will converge to stationary points of the objective, i.e. those satisfying the first-order necessary conditions for optimality as in Definition \ref{def: stationary}. This motivates our approach of studying the landscape of the policy gradient objective -- and in particular the quality of its  (approximate) stationary points -- rather than studying convergence properties of specific algorithms.  

\begin{definition}\label{def: stationary}
	Consider the optimization problem $\min_{x \in \X} f(x)$ where $\X \subset \mathbb{R}^d$ is a closed convex set and $f$ is continuously differentiable on an open set containing $\X$. A point $x \in \X$ is called a stationary point  if $\langle x'-x , \nabla f(x)\rangle \geq  0$ for all $ x' \in \X$. For $\Xc = \mathbb{R}^d$, a stationary point satisfies $\nabla f(x)=0$.  
\end{definition}	

We include an illustrative result showing that projected gradient descent converges asymptotically to a stationary point under appropriate smoothness and regularity conditions. This result can be generalized in numerous ways. For example, \cite[Theorem 10.15]{beck2017first} provides rates of convergence. A more complete treatment can be found in textbooks on nonlinear optimization \cite[see e.g][]{bertsekas1997nonlinear}. 

The result below covers two cases. The first assumes $\nabla f$ is Lipschitz, or, equivalently for twice differentiable functions, that the maximum eigenvalue of its Hessian is bounded above. The second relaxes this condition, only requiring regularity properties on the sublevel set of the initial iterate. This accommodates functions like $f(x)=x^4$, whose second derivative is unbounded. This result is possible because projected gradient descent with sufficiently small step-sizes is guaranteed to reduce cost in each iteration, so all iterates lie in certain sublevel sets. The restriction that $f$ has bounded sublevel sets is satisfied if the feasible region $\Xc$ is itself a bounded set or if the function is coercive, meaning $f(x) \to \infty$ as $\| x \| \to \infty$.  In problems where this is not naturally satisfied, it can sometimes be enforced by adding a small penalty function (e.g. a quadratic regularizer) to the objective. Recall that a point $x_\infty$ is said to be a limit point of a sequence $\{x_k\}$ if some subsequence converges to $x_\infty$. 
\begin{restatable}{lem}{convergencetostationary}\label{lem: pgd reaches global optimum}
	Consider the optimization problem $\min_{x \in \X} f(x)$ where $\X \subset \mathbb{R}^d$ is a closed convex set. Assume  $f$ is bounded below and its $\beta$--sublevel set  $\{x \in \Xc : f(x) \leq \beta \}$ is bounded for each $\beta \in \mathbb{R}$. Consider the sequence $x_{k+1} = {\rm Proj}_{\Xc}\left( x_k - \alpha \nabla f(x_k) \right)$ for $k\in \mathbb{N}$. 
	\begin{enumerate}
		\item \citep{beck2002convergence, beck2017first} Assume $f$ is differentiable on an open set containing $\X$ and  $\nabla f$ is Lipschitz continuous on $\Xc$ with Lipschitz constant $L$. 
		If $\alpha \in (0, 1/L]$, the sequence $\{x_k\}$ has at least one limit point and any limit point $x_{\infty}$ is a stationary point of $f(\cdot)$ on $\Xc$ satisfying $f(x_k) \downarrow f(x_\infty)$. 
		\item Given a fixed initial iterate $x_0$, suppose $f$ is continuously twice differentiable on an open set containing the sublevel set $\{x \in \Xc : f(x) \leq f(x_0)   \}$. For a sufficiently small $\alpha>0$, the sequence $\{x_k\}$ has at least one limit point and any limit point $x_{\infty}$ is a stationary point of $f(\cdot)$ on $\Xc$ satisfying $f(x_k) \downarrow f(x_\infty)$. 
	\end{enumerate}
\end{restatable}
\begin{proof}
	The proof for of the second part closely follows the proof for the first part as shown in \citep{beck2002convergence,beck2017first}. For brevity this is omitted, but to ensure reproducibility, details are given in the technical report \cite{technicalReport}.  
\end{proof}

\paragraph{Convergence rates under gradient dominance.}
Results like Lemma \ref{lem: pgd reaches global optimum} ensure that first order methods converge asymptotically to stationary points under mild regularity conditions. Then, even if the cost function is non-convex, such algorithms will converge toward a global optimum if all stationary points are optimal. A stronger property, called the Polyak-Lojasiewicz (PL) inequality and also commonly known as \textit{gradient dominance}, effectively requires that approximate stationary points are also approximately optimal. Combined with regularity conditions, this yields rates of convergence for first-order methods. Below, we introduce a notion of gradient dominance which might seem somewhat nonstandard, since many authors treat only unconstrained problems. For the unconstrained case, this result reduces to the well known PL inequality of \cite{polyak1963gradient}, $\min_{x' \in \R^d} f(x) \geq f(x) - \frac{c^2}{2\mu}\| \nabla f(x) \|_2^2$.

\begin{definition}
	\label{def:gradient_dominance}
	For $\X \subseteq \mathbb{R}^d$, we say $f$ is $(c,\mu)$--\textit{gradient dominated} over $\Xc$ if there exists constants $c>0$ and $\mu\geq 0$ such that
	\begin{align}\label{eq: second_order_grad_dom}
	\min_{x'\in \mathcal{X}} f(x')  \geq f(x) + \min_{x'\in \mathcal{X}} \, \left[ c\, \langle \nabla f(x),x'-x\rangle + \frac{\mu}{2} \norm{x-x'}^2_2 \right]  \,\, \quad \forall \, x\in \Xc. 
	\end{align}
	The function is said to be gradient dominated with degree one if $\mu=0$ and gradient dominated with degree two if $\mu>0$. 
\end{definition}
Any stationary point of a  gradient dominated function is globally optimal. To see this, note that if a point $x$ is stationary, meaning  $\langle \nabla f(x),x'-x\rangle \geq 0$ for every $x'\in \Xc$, then the minimizer of the right hand side in \eqref{eq: second_order_grad_dom} is $x$, implying $\min_{x'\in \mathcal{X}} f(x')  \geq f(x)$. More broadly, note how in \eqref{eq: second_order_grad_dom}, the optimality gap $\min_{x'\in \mathcal{X}} f(x') - f(x)$ can be bounded by a measure of how far $x$ is from stationarity, which is captured by the minimization problem on the right hand side.

Convex and strongly convex functions are gradient dominated. Recall that a differentiable function $f$ is said to be $\mu$--strongly--convex if it satisfies the inequality
\begin{equation}\label{eq:strongly_convex_def}
f(x') \geq  f(x)+\langle \nabla f(x)\,, \, x'-x  \rangle +  \frac{\mu}{2} \norm{x-x'}^2_2
\end{equation}
for every $x,x'\in \Xc$. Minimizing over $x'$ on each side of \eqref{eq:strongly_convex_def} shows that a $\mu$--strongly-convex function is $(1,\mu)$--gradient dominated. Similarly, convex functions satisfy \eqref{eq:strongly_convex_def} with $\mu =0$, and therefore are $(1,0)$--gradient dominated. Thus, gradient dominated functions satisfy a critical property of convex functions (relating the optimality gap to distance from stationarity). However, there are important classes of functions that are gradient dominated despite being nonconvex.   

Under gradient dominance conditions, popular first order optimization algorithms are assured to converge to the global minimum and a simple analysis provides finite time rates of convergence. As an illustrative result, part (a) of Lemma \ref{lemma:second_order_convergence_rate}  strengthens Lemma \ref{lem: pgd reaches global optimum} by providing an $\mathcal{O}(1/\sqrt{T})$ convergence rate. While part (b) of this lemma assumes $\Xc=\mathbb{R}^d$, \cite{karimi2016linear} also give a geometric convergence rate for projected gradient descent on constrained subsets, $\Xc \neq \mathbb{R}^d$ for $(1,L)$--gradient dominated functions with $L$-Lipschitz continuous gradients. Using similar arguments, it seems possible to also show a geometric rate for $(c,\mu)$--gradient dominated functions on constrained spaces when $\mu > 0$. 
\begin{restatable}[Convergence rates for gradient dominated smooth functions]{lem}{ratesgraddominance}
	\label{lemma:second_order_convergence_rate}
	Consider the problem, $\min_{x \in \Xc} f(x)$ where $\Xc \subseteq \R^d$ is nonempty. Assume $\nabla f$ is $L$--Lipschitz continuous on $\X$. Denote $f^*=\inf_{x'\in \Xc} f(x')$. Consider the sequence $x_{t+1} = {\rm Proj}_{\Xc}\left( x_t - \alpha \nabla f(x_t) \right)$.
	\begin{enumerate}
		\item Let $\Xc \subset \R^d$ be bounded. Set  $R=\sup_{x,x'\in \Xc} \| x-x'\|_{2}$ and $k=\sup_{x\in \Xc} \| \nabla f(x)\|_2$. If $\alpha \leq \min\{\frac{1}{k}, \frac{1}{L} \}$ and $f$ is $(c,0)$--gradient-dominated, then, 
		\[
		 f(x_T) - f^*  \leq \sqrt{\frac{2R^2c^2 \left( f(x_0) - f^* \right)}{\alpha T}}.
		\]
		\item Assume $\Xc=\R^d$ and $\alpha=1/L$. If $f$ is $(c,\mu)$--gradient-dominated for $\mu>0$, then, 
		\[ 
		f(x_T) - f^* \leq \left(1-\frac{\mu}{c^2 L}\right)^T  \left( f(x_0) - f^* \right).
		\]
	\end{enumerate}
\end{restatable}
\begin{proof}
	The proof of part (2) can be found in \citep{karimi2016linear,polyak1963gradient}. A result similar to Part (1) can be extracted from the textbook \citep{beck2017first}, which gives convergence rates to approximate stationary points. For brevity, this is omitted, but to ensure reproducibility, we give a detailed proof in the technical report \citep{technicalReport}.  
\end{proof}

%% file: lq_body.tex
\section{Motivation from linear quadratic control}
\label{sec:Motivation_LQ}
We first motivate and instantiate our general results for the special case of linear quadratic (LQ) control. Leveraging many of the closed form expressions available in this case, recent work of \cite{fazel2018global} showed that policy gradient methods converge to the globally optimal policy under some technical conditions. The key to their result is showing that the infinite horizon cost function, despite being non-convex, has no suboptimal stationary points (and is in fact gradient dominated). Given the presence of bad stationary points in Example \ref{ex: counterexample}, there must be some special problem structure driving this, but what? Quite different from \cite{fazel2018global}, our arguments involve classical understanding of the single period cost function underlying policy iteration, avoiding the complications of directly analyzing the infinite horizon cost function $\ell(\cdot)$. 

We highlight the two key properties of LQ control identified in Section \ref{subsec:contribution}. That is, we show that (a) the class of linear policies is closed under policy improvement and (b) the policy iteration  problem can be solved to optimality by a gradient method, since it is convex quadratic and therefore has no suboptimal stationary points. A short proof then shows how these two conditions imply that $\ell(\cdot)$ has no suboptimal station points. Like \cite{fazel2018global}, we simplify the presentation by studying deterministic LQ control, but it is easy to allow for noisy dynamics.


\begin{example}[Linear Quadratic Control]\label{ex: lq control}
	For symmetric positive definite matrices $R$ and $C$, we have the following optimal control problem:
	\begin{align*}
	{\rm Minimize} \quad  &  \sum_{t=0}^{\infty} \gamma^t \left( a_t^\top R a_t  + s_t^{\top} C s_t \right)\\
	{\rm Subject\,\, to} \quad & s_{t+1} = As_t + Ba_t, \quad s_0 \sim \rho
	\end{align*} 
	where $s_t \in \mathbb{R}^n$ is a continuous state variable and $a_t \in \mathbb{R}^k$ is the action chosen at time $t$. We assume that the second moment of the initial distribution $\E_{\rho} \left[ s_0 s_0^\top \right]$ is finite and positive definite. In this setting, a linear policy $\pi_{\theta}(s) = \theta s$ is known to be optimal for some $\theta \in \mathbb{R}^{k\times n}$. See for example \cite{bertsekas1995dynamic,bertsekas2011approximate,evans2005introduction}. We consider searching for the optimal $\theta$ via a gradient method. Unfortunately, the loss function $\ell(\theta)=(1-\gamma) \E_{\rho} [J_{\pi_{\theta}}(s_0)]$ is non-convex (see Appendix B in \cite{fazel2018global}), making it unclear if gradient descent on $\ell(\theta)$ would reach the global minimum. 
	
	For LQ control, if a linear policy $\pi_\theta$ is applied from a state $s_0$, then unrolling the linear dynamics we have $s_t= (A+B\theta)s_{t-1}= \cdots = (A+B\theta)^t s_0$. Then, we can write the cost-to-go function as:
	\[ 
	J_{\pi_\theta}(s_0) = \sum_{t=0}^{\infty} \gamma^t \left( s_t^\top \theta^\top R \theta s_t  + s_t^{\top} C s_t \right) = s_0^\top \underbrace{\left[ \sum_{t=0}^{\infty} \gamma^t \left( (A+B\theta)^t \right)^{\top}\left( \theta^{\top}R\theta +C \right) \left(A+B\theta\right)^t \right]}_{:=K_\theta \, \succeq \, 0}  s_0. 
	\] 
	A linear policy  $\pi_\theta$ is said to be \emph{stable} if its cost-to-go is finite from all initial states, or equivalently, if all eigenvalues of the matrix $\sqrt{\gamma}(A+B\theta)$ lie strictly within the unit circle. We	let $\Theta_{\text{S}} \subset \mathbb{R}^{k \times n}$ denote the set of all parameters defining stable linear policies and assume that this set is nonempty. One can show\footnote{This is stated formally in Lemma \ref{lem:lq_stability} in Appendix \ref{appendix: example details}. For completeness a proof is given in the technical report \citep{technicalReport}, but this is really a minor modification of standard linear systems theory to account for the effect of discounting. Indeed, if we consider the linear dynamical system $s_{t}=(A+B\theta)s_{t-1}$, then our condition is precisely equivalent to requiring that $\sqrt{\gamma}^t s_t \to \infty$ for every initial state. This is the appropriate definition of stability in discounted problems and, when $\gamma=1$, our definition reduces to the standard definition of a stable linear policy in undiscounted problems.} 		
	that $\ell(\theta)<\infty$ if $\theta \in \Theta_S$ and $\ell(\theta)=\infty$ if $\theta \notin \Theta_{\text{S}}$. 
	
	Even though the total cost function $\ell(\theta)$ is non-convex, it can be shown that starting from a stable linear policy, policy iteration (PI) converges to an optimal policy by solving a sequence of simpler single period optimization problems \citep{kleinman1968, hewer1971iterative} that arise when applying the Bellman operator, which in this case can be written as
	\begin{align}\label{eq: Bellman optimality operator LQ}
	(TJ_{\pi_{\theta}})(s) 
	= \min_{a\in \mathbb{R}^k} \, \left[\underbrace{a^\top R a +s^\top C s +  \gamma(As+Ba)^\top K_{\theta}(As+Ba)}_{= \, Q_{\pi_{\theta}}(s,a)}\right]. 
	\end{align}
	A single iteration of PI updates the stable linear policy $\pi_{\theta}$ to a new policy $\pi^{+}$ that selects the action $\pi^{+}(s) = \argmin_{a} Q_{\pi_{\theta}}(s,a)$. This can equivalently be expressed as $T_{\pi^{+}} J_{\pi_\theta}= TJ_{\pi_{\theta}}$. Typically, a PI update requires solving a unique optimization problem for each state, but given the convex quadratic nature of the problem in \eqref{eq: Bellman optimality operator LQ}, it can be easily checked that $\pi^{+}(s) = \thetabar s$ where $\thetabar=-\gamma(R+\gamma B^\top K_{\theta} B)^{-1} B^\top K_{\theta} A$. Thus, starting from a linear policy, a PI update yields yet another linear policy implying that for LQ control, the class of linear policies is closed under policy improvement. Policy iteration steps are sometimes called policy improvement steps because the new policy $\pi^+$ is assured to have lower cost-to-go, meaning that $J_{\pi^{+}}(s) \leq J_{\pi_{\theta}}(s)$ for all $s\in \Sc$ and the improvement is strict at some set of states if $\pi_{\theta}$ is not an optimal policy. Crucially, this implies that for LQ control the PI update $\pi^+$ is also a stable policy (see Lemma \ref{lem: lqr loss function} for a formal statement).

	A useful interpretation of the PI update is to view it as the minimizer of a weighted policy iteration cost,
	\begin{equation}\label{eq:weighted_pi_objective}
	\Bc(\theta' | \eta, J_{\pi_{\theta}}) :=  \intop (T_{\pi_{\theta}} J_{\theta})(s) \, \eta(ds) =  \intop Q_{\pi_\theta}(s, \pi_{\theta'}(s)) \, \eta(ds),
	\end{equation}
	over policy parameters $\theta'$. Under appropriate conditions on $\eta$, the solution is unique\footnote{Due to the quadratic structure of the objective, it is enough that $\eta$ has a finite and strictly positive definite second moment matrix. This is similar to our assumption about the initial distribution $\rho$ and it can be established using the same argument as given in the proof of Lemma \ref{lem: lqr stationary}.}. From \eqref{eq: Bellman optimality operator LQ}, note that $Q_{\pi_\theta}(s,a)$ is a convex quadratic function of action $a$, which implies $Q_{\pi_\theta}(s,\theta' s)$ is a convex quadratic function of $\theta'$ and so is \eqref{eq:weighted_pi_objective}. This shows that the weighted PI objective has no suboptimal stationary points, the second key property we identified in Section \ref{subsec:contribution}.

\end{example}
Because the per-stage cost functions in LQ control are unbounded, this example is technically beyond the scope of the problem formulation in Section \ref{section:problem_formulation}. Thankfully, the properties of Bellman operators that underlie our analysis (for example, monotonicity) hold for LQ control. See Lemma \ref{lem: bellman operators for LQ} in Section \ref{app:LQ} for details. As a consequence, \emph{the proofs} of our general results will essentially apply without modification to stable policies in LQ control. Nevertheless, to be formal, any results about LQ control will clearly specify that they apply to Example \ref{ex: lq control} and standalone proofs are given for completeness.  

To discuss policy gradient methods, we first need some smoothness properties of $\ell(\cdot)$. Beginning with an initial stable policy $\theta_0$ which incurs finite cost, first-order algorithms with an appropriate step-size are assured to decrease cost on every iteration, meaning that iterates remain in the sublevel set $\{ \theta\in \Theta : \ell(\theta) \leq \ell(\theta_0)  \} \subset \Theta_S$. The next lemma establishes regularity conditions on these sublevel sets which are sufficient to apply optimization results, like Lemma \ref{lem: pgd reaches global optimum}, to show that gradient descent converges to a stationary point of $\ell(\cdot)$. 
\begin{restatable}{lem}{LQloss}
	\label{lem: lqr loss function}
	Consider the LQ control problem formulated in Example \ref{ex: lq control}. The set $\Theta_{\text{S}}$ is open and $\ell$ is twice continuously differentiable on $\Theta_{\text{S}}$.  For any $\alpha\in \mathbb{R}$, the sublevel set $C_{\alpha}:=\left\{\theta \in \mathbb{R}^{n\times k} : \ell(\theta) \leq \alpha \right\}$ is a compact subset of $\Theta_{\text{S}}$ and, if it is nonempty, $\sup_{\theta \in C_{\alpha}}  \, \| \nabla^2 \ell(\theta) \|_2   < \infty$. 
\end{restatable} 
\begin{proof}
	These properties follow from \citep{toivonen1985globally, rautert1997computational}. Some additional details are provided in Appendix \ref{app:LQ}
\end{proof}
With this background, a simple proof shows that for LQ control, the policy gradient loss function has no suboptimal stationary points despite being non-convex. Essentially, for any stable linear policy that is suboptimal, we show that moving along the line segment toward a policy iteration update forms a descent direction, implying that it cannot be a stationary point. The two properties we identified, convexity of the weighted PI objective and closure with respect to the class of linear policies, are critical for this argument.
\begin{restatable}{lem}{LQstationary}
	\label{lem: lqr stationary}
	For the LQ control problem formulated in Example \ref{ex: lq control}, any stable linear policy $\theta$ satisfies 
	$\nabla \ell(\theta) = 0$ if and only if $J_{\pi_{\theta}} = J^*$. 
\end{restatable} 
\begin{proof}[Proof sketch]
	Consider a stable linear policy $\pi_\theta$ and take $\pi_\thetabar$ to be a policy iteration update. Standard analysis of policy iteration, using monotonicty of the Bellman operator, shows that $J_{\pi_{\thetabar}} \preceq TJ_{\pi_{\theta}} \preceq J_{\pi_{\theta}}$. Here, the second inequality is strict at some set of states unless $\pi_\theta$ is an optimal policy. This implies $\ell(\thetabar) < \ell(\theta)$ when $\pi_\theta$ is suboptimal. Now, consider a soft policy iteration update $\theta^{\alpha} = (1-\alpha)\theta + \alpha \thetabar \,\,\, \forall \,\alpha \in [0,1]$. Leveraging convexity of the policy iteration objective in \eqref{eq:weighted_pi_objective}, a short argument shows $\frac{d}{d\alpha} \ell(\theta^{\alpha}) \vert_{ \alpha=0} \leq 0$ and this inequality is strict unless $\pi_{\theta}$ is an optimal policy. See Appendix \ref{app:LQ} for a detailed proof. 
\end{proof} 
This idea of constructing a descent direction is strongly reminiscent to the arguments in \cite{kakade2002approximately} for finite MDPs.
While we find this proof to be intuitive, it relies not just on the closure property, but on convexity of the policy class\footnote{The class of linear policies is convex. That is, the policy $\alpha \pi_{\theta} + (1-\alpha)\pi_{\thetabar}$ is a linear policy for any given linear policies $\pi_{\theta}, \pi_{\thetabar}$ and some $\alpha \in [0,1]$. However, the class of threshold policies, used in the optimal stopping and the inventory control problems is not convex. If $\pi_{\theta}(s)=\mathbbm{1}(s\leq \theta)$ for $\theta \in \mathbb{R}$ is a threshold policy, then $\frac{1}{2} \pi_{\theta}  + \frac{1}{2} \pi_{\theta'}$ is not a threshold policy when $\theta\neq \theta'$.} as well as convexity of the policy iteration cost function, which will not hold in all of our examples. One contribution of this paper is to find a clean generalization of this argument, relaxing convexity conditions into Condition \ref{cond: 2ab} in the next section.

%% file: stationary_points.tex

\section{General results}
\label{sec:stationary_points}


We now generalize some of the insights discussed above for the LQ control example to identify properties which ensure the policy gradient objective has no suboptimal stationary points. In the next section, we show these properties hold for various problems settings beyond LQ control.

\subsection{Conditions on the policy iteration cost function}\label{subsec:conditions} 
Consider the weighted policy iteration or the ``Bellman'' cost function introduced in \eqref{eq:weighted_pi_objective}.
\begin{equation*}
\Bc(\pibar \mid \eta, J_{\pi})  = \intop (T_{\pibar} J_{\pi})(s) \, \eta(ds) = \intop Q_{\pi}(s,  \pibar(s)) \, \eta(ds),  
\end{equation*}
for a probability distribution $\eta$ over $\Sc$ and $J_{\pi}\in \Jc$. Here, the final equality follows by noting that  $(T_{\pibar} J_{\pi})(s) \equiv Q_{\pi}(s,\pibar(s))$ from \eqref{eq: q to bellman}. This Bellman cost function is a \textit{single period} objective, considering the cost-to-go of following $\pibar$ for a single period and following $\pi$ thereafter. We overload notation to write $\Bc(\theta \mid \eta, J)  = \Bc(\pi_\theta \mid \eta, J)$. 
When the state space is discrete and $\eta(s)>0$ for all $s \in \Sc$, classic policy iteration update can be equivalently written as,
\begin{equation}\label{eq:weighted_PI}
\pi_{k+1}= \argmin_{\pi \in \Pi} \,\, \Bc(\pi| \eta, J_{\pi_k}).
\end{equation} 
Policy iteration, like value iteration, indirectly optimizes the infinite horizon cost-to-go, $\ell(\cdot)$ by solving the sequence of simpler single period problems in \eqref{eq:weighted_PI}. On the other hand, policy gradient methods aim to directly minimize $\ell(\pi_\theta)$. Despite this crucial difference, our approach is to infer properties of the complex multi-period objective $\ell(\cdot)$ using some structure present in the single period problems. We outline this below.

\paragraph{Differentiability.}
Before arguing about any convergence properties, we first need conditions for the policy gradient itself to be well defined. Condition \ref{cond: diff} below states smoothness conditions, related to partial differentiability of the Bellman objective $\Bc(\cdot)$, that ensure $\ell(\cdot)$ is differentiable and its gradients satisfy a convenient formula used in practical implementations \citep{silver2014deterministic, marbach2001simulation,sutton2018reinforcement}. Condition \ref{cond: diff} arises quite naturally in calculating the derivatives of $\ell(\theta)$. To see this, we refer the readers to a short proof of Lemma \ref{lem: PG theorem} in Appendix \ref{appendix B}. 
\setcounter{cond}{-1}
\begin{cond}[Differentiability]\label{cond: diff}
	For each $\theta \in \Theta$, the functions $\thetabar \mapsto \Bc(\thetabar | \eta_{\pi_{\theta}}, J_{\pi_{\theta}})$ and $\thetabar \mapsto \Bc(\theta | \eta_{\pi_{\thetabar}}, J_{\pi_{\theta}})$ are continuously differentiable on an open set containing $\theta$. 
\end{cond}
As $\Bc(\thetabar | \eta_{ \pi_{\theta}}, J_{\pi_{\theta}} ) =  \intop Q_{\pi_{\theta}}(s, \pi_{\thetabar}(s)) \, \eta(ds)$, differentiability in $\thetabar$ follows if  $Q_{\pi_{\theta}}(s,\pi_{\thetabar}(s))$ is differentiable almost everywhere and the exchange of derivative and integral is permitted. Also note that $\Bc(\theta | \eta_{ \pi_{\thetabar}}, J_{\pi_{\theta}})= \intop J_{\pi_{\theta}}(s) \, \eta_{\pi_{\thetabar}}(ds)$. Thus, differentiability in $\thetabar$  is related to the existence of a weak derivative of the state occupancy measure \citep{pflug1988derivatives, pflug1990line}. A large literature  studies sufficient conditions for differentiability \citep{rhee2017lyapunov, glasserman1991gradient, asmussen2007stochastic}. We do not try to advance that literature, instead focusing on the convergence of policy gradient methods when these derivatives are well defined.

The next lemma writes the gradients of $\ell(\theta)$ in a form that may illuminate the sharp connections with policy iteration. In making an update to the policy parameter $\theta$, policy iteration within the parameterized policy class would solve $\min_{\thetabar \in \Theta} \Bc( \thetabar \mid \eta_{ \pi_{\theta}}, J_{\pi_{ \theta}})$. A policy gradient update takes a gradient step with respect to $\thetabar$ instead of solving this single period problem to optimality. It is through this viewpoint that we avoid analyzing $\ell(\cdot)$ directly. 
\begin{restatable}[Policy gradient theorem]{lem}{pgTheorem}\label{lem: PG theorem}
	Under Condition \ref{cond: diff}, $\ell(\theta)$ is continuously differentiable and
	\[
	\nabla \ell(\theta) =  \nabla_{\thetabar} \, \Bc(\thetabar \mid \eta_{\pi_\theta}, J_{\pi_\theta}) \bigg\vert_{\thetabar=\theta}  
	\]
\end{restatable}
\begin{rem}
When the exchange of a certain integral and derivative is permitted, this statement reduces to $\nabla \ell(\theta) = \intop \nabla_{\thetabar} Q_{\pi_{ \theta}}(s, \pi_{ \thetabar}(s)) \vert_{\thetabar=\theta} \eta_{ \pi_{\theta}}(ds)$. Our presentation differs from the familiar form of the policy gradient theorem  \citep{sutton2018reinforcement}, which is written as $\nabla \ell(\theta) = \E\left[ Q_{\pi_{\theta}} (s,a)  \nabla \log \pi_{\theta}(a|s) \right]$ where $\pi_{\theta}(a|s)$ is the probability of selecting a deterministic action $a$ in state $s$ and the expectation is taken over the distribution of states and actions under $\pi_{\theta}$. This form is useful for gradient estimation, but it applies only to stochastic policies and seems to obscure connections with the gradient of the weighted PI objective. The expression in Lemma \ref{lem: PG theorem} is more general and can be applied to both deterministic and stochastic policies (by taking $\pi_{\theta}(s)$ to be a probability vector as in Section \ref{sec:examples}). Lemma \ref{lem: PG theorem} was previously derived under somewhat stringent regularity conditions by \cite{silver2014deterministic}. Condition \ref{cond: diff} and our short proof appear to be new.
\end{rem}

\paragraph{Closure under policy improvement.}

We now introduce one of our main conditions, which we call closure under policy improvement. This is a consistency condition which essentially says that the policy improvement update can be solved within the policy class.


	\begin{cond}[Closure under policy improvement]\label{cond: closure}
		For each $\pi \in \Pi_\Theta$ , there exists $\pi^+ \in \Pi_\Theta$  such that $\Bc(\pi^+ | \eta_{\pi}, J_{\pi}) = \min_{ \pi' \in \Pi} \Bc(\pi' | \eta_{\pi}, J_{\pi})$.  
	\end{cond}
 \jb{It is not hard to show that Condition \ref{cond: closure} is equivalent to the following one: for each $\pi \in \Pi_{\Theta}$ there exists $\pi^+ \in \Pi_{\Theta}$ such that $(T_{\pi^+} J_{\pi})(s)=(TJ_{\pi})(s)$ almost surely under $s$ drawn from the occupancy measure $\eta_{ \pi}$. Under assumption \ref{ass: exploratory rho}, $\eta_{\pi^*}\ll \eta_{\pi}$ and so this condition can be thought of as imposing a closure property at all relevant parts of the state space.}

This closure assumption accommodates interesting examples in which a restricted class of policies is naturally aligned with the decision task, like the class of linear policies in LQ control or threshold policies for optimal stopping. We emphasize that this condition is weaker than requiring the policy class to contain nearly all stochastic policies. However, it is stronger than just requiring the policy class to contain an optimal policy. An extension in section \ref{sec:approximate_closure} bounds the optimality gap of stationary points when the policy is not closed but satisfies a relaxed closure condition. 

Some condition along these lines appears to be necessary. Indeed, Example \ref{ex: counterexample} in the introduction showed an extremely simple problem for which policy gradient methods can get stuck in a bad local minimum even though the policy class contains an optimal policy. There, we consider a two state ($\Sc=\{s_L, s_R \}$) deterministic MDP with actions corresponding to moving left ($L$) and right ($R$) respectively. An action $a\in \Ac = [0,1]$ indicates the probability of choosing the action right. We consider a restricted class of policies of the form $\pi_{\theta}(s_L)=\pi_{\theta}(s_R)=\theta$, which plays action $R$ in either state with probability $\theta \in [0,1]$. Simple calculations show why this policy class is not closed under policy improvement. For any policy $\pi_{\theta}$ we can write the Q-function as 
\begin{align*}
	Q_{\pi_{\theta}}(s_L, a) = (1-a)\cdot 1 + a \cdot 2 + \gamma \left( (1-a) J_{\pi_{\theta}}(s_L) + a J_{\pi_{\theta}}(s_R)\right) \\
	Q_{\pi_{\theta}}(s_R, a) = (1-a)\cdot 2 + a \cdot 0 + \gamma \left( (1-a) J_{\pi_{\theta}}(s_L) + a J_{\pi_{\theta}}(s_R)\right)
\end{align*}
Consider a policy where $\theta$ is nearly zero, so $\pi_{\theta}$ moves left with high probability. It is easy to check that $0=\argmin_{a\in [0,1]} Q_{\pi_{\theta}}(s_L, a)$. That is, in state $s_L$, it is optimal for the decision maker to move left assuming that the policy $\pi_{\theta}$ (which almost always moves left) will be applied in future periods. A similar argument shows that $1=\argmin_{a\in [0,1]} Q_{\pi_{\theta}}(s_R,a)$. Clearly, this policy iteration update is not contained in the restricted one dimensional policy class $\{\pi_{\theta} : \theta \in [0,1] \}$.

\paragraph{Stationary points of the weighted PI objective.}
As a first order method, policy gradients require additional local optimization structure to succeed. The following conditions ensure that first order methods are suitable for solving the weighted policy iteration problem. 
\begin{subtheorem}{cond}\label{cond: 2ab}
\begin{cond}[Stationary points of the weighted PI objective]\label{cond: stationary}
	For each $\pi \in \Pi_\Theta$, the function $\theta \mapsto \Bc(\theta \mid \eta_{\pi}, J_\pi)$ has no sub-optimal stationary points. 
\end{cond}
\begin{cond}[Gradient dominance of the weighted PI objective]\label{cond: convexity_strong} 
For any $\pi \in \Pi_{\Theta}$, the function $\theta \mapsto \Bc(\theta \mid \eta_{\pi}, J_{\pi})$ is $(c,\mu)$--gradient-dominated over $\Theta$. 
\end{cond}
\end{subtheorem}
It is worth emphasizing that the single period Bellman objective $\bar{\theta} \mapsto \Bc(\bar{\theta} | \eta_{\pi_{\theta}} , J_{\pi_{\theta}})$ is often much simpler than the infinite horizon objective $\ell(\thetabar)$. In LQ control it is a convex quadratic function and is therefore gradient dominated. For finite state and action MDPs, say for instance in Example \ref{ex: counterexample}, it is linear and hence is also gradient dominated, even though we showed that the total cost function $\ell(\theta)$ is non-convex and can have suboptimal local minima. Even for complex neural networks, a very active literature studies the quality of stationary points and local minima for certain single period loss functions \citep{du2018power, livni2014computational}.

\subsection{Closed policy classes and optimality of stationary points.}
Our first result establishes that the policy gradient objective has no suboptimal stationary points when the policy class is closed under policy improvement and the single period Bellman objective has no suboptimal stationary points. Having identified the right conditions and notation, the proof falls into place. 
\begin{thm}\label{thm: no bad stationary points general}
	Suppose Conditions \ref{cond: diff}, \ref{cond: closure}, and \ref{cond: stationary} hold. Then, $\ell$ is continuously differentiable and $\theta\in \Theta$ is a stationary point of $\ell(\cdot)$ if and only if $\ell(\pi_{\theta}) = \ell(\pi^*)$. 
\end{thm}

\paragraph{Proof of Theorem \ref{thm: no bad stationary points general}.} 
We first give a key lemma which establishes a Bellman-type equation that holds when the single period objective $\thetabar \mapsto \Bc(\thetabar \mid \eta_{\pi_\theta}, J_{\pi_\theta})$ has no bad stationary points. 
\begin{lem}\label{lem: bellman for stationary points}
	Suppose Condition \ref{cond: stationary} is satisfied. If $\theta$ is a stationary point of $\ell:\Theta \to \mathbb{R}$, then 
	\[ 
	\intop  J_{\pi_\theta} d\eta_{\pi_\theta}  = \min_{\pi\in \Pi_{\Theta}}\intop \left( T_{\pi}J_{\pi_{\theta}}\right) d\eta_{\pi_\theta}.
	\] 
\end{lem}
\begin{proof} 
	If $\theta$ is a stationary point of $\ell: \Theta \to \mathbb{R}$, then by the policy gradient theorem in Lemma \ref{lem: PG theorem}, it is also a stationary point of the function $\thetabar \mapsto \Bc(\thetabar \mid \eta_{\pi_\theta}, J_{\pi_\theta})$. Since Condition \ref{cond: stationary} holds, this implies 
	\begin{equation*}
		\Bc(\theta   \mid \eta_{\pi_\theta}, J_{\pi_\theta}) =  \min_{\thetabar \in \Theta} \Bc(\thetabar \mid \eta_{\pi_\theta}, J_{\pi_\theta}).
	\end{equation*}
	Recalling the definition of $\Bc(\theta   \mid \eta, J_\pi)$ in \eqref{eq:weighted_pi_objective} lets us rewrite both sides of this equation as,
	\begin{align*}
		\intop J_{\pi_\theta} d\eta_{ \pi_{\theta}} = \intop \left[ T_{\pi_\theta} J_{\pi_\theta}\right] d\eta_{ \pi_{\theta}} = \Bc(\theta   \mid \eta_{\pi_\theta}, J_{\pi_\theta}) = \min_{\bar{\theta}\in \Theta} \Bc(\bar{\theta} \mid \eta_{\pi_\theta}, J_{\pi_\theta}) = \min_{\bar{\theta}\in \Theta} \intop \left[T_{\pi_{\thetabar}}J_{\pi_{\theta}} \right] d\eta_{\pi_\theta}.
	\end{align*}
\end{proof} 
Next, in Lemma \ref{lem: on average bellman equation}, we state an ``average'' form of Bellman's equation which shows that under an exploratory initial distribution (see Assumption \ref{ass: exploratory rho}), an optimal policy has zero average Bellman error. 

\begin{restatable}[On average Bellman equation]{lem}{onAvgBellman}\label{lem: on average bellman equation}
	For any $\pi \in \Pi$, 
	\begin{equation*}
		\ell(\pi)  = \ell(\pi^*)  \,\,  \iff \,\,  \intop \left(J_{\pi} -TJ_{\pi} \right) d\rho = 0.
	\end{equation*}
\end{restatable}
\begin{proof}
	See Appendix \ref{app: on average Bellman} for a detailed proof.
\end{proof}
We now complete the proof of Theorem \ref{thm: no bad stationary points general}. 
\begin{proof}[Proof of Theorem \ref{thm: no bad stationary points general}]
	To show the first direction, note that $\ell(\theta) = \ell(\pi^*)$ implies that $\theta$ is a minimizer of $\ell(\cdot)$. By the first order necessary conditions of optimality, $\theta$ must be a stationary point of $\ell(\cdot)$. To prove the other direction, suppose that $\theta$ is a stationary point of $\ell(\cdot)$. Then,
	\[
	\intop  J_{\pi_\theta} d\eta_{\pi_\theta} = \min_{\pi\in \Pi_{\Theta}}\intop \left[ T_{\pi}J_{\pi_{\theta}}\right] d\eta_{\pi_\theta} = \intop  \left[TJ_{\pi_\theta}\right] d\eta_{\pi_\theta}. 
	\]
	where the first equality follows from Lemma \ref{lem: bellman for stationary points} (implied by Condition \ref{cond: stationary}) while the second equality uses the closure property in Condition \ref{cond: closure}.
	By the definition in \eqref{eq: discounted_state_occupancy_measure}, $\eta_{ \pi}\succeq (1-\gamma) \rho$. Using that $J_{\pi} \succeq T J_{\pi}$, we have
	\[ 
	0 = \intop \left[ J_{\pi_\theta}- TJ_{\pi_\theta}\right] d \eta_{\pi_\theta} \geq  (1-\gamma) \intop \left[ J_{\pi_\theta} - TJ_{\pi_\theta}\right] d\rho \geq  0.
	\]
	The ``on average Bellman equation'' in Lemma \ref{lem: on average bellman equation} lets us conclude that $\ell(\theta)=\ell(\pi^*)$.
\end{proof}

\subsection{Convergence rates for policy gradient methods.}
\label{subsec:PG_rates}
Theorem \ref{thm: no bad stationary points general}, which guarantees that the policy gradient objective has no suboptimal stationary points, is only an asymptotic result when viewed in context of Lemma \ref{lem: pgd reaches global optimum}. Our next result strengthens this by showing that $\ell(\theta)$ is gradient dominated (though possibly non-convex) for closed policy classes if the simpler weighted PI objective is gradient dominated. Recall, gradient dominance and smoothness of $\ell(\cdot)$ often imply fast convergence rates for first-order methods, using results like Lemma \ref{lemma:second_order_convergence_rate}. Our investigation here is inspired by \cite{fazel2018global}, who showed gradient dominance for LQ control by a careful manipulation of closed form expressions available in linear systems.   

Our result in Theorem \ref{thm: convergence_rates} below relies on a constant $\kappa_{\rho}$ that measures the efficacy of an exploratory initial distribution in a more refined way than Assumption \ref{ass: exploratory rho}. We call this the \textit{effective concentrability coefficient}, since it plays a role similar to the concentrability coefficients that are widely used in the analysis of approximate value and policy iteration algorithms \citep{munos2003error, munos2007performance, munos2008finite, farahmand2010error, kakade2002approximately, scherrer2014local, geist2017bellman}.  Intuitively, $\kappa_\rho$ captures how errors in the cost-to-go functions manifest in Bellman errors that are detectable by sampling from the exploratory initial distribution $\rho$. The somewhat opaque definition  in \eqref{def:concentrability_measure} is precisely the quantity we need in our analysis. Section \ref{sec:concentrability} provides more insights into the definition along with more interpretable bounds for $\kappa_{\rho}$. For example, we show it is always bounded as $\kappa_{\rho} \leq \| \frac{ d\eta_{ \pi^*}  }{d\rho}\|_{\infty}$, where this worst-case likelihood ratio has featured prominently in prior work as a measure of distribution shift. 
\begin{definition}
	Define the effective concentrability coefficient $\kappa_\rho$ for the class of cost-to-go functions $\Jc_{\Theta}=\{J_{\pi_\theta} : \theta \in \Theta \}$ to be the smallest scalar such that 
	\begin{align}
	\label{def:concentrability_measure}
	\|  J  - J^* \|_{1,\rho} \leq \frac{\kappa_{\rho}}{(1-\gamma)} \|  J  - TJ\|_{1, \rho} \quad \forall J \in \Jc_{\Theta}.
	\end{align}
	If no such scalar exists then we say $\kappa_{\rho} = \infty$. 
\end{definition}
Theorem \ref{thm: convergence_rates} gives a gradient dominance condition on $\ell(\cdot)$ for closed policy classes under Condition \ref{cond: convexity_strong}. Subsequently, Corollary \ref{cor: convexity to gradient dominance of pg} holds as (strongly) convex functions are gradient dominated. 


\begin{restatable}[]{thm}{mainthm}\label{thm: convergence_rates}
	If Conditions \ref{cond: diff}, \ref{cond: closure}, and \ref{cond: convexity_strong} hold, then $\ell(\cdot)$ is $\left(\frac{\kappa_{\rho}}{(1-\gamma)}\cdot c\, ,\, \frac{\kappa_{\rho}}{(1-\gamma)}\cdot \mu \right)$--gradient dominated.
\end{restatable}
\begin{cor}\label{cor: convexity to gradient dominance of pg}
	Suppose Conditions \ref{cond: diff} and \ref{cond: closure} hold. If, for every $\pi \in \Pi_{\Theta}$, the function $\theta \mapsto \Bc(\theta \mid \eta_\pi, J_{\pi})$ is convex, then $\ell(\theta)$ is gradient dominated with degree one. If $\theta \mapsto \Bc(\theta \mid \eta_\pi, J_{\pi})$ is strongly convex, then $\ell(\theta)$ is gradient dominated with degree two. 
\end{cor}

\paragraph{Proof of Theorem \ref{thm: convergence_rates}.}
Our proof can be divided into two key steps. First, we use closure property (Condition \ref{cond: closure}) to bound the optimality gap of a policy, $\ell(\pi) - \ell(\pi^*)$, by the improvement under a weighted PI update. The second step uses the policy gradient theorem in Lemma \ref{lem: PG theorem} to translate this inequality into a gradient dominance condition on $\ell(\cdot)$. It is noteworthy that our results crucially depend on using an exploratory initial distribution under which $\kappa_\rho < \infty$.
\begin{proof}
	We first derive a consequence of the closure condition:
	\begin{align*}
		\ell(\pi_{\theta}) - \min_{\pi\in \Pi}\ell(\pi) = (1-\gamma) \intop \left[ J_{\pi_{\theta}} - J^* \right] d\rho &\overset{(a)}{=} (1-\gamma)\| J_{\pi_{\theta}} - J^{*}\|_{1,\rho} \\
		&\overset{(b)}{\leq} \kappa_\rho \|  J_{\pi_{\theta}} - TJ_{\pi_{\theta}} \|_{1, \rho} \\
		&\overset{(c)}{\leq} \frac{\kappa_\rho}{(1-\gamma)} \|  J_{\pi_{\theta}} - TJ_{\pi_{\theta}} \|_{1, \eta_{\pi_{\theta}}} \\
		&= \frac{\kappa_\rho}{(1-\gamma)} \intop \left[ J_{\pi_\theta} - T J_{\pi_{\theta}} \right] d\eta_{\pi_{\theta}}\\
		&\overset{(d)}{=} \frac{\kappa_\rho}{(1-\gamma)} \left(  \intop J_{\pi_\theta} d\eta_{\pi_{\theta}}  -   \min_{\pi \in \Pi_{\Theta}}  \intop \left[ T_{\pi} J_{\pi_\theta} \right] d\eta_{\pi_{\theta}} \right) \\
		&= \frac{\kappa_\rho}{(1-\gamma)} \left( \Bc(\theta \mid \eta_{\pi_{\theta}}, J_{\pi_{\theta}}) - \min_{\theta' \in \Theta} \, \Bc(\theta' \mid \eta_{\pi_{\theta}}, J_{\pi_{\theta}})  \right).
	\end{align*}
	Here (a) uses that $J_{\pi_{\theta}} \succeq J^*$, (b) applies the definition of  $\kappa_\rho$ in \eqref{def:concentrability_measure}, (c) uses that $\eta_{\pi_{\theta}} \succeq (1-\gamma) \rho$ (see the definition in \eqref{eq: discounted_state_occupancy_measure}) and (d) uses closure property of the policy class. 
	
	By Condition \ref{cond: convexity_strong}, $\thetabar \mapsto \Bc(\thetabar \mid \eta_{\theta}, J_{\pi_{\theta}})$ is ($c$, $\mu$)--gradient dominated. Using gradient dominance and the policy gradient theorem in Lemma \ref{lem: PG theorem}, we find
	\begin{align*}
		\Bc(\theta \mid \eta_{\pi_{\theta}}, J_{\pi_{\theta}}) - \min_{\theta' \in \Theta} \Bc(\theta' \mid \eta_{\pi_{\theta}}, J_{\pi_{\theta}}) &\leq - \min_{v \in \Theta} \, \left[ c \, \langle \nabla_{\thetabar} \, \Bc(\thetabar \mid \eta_{\pi_{\theta}}, J_{\pi_{\theta}}) \bigg\vert_{\thetabar = \theta}, v-\theta \rangle + \frac{\mu}{2} \norm{v-\theta}^2_2 \right] \\
		&\leq - \min_{v \in \Theta} \, \left[ c \, \langle \nabla_{\theta} \ell(\theta), v-\theta \rangle + \frac{\mu}{2} \norm{v-\theta}^2_2 \right].
	\end{align*}
	Combining this with the preceding calculation yields the desired result. 
\end{proof}

\subsection{Beyond closed policy classes: the case of non-stationary policies.}
\label{subsec:nonstationary_policies}
For finite horizon problems with a class of non-stationary policies, we can guarantee that the policy gradient objective has no spurious local minima under a much weaker condition. Rather than require the policy class to be closed under improvement, it is sufficient that the policy class contains an optimal policy\footnote{This is a weaker property as closure of the policy class implies that it contains an optimal policy.}. For this reason, our theory will cover  a broad variety of finite horizon dynamic programming problems for which structured policy classes are known to be optimal, \jb{even if the policy class is not closed under policy improvement}. Interestingly, this result relies critically on the use of a non-stationary policy class. Recall how Example \ref{ex: counterexample} shows that for stationary policy classes, policy gradient methods can get stuck in bad local minima even if the policy class contains an optimal policy.

We can state our formal result without introducing new notation for the finite horizon setting by a well known trick that treats finite-horizon time-inhomogeneous MDPs as a special case of infinite horizon MDPs (see e.g. \cite{osband2017deep}). Essentially, one can imagine that the state space factorizes into $H+1$ components, thought of as stages of the decision problem. For any policy, a state $s \in \Sc_i$ transitions to a state in $\Sc_{i+1}$ until stage $H+1$ is reached and the interaction effectively ends. We also assume the policy class factors into separate components. This structure allows us to change the policy in stage $h$ without influencing the policy at other stages, essentially encoding time-inhomogeneous policies.  
\begin{restatable}[]{cond}{condFiniteHorizon}
	\label{cond: finite_horizon}
	Suppose the state space factors as $\Sc=\Sc_1 \cup \cdots \cup \Sc_{H} \cup \Sc_{H+1}$, where for a state $s\in \Sc_h$ with $h\leq H$, $P( \Sc_{h+1}| s, a)=1$ for all $a\in \Ac_s$. The final subset $\Sc_{H+1}=\{\tau\}$ contains a single costless absorbing state, with $P(\{\tau\} | \tau,a)=1$ and $g(\tau,a)=0$ for any action $a$. The parameter space is the product set $\Theta=\Theta_{1} \times \cdots \times \Theta_H$, where  a policy parameter $\theta=(\theta_{1}, \ldots, \theta_{H})\in \Theta$ is the concatenation of $H$ sub-vectors. For any fixed $s\in \Sc_h$, $\pi_{\theta}(s)$ depends only on $\theta_{h}$. 
\end{restatable}
We now state the main result for this subsection which applies under conditions much weaker than those for Theorem \ref{thm: no bad stationary points general}. First, we only require the policy class to contain the optimal policy. Second, we relax Condition \ref{cond: stationary}, which considered stationary points of the single-period Bellman objective $\theta \mapsto \Bc(\theta|\eta_{\pi},J_\pi)$, induced by any policy $\pi \in \Pi_{\Theta}$. Instead, we only need to impose a regularity condition on the Bellman objective corresponding to the optimal cost-to-go function, $\Bc(\theta|\eta,J^*)$. This is helpful as the cost-to-go function of an optimal policy often satisfies nice regularity properties (e.g. convexity) than that of an arbitrary policy. \jb{Example \ref{ex:inventorycontrol} illustrates how we exploit both these properties for finite horizon inventory control with the class of base-stock policies.}
\begin{cond}
	\label{cond:finite_horizon_stationarity}		
	For any $\eta \in \{ \eta_{ \pi} : \pi \in \Pi_{\Theta} \}$, the problem $\min_{\theta \in \Theta} \Bc(\theta|\eta,J^*)$ has no suboptimal stationary points.
\end{cond}
However, for our argument to work only with Condition \ref{cond:finite_horizon_stationarity}, we do require a stronger regularity property of the initial distribution $\rho$ as compared to Assumption \ref{ass: exploratory rho}. 
\begin{assumption}\label{ass: exploratory rho finite horizon}  
For any $\pi \in \Pi_{\Theta}$, $\eta_{ \pi}$ is absolutely continuous with respect to $\rho$.  
\end{assumption}
Assumption \ref{ass: exploratory rho finite horizon} is crucial for our proof as it enables us to relate stationary points of $\Bc(\theta|\eta,J_\pi)$ to those of $\Bc(\theta|\eta,J^*)$. See Appendix \ref{subsec: finite horizon proof} for details. Also note that in context of finite horizon problems, Assumption \ref{ass: exploratory rho finite horizon} implies that the agent may begin in a sub-problem with fewer than $H$ periods remaining. This would typically be possible in simulation based optimization and it is necessary for the results we derive. Intuitively, it ensures the simulated agent encounters all possible `scenarios' when the policy is being trained. 

\begin{restatable}[]{thm}{finiteHorizon}
	\label{thm: finite_horizon}
	Suppose Conditions \ref{cond: finite_horizon} and \ref{cond:finite_horizon_stationarity} hold. If the parameterized policy class $\Pi_\Theta$ contains an optimal policy, then any stationary point $\theta$ of $\ell:\Theta \to \mathbb{R}$ satisfies $\ell(\pi_\theta) = \ell(\pi^*)$. 
\end{restatable}
\begin{proof}[Proof Sketch]
	The proof is given in Appendix \ref{subsec: finite horizon proof} and proceeds by backward induction. We first show that at any stationary point $\theta$ of $\ell(\cdot)$, $J_{\pi_{\theta}}(s)=J^*(s)$ holds $\rho$-almost surely for $s \in \Sc_H$. We then argue that the same statement holds for $s \in \Sc_h$ for all $h < H$ as well. Our result then follows by invoking Lemma \ref{lem:minimizers_of_ell}.
	
\end{proof}
Note that Theorem \ref{thm: finite_horizon} only gives a characterization of the stationary points of $\ell(\cdot)$. We leave the study of a gradient dominance condition for finite horizon problems as future work. 

\jb{
	\begin{rem}[On comparison with Policy Search by Dynamic Programming]
	 We briefly remark on how Theorem \ref{thm: finite_horizon} compares to results of \cite{bagnell2004policy} on the Policy Search by Dynamic Programming (PSDP) algorithm. Note that the spirit of our work and PSDP is quite different. While we focus on the optimization landscape of policy gradient methods, PSDP is a dynamic programming based approach which \textit{approximately} solves the policy iteration problem, $\min_{ \pi \in \Pi_{\Theta}} \Bc( \pi| \mu, J_{\pi})$ over some base distribution $\mu$. It does this recursively, starting from the end of the horizon. If $\mu$ is a good guess of the state occupancy measure of the optimal policy $\eta_{\pi^*}$, and the policy class contains a near-optimal policy,  PSDP returns an approximately optimal policy. Therefore, PSDP requires the base distribution $\mu$ to be equal to the state-occupancy measure of the optimal policy, i.e. $d(\eta_{\pi^*},\mu)=0$ for an appropriate metric $d(\cdot, \cdot)$ between two distributions. This is a much stronger condition than Assumption \ref{ass: exploratory rho finite horizon}, which only requires $\eta_{\pi} << \rho$ for all $\pi \in \Pi_{\Theta}$ and can be ensured for example by letting $\rho$ to be a density supported on the state space in each period, $\Sc_1 \times \Sc_2 \times \ldots \times \Sc_{H}$.
	\end{rem}
}

%% file: examples.tex
\section{Examples }\label{sec:examples}
 We now show how our general results in Section \ref{sec:stationary_points} apply to several examples.

\subsubsection*{Finite state and action MDPs with natural parameterization.}

\begin{example}[Finite state-action MDPs]
	\label{ex:tabularMDPs}
	Consider a problem with finite state space $\Sc=\{1,\cdots, n\}$. For simplicity, we assume the set of feasible actions $\Ac_s$ is the same for every state $s$ and denote this by $\Ac$. We also assume there is a finite set of $k$ deterministic actions to choose from and take $\Ac=\Delta^{k-1}$ to be the set of all probability distributions over these actions. That is, any action $a \in \Ac$ is a probability vector where each component $a_i$ denotes the probability of taking the $i$-th deterministic action. Cost and transition functions can be naturally extended to functions on the probability simplex by defining:
	\begin{align}\label{eq: tabular costs and transitions}
	g(s,a) = \sum_{i=1}^{k} g(s,e_i) \, a_i && P(s'| s,a)=\sum_{i=1}^{k} P(s'| s, e_i) \, a_i,
	\end{align}
	where $e_i$ is the $i$-th standard basis vector, representing one of the $k$ possible deterministic actions. 
	
	For this tabular setting, a natural parameterization considers the policy $\pi_{\theta}(s) = \theta_s \in \Delta^{k-1}$ which associates each state with a probability distribution over actions. Rather than track the policy parameter $\theta=(\theta_s : s=1,\cdots, n)\in \mathbb{R}^{n\times k}$ we work directly with a stochastic policy $\pi\in \mathbb{R}^{n \times k}$, viewed as a matrix whose rows are probability vectors. In this case, the set of all stationary randomized policies can be written as $\Pi  = \{ \pi \in \mathbb{R}^{n\times k}_{+}  :  \sum_{i=1}^{k} \pi_{s,i}  =1  \,\, \forall s\in \{1,\cdots, n\}  \}$. (When taking gradients, it can be helpful to view $\pi$ as a stacked vector rather than a matrix.)
\end{example}
 Since $\Pi$ contains all stationary policies, it is clearly closed under policy improvement. It is also worth noting that for any $\pi \in \Pi, s \in \Sc$ and $a \in \Delta^{k-1}$, the Q-function is linear in $a$, as we can write: $Q_{\pi}(s,a) = \sum_{i=1}^k Q_{\pi}(s,e_i) a_i = \langle Q_{\pi}(s,\cdot), a \rangle$. Therefore, the weighted policy iteration objective, 
\begin{equation}
	\label{eq:finiteMDP_weightedPI}
\Bc(\pi'|\eta_{\pi}, J_{\pi}) = \sum_{s \in \Sc} \eta_{\pi}(s) \sum_{i=1}^{k}  Q_{\pi}\left(s,e_i\right) \pi'_{s,i}	 
\end{equation}
is convex (linear) in $\pi'$. This implies that the weighted Bellman objective has no suboptimal stationary points (Condition \ref{cond: stationary}) and in fact is \jb{(1,0)-gradient dominated} (Condition \ref{cond: convexity_strong}). Condition \ref{cond: diff} is not hard to verify\footnote{We sketch the argument. It is clear that \eqref{eq:finiteMDP_weightedPI} is continuously differentiable in $\pi'$, since it is linear in $\pi'$. On the other hand, one can show to show that $\Bc(\pi | \eta_{\pi'}, J_{\pi})$ is continuously differentiable in $\pi'$ by observing that it is linear in $\eta_{\pi'}$ and that $\eta_{\pi'}$ can be written in matrix form as $\eta_{\pi'} = (1-\gamma) \rho (I-\gamma P{\pi'})^{-1}$ where $\rho$ is viewed as a row vector and $P_{\pi'}$ is the transition matrix of the Markov chain induced by $\pi'$. Here $I-\gamma P_{\pi'}$ is invertible and linear in $\pi'$, so the derivative of this inverse matrix exists and is continuous. }. Therefore, Theorems \ref{thm: no bad stationary points general} and \ref{thm: convergence_rates} confirm that for tabular MDPs, $\ell(\cdot)$ has no suboptimal stationary points and is gradient dominated with degree one. 


\jb{
	We give an illustrative convergence rate result for this example by appealing to part (1) of Lemma \ref{lemma:second_order_convergence_rate} and computing all the relevant quantities. Notice that application of Lemma \ref{lemma:second_order_convergence_rate} also requires smoothness properties. For this, we use a result in \citep{agarwal2020optimality} which shows that for tabular MDPs with natural parameterization, $\nabla \ell$ is Lipschitz continuous with constant $L=\frac{2\gamma |\Ac|}{(1-\gamma)^2}$. See Appendix \ref{app:tabular MDPs} for a detailed proof.
	\begin{restatable}[Convergence rate for tabular MDPs]{lem}{convergencerateresult}\label{lemma: convergence rate tabular}
		Consider the finite state action MDP with natural parameterization as formulated in Example 3. Assume $\gamma \geq 1/3$ and per-period  costs are normalized with $\|g\|_{\infty}\leq 1$. For projected gradient descent, $\pi^{(t+1)} = {\rm Proj}_{\Pi} (\pi^t - \alpha \nabla \ell(\pi^t)))$ with $\alpha =  \frac{(1-\gamma)^2}{2\gamma |\Ac|}$, 
		\[
		   \ell(\pi^T) - \ell^* \leq \sqrt{ \frac{ 8\gamma |\Sc| |\Ac| \kappa_{\rho}^2}{(1-\gamma)^4} \frac{ (\ell(\pi^0) - \ell^*)}{ T}},
		\]
		where $\ell^*=\inf_{\pi \in \Pi}\ell(\pi)$. 
	\end{restatable}
	\begin{rem}
		Note that the dependence on various factors such as $\gamma, |\Sc|, |\Ac|, \kappa_\rho, T$ is essentially the same as in the result of \citep{agarwal2020optimality} except for a factor of $(1-\gamma)^2$ due to our definition of the policy gradient objective $\ell(\cdot)$ as a discounted average cost.
	\end{rem}
}

\subsubsection*{Regularized finite state and action MDPs with natural parameterization.}
We now consider a modification of example \ref{ex:tabularMDPs} where a strongly convex regularizer is added to the single period cost functions. This smooths the problem and ensures that optimal policies are strictly stochastic. See \cite{geist2019theory} and the references therein for a formulation that is analogous to ours, \jb{except that while they use a negative entropy regularizer, we propose to use a more aggressive log-barrier based regularizer}\footnote{Like \cite{geist2019theory}, we regularize single stage cost functions, which produces a modified MDP with continuous actions to which our results apply. Directly adding a generic regularizer to $\ell(\theta)$ before optimizing over policies might produce objectives that are not additive across time. }.

\begin{example}(Regularized Finite MDPs)\label{ex:regularized_tabular}
	As in Example \ref{ex:tabularMDPs}, let $\Sc = \{1,\cdots, n\}$ and $\Ac=\Delta^{k-1}$. Define $\mathcal{R}(a):= \lambda D_{\rm KL}(U||a) = \lambda \sum_{i=1}^{k} \left(\frac{1}{k}\right)\log\left(  \frac{1/k}{a_i} \right)$ to be proportional to the Kullback–Leibler (KL) divergence (a.k.a relative entropy) between a uniform distribution and $a$. Assume that for each $s$ there exists $g_s\in \mathbb{R}^k_+$ such that  \begin{align}\label{eq: entropy regularized costs and transitions}
		g(s,a) = g_s^\top a + \mathcal{R}(a)  && P(s'| s,a)=\sum_{i=1}^{k} P(s'| s, e_i) \, a_i.
	\end{align}
	\jb{The regularizer $\mathcal{R}(a)$ is a strongly-convex extended-real-valued function with ${\rm dom}(R) = \{a \in \Delta^{k-1} : \mathcal{R}(a)< \infty\}=\{a\in \Delta^{k-1}: \min_{i} a_i >0\}$.}  
\end{example}
\jb{Under this example, the weighted Bellman objective takes the form 
	\begin{equation}\label{eq:weighted_bellman_regularized}
		\Bc(\pi' \mid \eta_{ \pi}, J_{\pi}) = \sum_{s \in \Sc} \eta_{\pi}(s) \sum_{i=1}^{k}  Q_{\pi}\left(s,e_i\right) \pi'_{s,i} +\sum_{s \in \Sc} \eta_{\pi}(s) \mathcal{R}((\pi'_{s,1},\cdots, \pi'_{s,k}) ),
	\end{equation} 
	which is the sum of a linear function and a strongly convex regularizer. Strong convexity of $\Bc$ implies it has no suboptimal stationary points, and in fact it is gradient dominated with degree 2 (Condition \ref{cond: 2ab}). The class of policies is trivially closed under policy improvement (Condition \ref{cond: closure}). Together, these imply $\pi \mapsto \ell(\pi)$ is gradient dominated with degree two.}

	\jb{Notice that relative-entropy regularization is incorporated into the single-period cost functions. Our results imply that appropriate first-order methods applied to a regularized objective converge to the optimum of that objective. We complement this by establishing that, when $\lambda$ is small, an optimal policy for an entropy regularized objective is near-optimal for an un-regularized one. See Appendix \ref{app:regularized} for a proof. \cite{geist2019theory} provides similar bounds albeit with a negative entropy regularizer; these bounds do not apply here as the log-barrier regularizer is unbounded.  
		\begin{restatable}[Impact of regularization]{lem}{regularizationimpact}\label{lem:impact-of-regularization}
		 Let $\ell_{\lambda}(\pi)$ denote the average cost function for the problem described in Example \ref{ex:regularized_tabular} with a given regularization  parameter $\lambda \geq 0$.  Then, if $\pi_{\lambda}^* \in \argmin_{\pi \in \Pi} \ell_{\lambda}(\pi)$,
		\[
		\ell_{0}\left(\pi_{\lambda}^* \right)\leq \min_{\pi \in \Pi} \ell_{\lambda}(\pi)+ \lambda \left(1 + \log\left( 1+\frac{c}{\lambda} \right) \right) \quad \text{ where} \quad c=2(\max_{s,i} |g_{s,i}| ) / (1-\gamma). 
		\]
	\end{restatable}}

\subsubsection*{Regularized finite state and action MDPs with nonlinear parameterization.}
Our next example discusses the popular softmax policy parameterization for tabular MDPs.   
\begin{example}[Softmax policies]\label{ex:regularized softmax}
	\label{ex:vanishin_grads_softmax}
Consider the setting of Example \ref{ex:regularized_tabular} but with a different policy parameterization. 
Suppose policies are parameterized by $\theta \in \mathbb{R}^{n \times (k-1)}$ where for any state $s$, $\pi_{\theta}(s) \in \Delta^{k-1}$ is a probability distribution whose components  $\pi_{\theta}(s)\equiv(\pi_{\theta}(1|s),\cdots, \pi_{\theta}(k|s))$ satisfy
	\[
	\pi_{\theta}(1|s)=\frac{1}{1+\sum_{j=2}^k e^{\theta_{s,j}}} \qquad \& \qquad \pi_{\theta}(i|s)=\frac{e^{\theta_{s,i}}}{1+\sum_{j=2}^k e^{\theta_{s,j}}} \quad  i\neq 1. 
	\]
Here, we have simplified our discussion by tracking $k-1$ parameters per state, denoted $(\theta_{s,2},\cdots, \theta_{s,k})$, and effectively fixing $\theta_{s,1}=0$. This convenient as it implies that each policy in the policy class corresponds to a unique parameter vector.  
\end{example}

\jb{The cost-function $\theta \mapsto \ell(\theta)$ is coercive, since $\ell(\pi_{\theta})\geq \sum_{s} \rho(s) R(\pi_{ \theta}(s))\to \infty$ as any component $\theta_{s,i}\to \infty$. Lemma \ref{lem: pgd reaches global optimum} therefore implies that gradient descent converges to a stationary point of $\theta \mapsto \ell(\pi_{ \theta})$. Our goal is to show that any stationary point corresponds to an optimal policy.} 
	

\jb{First, note that with this relative entropy regularizer, the function $\pi'\mapsto \Bc(\pi' \mid \eta_{ \pi}, J_{\pi})$ which takes the form in \eqref{eq:weighted_bellman_regularized}, is an extended-real-valued strongly-convex function \citep{boyd2004convex}. Its domain ${\rm dom}(B)= \{\pi' \in \Pi : \Bc(\pi' | \eta_{ \pi}, J_{\pi})<\infty\}$ is the set of strictly stochastic policies $\Pi^{\circ}=\{\pi' \in \Pi : \min_{s,i} \pi'_{s,i}>0 \}$. For any $\pi$, $\argmin_{\pi'} \Bc(\pi' | \eta_{ \pi}, J_{\pi}) \in \Pi^{\circ}$, so the class of policies $\Pi^{\circ}$ is closed under policy improvement.  Clearly,  $\theta \mapsto \pi_{ \theta}$ is an isomorphism\footnote{That is, the function is one-to-one, meaning no two distinct choices of parameter yield the same policy, and onto, meaning that every strictly stochastic policy can be represented by some choice of $\theta$. } between $\mathbb{R}^{n\cdot(k-1)}$ and the set of strictly stochastic policies,  $\Pi^{\circ}$.  Closure under policy improvement (Condition \ref{cond: closure}) follows from this.}  

\jb{Next, we show Condition \ref{cond: stationary}, that $\theta \mapsto \Bc(\pi_{ \theta}| \eta_{\pi}, J_{\pi})$ has no suboptimal stationary points. The intuition behind our argument is that optimizing the single period objective in  \eqref{eq:weighted_bellman_regularized} locally over $\pi'\in \Pi^{\circ}$ is, in a suitable sense, equivalent to optimizing locally over the parameter of a softmax policy. Since the arbitrary policy $\pi$ is fixed throughout, we use the shorthand notation $\Bc(\pi_{\theta}) \equiv \Bc(\pi_{\theta} \mid \eta_{\pi}, J_{\pi})$. Suppose  $\theta_0$ is a stationary point of $\theta \mapsto \Bc(\pi_{ \theta})$. Then, Corollary \ref{cor:softmax_to_direct_stationarity} shows $\pi_{\theta_0}$ is a stationary point of $\pi' \mapsto \Bc(\pi')$. Since $\Bc(\pi')$ is strictly convex in $\pi'$, it has no suboptimal stationary points. Hence, $\pi_{ \theta_{0}}$ must be a minimizer of $\Bc$.}


\jb{ In the next result, $\partial \pi_{ \theta}/ \partial \theta$ is a Jacobian matrix. For completeness, a proof is given in Appendix \ref{app:softmax}. It acts a local change of coordinates. This can also be critical in practical policy gradient implementations \citep{kakade2002natural,schulman2015trust}. 
	\begin{lem}[Diffeomorphism]\label{lem:softmax_diffeomorphism}
		Given a softmax policy $\pi_{ \theta}$ as in Example \eqref{ex:vanishin_grads_softmax} and any vector $D$ such that $\pi_{ \theta}+\alpha D\in \Pi$ for sufficiently small $\alpha>0$, there exists a vector $N$ such that $\left[\frac{\partial \pi_{ \theta}}{\partial \theta}\right] N=D$.
	\end{lem}
	\begin{cor}[Correspondence of stationary points]\label{cor:softmax_to_direct_stationarity}
		If $\nabla_{\theta} \Bc(\pi_{\theta}) =0$, then $\left\langle\nabla_{\pi_{\theta}}\Bc(\pi_{ \theta})\, ,\, \pi'-\pi_{ \theta} \right\rangle \geq 0 $ for each $\pi' \in \Pi$.  		
	\end{cor}
	\begin{proof}
		Suppose otherwise and there exists $\pi' \in \Pi$ and a descent direction $D=\pi'-\pi_{ \theta}$ with $\left\langle\nabla_{\pi_{\theta}}\Bc(\pi_{ \theta})\, ,\,  D\right\rangle < 0$. By convexity of $\Pi$, $D$ is feasible, i.e. $\pi_{ \theta}+\alpha D = (1-\alpha)\pi_{ \theta}+\alpha \pi'  \in \Pi$ for $\alpha \in (0,1)$. Lemma \ref{lem:softmax_diffeomorphism} therefore applies and we can take $N$ as in Lemma \ref{lem:softmax_diffeomorphism}. Then, by the chain rule 
		\[
		\langle \nabla_{\theta} \Bc(\pi_{ \theta}) \, , \, N\rangle = \left\langle \left[ \frac{\partial \pi_{\theta}}{\partial \theta} \right]^\top \nabla_{\pi_{\theta}} \Bc\left(  \pi_{ \theta} \right)  \, ,\, N \right\rangle = \left\langle \nabla_{\pi_{\theta}} \Bc\left(  \pi_{ \theta} \right)\,, \, \left[ \frac{\partial \pi_{\theta}}{\partial \theta} \right] N \right\rangle = \left\langle \nabla_{\pi_{\theta}} \Bc\left(  \pi_{ \theta} \right)\,, \,  D \right\rangle < 0,
		\]
		yielding a contradiction.  
	\end{proof}
	Therefore, under Conditions \ref{cond: closure} and \ref{cond: stationary}, Theorem \ref{thm: no bad stationary points general} implies that $\theta \mapsto \ell(\pi_{\theta})$ has no suboptimal stationary points. Lemma \ref{lem:impact-of-regularization} shows that solving this relative entropy regularized problem with a very small regularization parameter $\lambda$ will produce a near-optimal policy for the unregularized objective as well. It is worth emphasizing, however, that our results in Theorem \ref{thm: no bad stationary points general} do not apply to the case of softmax policies with unregularized finite MDP (i.e if we take $\lambda=0$). In that case the cost function $\theta \mapsto \ell(\pi_{\theta})$ has no minimizer, since an optimal policy is not strictly stochastic. Similarly, it has no stationary points. A minimum can be approached as certain components of $\theta$ tend to infinity, but  is never attained. The choice of policy parameterization leads to a degenerate optimization problem, even in a seemingly trivial problem with just a single state and single time period $(\gamma=0)$. 
}

\subsubsection*{Linear MDPs}

\jb{
Since their introduction in \cite{jin2020provably}, the class of linear MDPs has been widely used in theoretical analysis of reinforcement learning. This model assumes instantaneous cost functions and transitions kernels have a low-dimensional structure that ensures cost-to-go functions lie in a low dimensional function class. The next example treats a regularized linear MDP, where a small entropy regularizer is added to the single-period costs. This choice smooths the problem, making it easy to apply and analyze policy gradient methods.}

\begin{example}[\jb{Entropy regularized linear MDPs}]\label{ex:linear_mdp}
	\jb{As in Examples \ref{ex:tabularMDPs}, \ref{ex:regularized_tabular}, let $\Ac=\Delta^{k-1}$ be the set probability distributions over $k$ deterministic actions. The standard basis vectors $e_1, \cdots, e_k$ correspond to deterministic actions. Suppose there exists a feature map $\phi:\Sc \times \Ac \to \mathbb{R}^{m+1}$, a parameter $\beta \in \mathbb{R}^{m+1}$, and signed measures $\mu_0, \cdots, \mu_m$ over $\Sc$ such that for each state $s \in \Sc$,  action distribution $a\in \Ac$ and subset $\Mc\subset \Sc$,
	\[
	g(s,a) = \sum_{j=0}^{m}\phi_{j}(s,a) \beta_j \quad \text{and} \quad P( \Mc |s,a) = \sum_{j=0}^{m} \phi_{j}(s,a) \intop_{\Mc} \mu_{i}(ds').  
	\]
	In keeping with the interpretation of an action $a$ as encoding a distribution over a set of $k$ deterministic actions,  we assume $\phi_{j}(s,a)= \sum_{i=1}^{k} \phi_{j}(s,e_i) a_i$ for $j\in \{1,\cdots, m\}$.} 
	
	\jb{To smooth the problem, we include a pair of basis elements $(\phi_{0}, \mu_{0})$ satisfying:  
	$$\phi_{0}(s,a) = \mathcal{R}(a):= \lambda D_{\rm KL}(U ||a)  \quad \text{and} \quad   \mu_{0}(\Mc)=0  \quad \forall s\in \Sc, a\in \Ac, \Mc \subset \Sc.$$ 
	Take $\beta_0=1$. As in Example \ref{ex:regularized_tabular}, one should interpret this as modifying the original objective of the MDP by adding  the relative-entropy term  $\mathcal{R}(a)$  to the single period cost function. This is useful for us because it ensures that the optimal policy is a smooth function of problem parameters (like $\beta$). Lemma \ref{lem:impact-of-regularization} shows that when $\lambda$ is very small, a policy that is optimal with respect to this modified objective will be near optimal for the original objective.}
	
	\jb{The inherent low-dimensional structure of linear MDPs ensures that each state action value function lies in a low dimensional class of functions spanned by the basis vectors $\phi_0, \cdots, \phi_d$. In particular, 
	\begin{equation}\label{eq:linear-mdp-q-pi} 
		Q_{\pi}(s,a) = \sum_{j=0}^{m} \phi_{j}(s,a) \beta_j +  \gamma \sum_{j=0}^{m} \phi_{j}(s,a) \intop_{\Sc}  J_{\pi}(s')\mu_{j}(ds') = \mathcal{R}(a) + \sum_{j=1}^{m} \phi_{j}(s, a) \theta_j
	\end{equation}
	where we choose $\theta_j = \beta_j + \gamma \intop_{\Sc}  J_{\pi_{\theta}}(s')\mu_{j}(ds')$.}
\end{example}

\jb{Motivated by \eqref{eq:linear-mdp-q-pi}, define the low dimensional class of state-action value function of the form 
\[
Q^{\theta}(s,a)= a^\top \Phi(s) \theta + R(a).
\]
where $\theta \in \mathbb{R}^m$ and  $\Phi(s) \in \mathbb{R}^{k\times m}$ has elements $\Phi(s)_{ij}= \phi_{j}(s,e_i)$. Take $\pi_{\theta}(s)= \argmin_{a\in \Delta^{k-1}} Q^{\theta}(s,a)$ to be the policy induced by minimizing such a state-action value function. 
\begin{lem}\label{lem:linear_mdps}
	The class of policies $\Pi_{\Theta} = \{\pi_{\theta} : \theta \in \mathbb{R}^m \}$ satisfies Condition \ref{cond: closure} (closure) and for any $\pi \in \Pi$, $\theta \mapsto \Bc\left(\pi_{\theta} \mid \eta_{ \pi}, J_{\pi} \right)$ satisfies Condition \ref{cond: stationary} (no sub-optimal stationary points).
\end{lem}
\begin{proof}[Proof sketch]
	To establish the closure condition, fix an arbitrary policy $\pi$. By \eqref{eq:linear-mdp-q-pi}, one can pick $\theta^+$ such that $Q_{\pi}=Q^{\theta^+}$. Then, by construction, $\pi_{\theta^+}(s)$ is a policy iteration update to $\pi$.   Now, we want to show that the single period objective $\theta \mapsto \Bc(\pi_{\theta} \mid \eta_{\pi}, J_{\pi} )$ does not have any suboptimal stationary points. Given $\theta$,  take $\theta^{\alpha} = \theta + \alpha(\theta^+ - \theta)$. We show in Appendix \ref{subsec:app_linear_mdp_proof} that   
	\[
	\frac{d}{d\alpha}  \Bc(\pi_{\theta^{\alpha}} \mid \eta_{\pi}, J_{\pi}) = \intop   \frac{d}{d\alpha} Q_{\pi}(s, \pi_{\theta^{\alpha}}(s))  \eta_{\pi}(ds) = \intop   \frac{d}{d\alpha} Q^{\theta^+}(s, \pi_{\theta^{\alpha}}(s))  \eta_{\pi}(ds) < 0. 
	\]
	To see this intuitively, consider some fixed $s$. The policy iteration update to $\pi$, is a greedy policy defined by optimizing $Q_{\pi}=Q^{\theta^+}$. Precisely, $\pi_{\theta^+}(s) \in \argmin_{a\in \Ac} Q^{\theta^+}(s,a)$ minimizes an entropy regularized linear cost function over the probability simplex. The current (sub-optimal) policy $\pi_{\theta}$ optimizes a different (incorrect) cost function. It is a greedy policy to $Q^{\theta}$, with $\pi_{\theta}(s) \in \argmin_{a\in \Ac} Q^{\theta}(s,a)$. Changing the optimization objective marginally, from $Q^{\theta}(s, \cdot)$ towards the (correct) cost function $Q^{\theta^+}(s,\cdot)$, reduces the cost of the selected action.
\end{proof}}

\jb{
\begin{rem}
	Unlike other examples, here a parameterized policy is defined implicitly as the solution of a certain optimization problem. One way to derive gradients of this policy is to use implicit differentiation, as in our proof in Appendix \ref{subsec:app_linear_mdp_proof}.
\end{rem}
}

\subsubsection*{Linear quadratic control with linear policies.}
We briefly revisit the linear quadratic control example in Section \ref{sec:Motivation_LQ}. As discussed there, this example technically falls outside the scope of our formulation because single-period costs are not uniformly bounded. At the same time, \emph{the proofs} of Theorems \ref{thm: no bad stationary points general} and \ref{thm: convergence_rates} apply essentially without modification to the class of stable policies. In particular, by Corollary \ref{cor: convexity to gradient dominance of pg}, because the policy class is closed and the single period weighted PI objective is quadratic, the policy gradient cost function is gradient dominated with degree two. This may help illuminate the gradient dominance result previously proved directly by \cite{fazel2018global}. \cite{fazel2018global} also give a detailed analysis of the rate of convergence attained by several specific algorithms. Despite gradient dominance, a subtlety that complicates such convergence analysis is that  the  smoothness properties of the loss function in Lemma \ref{lem: lqr loss function} only hold over sublevel sets. Smoothness over sublevel sets is sufficient to apply Lemma \ref{lem: pgd reaches global optimum}, implying convergence of gradient descent to the global optimum. But as currently stated,  the convergence rate in Lemma \ref{lemma:second_order_convergence_rate} does not apply since that lemma assumes a global bound on the norm of the Hessian. We believe Lemma \ref{lemma:second_order_convergence_rate} can be extended along the lines of Lemma \ref{lem: pgd reaches global optimum} to give a geometric convergence rate under constant stepsizes chosen to ensure that iterates always remain in sublevel sets. 

\subsubsection*{Optimal stopping with threshold policies.} We now turn to an example with a structured policy class.

\begin{example}[Optimal Stopping]
	\label{ex:stopping}
	We formulate the optimal stopping problem as a reward \emph{maximization} problem\footnote{One could imagine costs to be the negative of reward in order to be consistent with the formulation in Section \ref{section:problem_formulation}.}.  In each round the agent observes a state variable $x_t$ taking values in a finite set $\Xc$, which evolves according to an uncontrolled Markov chain with time-homogeneous transition probabilities $\Prob(x_{t+1}=x'| x_t=x) = p(x'|x)$.  Conditioned on $x_t$, the agent receives an offer $y_t \in \mathbb{R}$ drawn i.i.d from some probability density function $q_{x_t}(\cdot)$. We assume that for each $x\in \Xc$, $q_{x}(\cdot)$ has support $\{ y\in \mathbb{R}  :  q_{x}(y)>0  \}=[y_{\min}, y_{\max}]$ where $0<y_{\min}<y_{\max} < \infty$ and has a continuous derivative throughout its support\footnote{To be rigorous, at the boundaries it has a directional derivative only.}. Set $\Yc=[y_{\min}, y_{\max}]$. If the offer is accepted in round $t$, the process terminates and the agent accrues a reward of $\gamma^t y_t$. Rejecting the offer in any round is costless. The agent's objective is to maximize the expected revenue.  
	
	This problem can be formalized as a Markov decision process with the state-space $\Sc= \Sc_{\rm C}\cup \{\tau \}$, consisting of a set of continuation states $\Sc_{\rm C} = (\Xc \times \Yc)$ and a terminal state $\tau$ that is costless ($g(\tau,a)=0$) and absorbing ($P(\{\tau\} |\tau,a)=1$). For convenience, we assume the initial distribution $\rho$ assigns zero probability of trivial problem instances that start in the terminal state. We also assume that $\rho$ factorizes over continuation states as $\rho(x,dy) = \nu(x) q_x(y)dy$ where $\nu(x) > 0$ for every $x \in \Xc$. 
	The action $a=0$ corresponds to accepting the offer and terminating while action $a=1$ continues the game by transitioning to a new state with probabilities given by
	\[
	\mathbb{P}\left[s_{t+1}=(x',dy') \mid s_t=(x,y),a=1\right] = p(x' | x) q_{x'}(y')dy'.
	\]
	We consider the class of threshold policies where the vector $\theta \in \Theta:=[y_{\min}, y_{\max}]^{|\Xc|}$ specifies one stopping threshold per context. The policy $\pi_{\theta}(x,y) = \mathbbm{1}\left(  y < \theta_x \right)$ rejects all offers below $\theta_x$.
\end{example}
It is easy to show closure of the policy class. For a threshold policy $\pi$, a policy iteration step updates to yet another threshold policy which accepts an offer $y$ in $x$ if and only if it exceeds the continuation value, $c_{\pi}(x) = \gamma\E[J_{\pi}( x_{t+1}, y_{t+1} )  | x_t = x, a_t=1 ]$. See Appendix \ref{subsec:app_optimalstopping} for details where we also show that Conditions \ref{cond: diff}, \ref{cond: closure} and \ref{cond: stationary} hold. It is also possible to prove gradient dominance and smoothness results that, due to Lemma \ref{lemma:second_order_convergence_rate}, imply convergence rates for policy gradient methods. We omit those results for brevity, but the interested reader can find precise statements in the Appendix and the proofs in the technical report \citep{technicalReport}.

\subsubsection*{Finite horizon inventory control with base stock policies.}
We now apply Theorem \ref{thm: finite_horizon} to a finite horizon inventory control problem with the class of multi-period base stock policies. \cite{kunnumkal2008using} previously showed through a somewhat intricate analysis that a stochastic approximation algorithm converges to the optimal policy, despite non-convexity of the objective. See also the follow up work by \cite{huh2013online}. 

One unconventional feature of our formulation is that it involves both time periods $t\in (0,1,\cdots)$ and ``stages'' $h_0, \cdots, h_t, \cdots, H$. This is consistent with the way we formulated finite-horizon problems in Section \ref{subsec:nonstationary_policies}. We have in mind an inventory control problem that is optimized in simulation. Consistent with Assumption \ref{ass: exploratory rho finite horizon},  the initial stage is random, meaning that different simulated scenarios may begin with a different number of periods remaining. This effectively ensures the policy is optimized from a diverse set initial scenarios. This appears to be necessary for general finite horizon MDPs, but an exploratory initial distribution may not be necessary due to the special structure of this particular problem.

\begin{example}[Finite horizon inventory control]
	\label{ex:inventorycontrol}
	Consider a multi-period inventory control problem (also popularly known as the multi-period newsvendor problem) with backlogged demands. In time period $t$, the seller selects a non-negative quantity $a_t\geq 0$ of inventory to order on the basis of the current inventory level $x_t \in \R$. After ordering inventory, a random i.i.d demand $w_t$ is realized and the inventory level evolves as: $x_{t+1} = x_t + a_t - w_t$. We assume the demand distribution has a twice differentiable\footnote{That this PDF is twice differentiable is used to simplify a proof of Condition \ref{cond: diff} given in the appendix. We give a remark there on how the proof could be completed without that condition.} PDF supported over some bounded set $[0,w_{\max}]$. Negative inventory levels correspond to backlogged demand that is filled when additional inventory becomes available. 
	
	The seller begins at some \emph{stage} $h_0$ in the initial period and the stage advances by one in each time period, i.e. $h_{t+1} =h_t +1$, until a final state $H$ is reached. The seller's objective is to minimize total expected cost over the horizon,
	\begin{align}
	\label{eq:inventory_cost_to_go}
	\E \left[ \sum_{t=0}^{H-h_0} \gamma^t (ca_t + b\max\{x_t+a_t-w_t, 0\} + p\max\{-x_t-a_t+w_t, 0\} ) \right]
	\end{align}
	where $c,b,p > 0$ denote the per unit costs of ordering, holding and backlogging items, respectively. We assume that $p > c$. Otherwise, the optimal policy may never order in any period. It is well known that a (multi-period) base-stock policy is optimal for this setting \citep{bertsekas1995dynamic}.
	
	 Let $s_t= (x_t, h_t)$ denote the state at time $t$, encoding all information needed to make an optimal ordering decision. A base-stock policy depends on a vector of target inventory levels $(\theta_1, \cdots, \theta_{H}) \in \mathbb{R}^H_+$ \  At state $(x,h)$, the policy orders inventory $\pi_{\theta}((x, h)) =  \max\{0, \theta_h - x\}$. That is, it orders enough inventory to reach a target level $\theta_h$, whenever feasible. We restrict to the policy class $\Pi_{\Theta}=\{ \pi_{\theta} : \theta \in \Theta \}$ with bounded parameter space $\Theta=  [0, 2w_{\max}]^{H}$. With a selection of $\theta_{h_t} = 2w_{\max}$, the seller can ensure $x_{t+1} \geq w_{\max}$, which is large enough to meet all demand with probability one. Inventory levels above this are clearly suboptimal, due to their excessive holding costs. As the demand distribution is bounded and the target inventory level is in $[0,2w_{\max}]$ 	the feasible inventory levels at every period are trivially bounded in $\mathcal{I} = [- w_{\max}, 2w_{\max}]$, as long as the initial inventory level does not fall outside this set. The initial state $(x_0, h_0)$ is drawn from the initial distribution $\rho$. We assume that $h_0$ has a positive probability of taking on any value in $\{1,\cdots, H\}$ and that, conditioned on $h_0$, the distribution of $x_0$ has a twice differentiable PDF supported on $\mathcal{I}$.  
\end{example}
This is clearly an example of a finite horizon problem with a non-stationary policy class and hence has the structure noted in Condition \ref{cond: finite_horizon}. Condition \ref{cond: diff} follows essentially from using the Leibniz rule along with the fact that base-stock policies are differentiable everywhere except at a single point. Condition \ref{cond:finite_horizon_stationarity} follows from a classic result in inventory control theory which shows that the optimal state-action cost-to-go function $Q^*(s,a)$ is convex in $a$ \citep{bertsekas1995dynamic}. Details are given in Appendix \ref{appendix: example details}.

%% file: initial_distribution.tex

\section{Bounds on the concentrability coefficient}
\label{sec:concentrability}
To show a gradient dominance result, we need to relate the optimality gap to the magnitude of errors in the Bellman equation. In Section \ref{subsec:PG_rates}, we defined the effective concentrability coefficient $\kappa_{\rho}$ for the set of cost-to-go functions $\Jc_{\Theta}=\{J_{\pi_\theta} : \theta \in \Theta \}$ to be the smallest scalar such that 
\begin{align}
	\label{eq:concentrability_measure_restate}
	\|  J  - J^* \|_{1,\rho} \leq \frac{\kappa_{\rho}}{(1-\gamma)} \|  J  - TJ\|_{1, \rho} \quad \forall J \in \Jc_{\Theta}.
\end{align}
To motivate our definition, note that whenever the Bellman operator is a contraction in some norm $\| \cdot \|$ with modulus $\gamma$, the following inequality holds: $\|J-J^*\|\leq (1-\gamma)^{-1} \| J - TJ\|$ (see \eqref{eq: simple near solutions to Bellmans equation} in Appendix \ref{appendix B}). For bounded cost problems, $T$ is a contraction in the maximum norm and can sometimes also be shown to be contractive in a weighted norm which reflects state relevance. See \cite{bertsekas1995dynamic} or \cite{puterman2014markov}. Essentially, the constant $\kappa_{\rho}$ enables the above inequality in the weighted norm $\|\cdot\|_{1,\rho}$, in which $T$ is typically not contractive. 

The focus on this norm is motivated by two factors. First, the optimality gap can be written as $\ell(\pi_\theta) - \min_{\pi \in \Pi} \ell(\pi) = (1-\gamma) \| J_{\pi_{\theta}} - J^* \|_{1,\rho}$, mirroring the left hand side of \eqref{eq:concentrability_measure_restate} modulo a constant factor. Second, the policy gradient theorem in Lemma \ref{lem: PG theorem} reveals a natural dependence on the errors in Bellman's equation weighted under the state occupancy measure $\eta_{\pi}$. As $\eta_{\pi}\succeq (1-\gamma) \rho$, it makes sense to measure the Bellman errors in $\|\cdot\|_{1,\rho}$. It is worth noting that, because our definition of $\kappa_{\rho}$ depends only on the subclass of cost-to-go functions $\Jc_{\Theta}$, it allows for stronger bounds when these functions obey certain regularity properties. We now provide several useful upper bounds on $\kappa_{\rho}$. See Appendix \ref{subsec: app_concentrability} for proof details.

\begin{restatable}[]{thm}{concentrabilityGeneral}\label{thm: concentrability general} The following results apply under the general problem formulation in Section \ref{section:problem_formulation}. 
	\begin{enumerate}[label=(\alph*)]
		\item If $\Sc$ is finite, then $\kappa_{\rho} \leq 1/ \left(\min_{s \in \Sc} \rho(s) \right)$. 
		\item 	Let $\pi^*$ denote any optimal stationary policy. Then, 
		$\kappa_{\rho} \leq \left\|\frac{d\eta_{\pi^*}}{d\rho} \right\|_{\infty}.$
		\item The bound $\kappa_{\rho} \leq C/c$  holds if $T$ is a contraction with modulus $\gamma$ in a norm $\|\cdot\|$ that satisfies 
		\begin{equation}
			\label{eq: norm equivalence}
			c\| J \| \leq \| J\|_{1,\rho} \leq C \|J \|  \qquad \forall J \in \Jc_{\Theta}.
		\end{equation}
	\end{enumerate}
\end{restatable}
The bound in part (a) is simple and can be derived as a special case of the result in part (b). The bound in part (b) depends on the worst-case likelihood ratio between the state occupancy measure under the optimal policy and the initial distribution. Note, $\frac{d\eta_{\pi^*}}{d\rho}$ is the Radon-Nikodym derivative term, which exists because of Assumption \ref{ass: exploratory rho}. Put differently, this result implies that $\kappa_{\rho} \leq C$ if $\eta_{ \pi^*}(\Mc) \leq C\rho(\Mc)$ for every measurable set $\Mc \subset \Sc$. This is, essentially, a restatement of a key observation in \cite{kakade2002approximately}. Such distributional mismatch terms also appears in the works of  \cite{scherrer2014local, agarwal2020optimality}.  

The result in part (c) gives an alternative approach to bounding $\kappa_{\rho}$. It is potentially useful for many problems where the Bellman operator is a contraction with respect to a certain weighed norm, as it suggests $\rho$ should be chosen in a manner which aligns with that norm's state weighting. The optimal stopping problem is one such special case where it can be shown that $T$ is a contraction in $\|\cdot\|_{1, \mu}$, where $\mu$ is the stationary distribution of the underlying Markov chain -- assuming it is never interrupted by stopping. Choosing $\rho=\mu$ implies $\kappa_{\rho} \leq 1$ using \eqref{eq: norm equivalence}. In practical problems, one could easily sample initial states from $\rho$ by simulating this Markov process.
\begin{restatable}[]{lem}{concentrabilityStopping}\label{lemma: concentrability Stopping}
	For the optimal stopping problem in Example \ref{ex:stopping}, consider a policy $\pi_{\rm C}$ that never stops, i.e. $\pi_{\rm C}(s)=1$ for all $s \in \Sc_{\rm C}$. Let $\mu$ be a stationary distribution of the induced Markov process, meaning $\mu(\Mc) = \intop P( \Mc | s', 1)\mu(ds')$ for any $\Mc \subset \Sc$. Then, choosing $\rho=\mu$ implies $\kappa_{\rho} \leq 1$.
\end{restatable}
The LQ control problem technically falls outside the scope of our general formulation as per-stage costs are not bounded. Therefore, we restrict our attention to the cost-to-go functions corresponding to stable linear policies. For this result, we are able to leverage certain regularity properties of this which imply better bounds than the generic bound implied by part (b) of Theorem \ref{thm: concentrability general}. In particular, because the class of cost-to-go functions induced by linear policies are quadratic, we need only the initial distribution to explore the basis of the state space sufficiently, rather than requiring it to almost perfectly mimic the steady state distribution of the (unknown) optimal policy.
\begin{restatable}[]{lem}{concentrabilityLQ}\label{lemma: concentrability LQ}
	Consider the LQ control problem formulated in Example \ref{ex: lq control}. Let  $\theta^*\in \mathbb{R}^{n\times k}$ denote the parameter of an optimal policy and define $\Sigma_{\rho}:= \E_{\rho} \left[ s_0 s_0^\top \right]$ and $\Sigma_{\eta_{ \pi_{\theta^*}}} := \E_{\eta_{\pi_{ \theta^*}}} \left[ s_0 s_0^\top \right]$. Then, 
		\[
		\|  J  - J^* \|_{1,\rho} \leq \frac{\kappa}{(1-\gamma)} \|  J  - TJ\|_{1, \rho}  \qquad \forall J \in  \{ J_{\pi_{\theta}}: \theta \in \Theta_{\rm S} \}  
		\]
		when
		\begin{equation}\label{eq: lqr conc bound}
			\kappa =  \frac{\lambda_{\max}\left( \Sigma_{\eta_{ \pi_{\theta^*}}} \right) }{ \lambda_{\min}\left( \Sigma_\rho \right) }.
		\end{equation}
\end{restatable}
In the claim above, $\Sigma_\rho$ is the second moment matrix of the initial state distribution, which we assumed to be non-singular, and $\Sigma_{\eta_{ \pi_{\theta^*}}}$ is the second moment matrix of the state-occupancy measure under the optimal policy. That is, $\Sigma_{\eta_{ \pi_{\theta^*}}} = (1-\gamma) \sum_{t=0}^{\infty} \gamma^t \E[s_t s_t^\top]$ where $s_t = [A+B\theta^*]^t s_0$ evolves according to linear dynamics from the initial state $s_0\sim \rho$. Since the optimal policy is stable, this infinite sum converges to a finite limit. Generally, this term is large when the system is only barely stable even under the optimal policy, meaning the spectral radius of $\sqrt{\gamma}(A+B \theta^*)$ is close to one.

%% file: approx_closure.tex
\section{Closure under approximate policy improvement.}
\label{sec:approximate_closure}
So far, our results crucially depend on the closure property of the policy class, which applies to many classical dynamic programming problems with structured policy classes. A natural question to ask is whether we can relax this closure condition. In this section, we present results for the case where our policy class is only \textit{approximately} closed under improvement. One would expect expressive policy classes such as those parameterized by a deep neural network, a Kernel method \citep{rajeswaran2017towards}, or using state aggregation \cite{singh1995reinforcement,ferns2004metrics,bertsekas2019feature} to follow this condition. \jb{We discuss the example of state-aggregation at the end of this section.} Recall that $\Pi$ denotes the class of all stationary policies and $\Pi_{\Theta}$ denotes the parameterized policy class over which we search.
\begin{cond}[Closure under approximate policy improvement]\label{cond:approx_closure}
	There exists $\epsilon\geq 0$ such that for every $\pi \in \Pi_\Theta$, 
	\begin{equation}\label{eq:approx_closure}
	\min_{\pi^+\in \Pi_{\Theta}} \Bc(\pi^+ | \eta_{\pi}, J_{\pi}) \leq \min_{ \pi' \in \Pi} \Bc(\pi' | \eta_{\pi}, J_{\pi}) + \epsilon.
	\end{equation}
\end{cond}
\jb{If $\Pi_\Theta$ were closed under policy improvement steps, the approximation error would be zero since there would exist a $\pi^+ \in \Pi_{\Theta}$ such that  $T_{\pi^+}J_{\pi}(s) = TJ_{\pi}(s)$ almost surely for $s$ drawn from $\eta_{\pi}$.}
Condition \ref{cond:approx_closure} measures the deviation from this ideal case, in a norm that weights states by the discounted state occupancy measure under the current policy. We refer to $\epsilon$ as the \textit{inherent Bellman error} of the policy class. We discuss this condition further after the next theorem.

We state our formal result in Theorem \ref{thm: approximate closure} below to show that for approximately closed policy classes, any stationary point of the policy gradient objective is nearly optimal, where this optimality gap is a function of the inherent Bellman error in Condition \ref{cond:approx_closure}. Our result is reminiscent of results in the study of approximate policy iteration methods, pioneered by \cite{bertsekas1996neuro, munos2003error, antos2008learning, munos2008finite, bertsekas2011approximate}, among others. The primary differences are that (1) we directly consider an approximate policy class whereas that line of work considers the error in parametric approximations to the $Q$-function and (2) we make a specific link with the stationary points of a policy gradient method. Recall the definition of $\kappa_\rho$ in \eqref{def:concentrability_measure}, which relates the optimality gap to errors in the Bellman equation weighted under $\rho$.
\begin{thm}\label{thm: approximate closure}
	Suppose Conditions \ref{cond: diff}, \ref{cond: stationary} and \ref{cond:approx_closure} hold. Then, $\ell$ is continuously differentiable and any stationary point  $\theta$  of $\ell(\cdot)$ satisfies, 
	\[ 
	\ell(\pi_\theta) - \ell(\pi^*) \leq  \frac{\kappa_\rho}{(1-\gamma)} \cdot \epsilon 
	\]
\end{thm}
\begin{proof}
	Suppose $\theta$ is a stationary point of $\ell:\Theta \to \mathbb{R}$. Under Conditions \ref{cond: stationary} and \ref{cond:approx_closure}, we have
	\begin{align*}
		\epsilon \geq \min_{\pi^+\in \Pi_{\Theta}} \Bc(\pi^+ | \eta_{\pi_{\theta}}, J_{\pi_{\theta}}) - \min_{ \pi' \in \Pi} \Bc(\pi' | \eta_{\pi_{\theta}}, J_{\pi_{\theta}}) &= \min_{\pi^+\in \Pi_{\Theta}} \left( \intop \left[ T_{\pi^+}J_{\pi_{\theta}} \right] d\eta_{\pi_{\theta}} \right) - \intop \left[TJ_{\pi_{\theta}}\right] d\eta_{\pi_{\theta}} \\
		&= \intop \left[ J_{\pi_{\theta}} - TJ_{\pi_{\theta}} \right] d\eta_{\pi_{\theta}} \\
		&= \| J_{\pi_\theta} - TJ_{\pi_\theta} \|_{1, \eta_{\pi_\theta}},
	\end{align*}
	where the second equality follows from Lemma \ref{lem: bellman for stationary points} and the final equality uses that $J_{\pi_{\theta}} \succeq TJ_{\pi_\theta}$ for any $\pi_{\theta} \in \Pi_{\Theta}$. Then, we have
	\begin{align*}
		\ell(\pi_{\theta}) - \min_{\pi}\ell(\pi) = (1-\gamma) \intop \left[ J_{\pi_{\theta}} - J^* \right] d\rho &= (1-\gamma)\| J_{\pi_{\theta}} - J^{*}\|_{1,\rho} \\
		&\leq \kappa_\rho \|  J_{\pi_{\theta}} - TJ_{\pi_{\theta}} \|_{1, \rho} \\
		&\leq \frac{\kappa_\rho}{(1-\gamma)} \|  J_{\pi_{\theta}} - TJ_{\pi_{\theta}} \|_{1, \eta_{\pi_{\theta}}} \\
		&\leq \frac{\kappa_\rho \cdot \epsilon}{(1-\gamma)}.
	\end{align*}
	The first inequality follows from the definition of $\kappa_\rho$ while the second uses that $\eta_{\pi_{\theta}} \succeq (1-\gamma) \rho$.
\end{proof}

\jb{Intuitively, it seems we might expect the inherent Bellman error in condition \eqref{cond:approx_closure} to be small if the policy class is sufficiently rich. Indeed, it holds with $\epsilon=0$ when $\Pi_\Theta$ contains all possible policies. For the linear MDP problem in Example \ref{ex:linear_mdp}, the condition also holds with $\epsilon=0$ once the policy class contains the low-dimensional family referenced in Lemma \eqref{lem:linear_mdps}. The situation is not so simple, however. For a given policy $\pi$, it is easier to satisfy the policy improvement property \eqref{eq:approx_closure} when $\Pi_{\Theta}$ contains more policies, but if $\Pi_{\Theta}$ is larger we require \eqref{eq:approx_closure} to hold for more choices of $\pi$. For this reason, the inherent Bellman error does not necessarily reduce monotonically as the policy class grows.} 

\jb{Let us explore the illustrative example of state aggregation, which has a long history of being employed in reinforcement learning \citep{tsitsiklis1996feature, van2006performance}. Consider a continuous state, finite action MDP. When the cost functions and transitions dynamics are appropriately smooth, we expect there is little benefit to distinguishing between states that are very close together. A state aggregated policy is one that partitions the state space into disjoint regions and assigns an action distribution to each region. Now consider an increasing sequence of policy classes generated by taking progressively finer partitions of the state space. Does condition \eqref{cond:approx_closure} hold once the partition is sufficiently fine? The next lemma provides a coarse upper bound confirming that inherent Bellman error is small once the policy class is rich enough. To see this, imagine a sequence of state aggregation rules where the distance $\|s-\phi(s)\|$ between a state and its representer tends to zero uniformly. If $g(s,a)$ and $P(\cdot | s, a)$  are Lipshitz continuous in $s$ (in total-variation metric for $P$), then \eqref{eq:state_agg_model_error} tends to zero.} 
\jb{\begin{lem}(Approximate closure of state-aggregated policies)
	Suppose $\Sc = \cup_{i=1}^{m} \Sc_i$ is the union of $m$ disjoint subsets, $\Ac=\Delta^{k-1}$ is the set of probability distributions over $k$ elements, and $\|g \|_{\infty}\leq 1$. From each $\Sc_i$, pick a representer $\bar{s}_i \in \Sc_i$ and define $\phi: \Sc \to \Sc_{i}$ such that $\phi(s) = \bar{s}_i$ if and only if $s\in \Sc_i$. For any $\theta = (\theta_1, \cdots, \theta_m) \in \Theta := \Delta^{k-1} \times \cdots \times \Delta^{k-1}$, define $\pi_{ \theta}(s) = \theta_{\phi(s)}$.  The class of policies $\Pi_{\Theta}$ satisfies Condition \ref{cond:approx_closure} with 
	\begin{equation}\label{eq:state_agg_model_error}
	 \epsilon 
	 = 2 	\sup_{s\in \Sc, a \in \Ac} \Big[| g(s,a)-g(\phi(s),a)| + \frac{\gamma}{1-\gamma} \| P(\cdot \mid s,a) - P(\cdot  \mid \phi(s), a)\|_{\rm TV} \Big].
	\end{equation}
\end{lem}
\begin{proof}
 Fix $\pi \in \Pi_{\Theta}$. Applying
  \eqref{eq:state_agg_model_error} with the definition in \eqref{eq: def q function} ensures  $|Q_{\pi}(s,a) - Q_{\pi}( \phi(s), a)| \leq \epsilon/2$ for all $s\in \Sc$ and $a\in \Ac$. Take $\pi^+$ to a be an unconstrained policy iteration update, with $\pi^+(s) \in \argmin_{a\in \Ac} Q_{\pi}(s,a)$ , and $\pi_{\theta}$ to be a state-aggregated policy which satisfies $\pi_{ \theta}(s) \in \argmin_{a\in \Ac} Q_{\pi}(\phi(s),a)$. Then, 
 \[
 Q_{\pi}(s, \pi_{ \theta}(s))\leq Q_{\pi}( \phi(s), \pi_{ \theta}(s))+\epsilon/2 = \min_{a\in \Ac} Q_{\pi}( \phi(s), a)+\epsilon/2 \leq \min_{a\in \Ac} Q_{\pi}( s, a)+\epsilon = Q_{\pi}( s,\pi^+(s))+\epsilon.
 \]
 Taking an expectation over $s\sim \eta_{ \pi}$ and using the definition of $\Bc\left(\cdot \mid \eta_{ \pi}, J_{\pi}\right)$ establishes that $\Bc(\pi_{ \theta}| \eta_{ \pi}, J_{\pi}) \leq \Bc(\pi^+| \eta_{ \pi}, J_{\pi}) + \epsilon$. 
\end{proof}}

%% file: conclusion.tex

\section{Conclusion}
In this paper, we uncover structural properties of the underlying MDP which guarantee that policy gradient methods converge to globally optimal solutions as well as characterize their convergence rates, even though the optimization objective is non-convex. Our results rely on a key connection with policy iteration, a classic dynamic programming algorithm which solves a single period optimization problem at every step that often has special structure. We show how the policy gradient objective inherits this nice structure, making it amenable for gradient based algorithms to find the optimal policy at a fast rate. There are a number of research directions to extend our work, including results for the function approximation setting with neural network or kernel based parameterization of the cost-to-go function as well as designing principled exploration approaches which can be efficiently combined with policy gradient methods.

%% file: notation.tex
\newpage
\section{Notation}

\begin{table}[htbp]\caption{Table of Notation}
	\label{tab:notation}
	\centering 
	\begin{tabular}{r c p{10cm} }
		\toprule
		$\gamma$ &$\triangleq$& Discount factor. \\
		$\Sc$ &$\triangleq$& State space. \\
		$\Ac_s\subset \mathbb{R}^k$ &$\triangleq$& Convex set of feasible actions when in state $s$. \\
		$\Pi$ &$\triangleq$& Set of all stationary policies. \\
		$\Jc$ &$\triangleq$& Set of bounded measurable functions on $\Sc$. \\
		$g(s,a)$ &$\triangleq$& Single period expected cost of action $a$ in state $s$. \\
		$P(\Mc | s,a)$ &$\triangleq$&  Transition probability to set $\Mc \subset \Sc$ \\
		$g_{\pi}\in \Jc$ &$\triangleq$&  Single period cost function under policy $\pi$. \\
		$J_\pi \in \Jc$  &$\triangleq$&  Cost-to-go function under policy $\pi$. \\
		$Q_{\pi}: \Sc \times \Ac \to \mathbb{R}$ &$\triangleq$&  State-action cost-to-go function under policy $\pi$. \\ 
		$J^* \in \Jc$  &$\triangleq$& Optimal cost-to-go function. \\
		$\pi^*$ &$\triangleq$&  An optimal policy (satisfying $J_{\pi^*}=J^*$). \\
		$Q^*=Q_{\pi^*}$ &$\triangleq$&  State-action cost-to go function associated with an optimal policy. \\
		${T}_{\pi}: \Jc\to \Jc$ & $\triangleq$ & Bellman operator associated with policy $\pi$. \\
		$T:\Jc\to \Jc$ & $\triangleq$ & Bellman optimality operator. \\
		$\rho$ &$\triangleq$& Initial distribution. \\
		$\eta_{\pi}$ &$\triangleq$& The discounted state occupancy measure under policy $\pi$.  See \eqref{eq: discounted_state_occupancy_measure}. \\
		$\ell(\pi)=(1-\gamma)\intop J_\pi d\rho$ &$\triangleq$& Expected discounted cost under a random initial state, policy $\pi$.\\
		$\Theta\subset \mathbb{R}^d$ &$\triangleq$& Convex set of policy parameters. \\
		$\Pi_\Theta= \{\pi_\theta: \theta \in \Theta\}$ &$\triangleq$&  Parameterized policy class. \\
		$\Jc_\Theta= \{J{\pi}: \pi \in \pi_\Theta\}$ &$\triangleq$&  Set of cost-to-go functions under parameterized policies. \\
		$\ell(\theta)=\ell(\pi_\theta)$   &$\triangleq$& Overloaded notation for $\ell(\pi_\theta)$. \\ 
		$\Bc(\pi' | \eta, J_{\pi})$  &$\triangleq$& ``Bellman'' objective or the weighted policy iteration objective \\ 
		$\Bc(\theta | \eta, J_{\pi})=\Bc(\pi_{\theta} | \eta, J_{\pi})$  &$\triangleq$& Overloaded notation for the policy iteration objective at $\pi_{\theta}$. \\ 
		$\kappa_\rho$ &$\triangleq$& Effective concentrability coefficient. See \eqref{def:concentrability_measure}. \\
		$\|J \|_{\infty}$ &$\triangleq$& Max-norm $\sup_{s} |J(s)|$. \\
		 $\|J\|_{1,\eta}$ &$\triangleq$& Weighted 1-norm $\intop |J(s)| \, d\eta$. \\
		 $\| A \|_p$ for $A\in \mathbb{R}^{n\times m}$ &$\triangleq$& Matrix operator norm $\max\{ \|Ax\|_p : \| x\|_p = 1 \}$ \\
		 $\lambda_{\min}(A)$ for $A=A^T\in \mathbb{R}^{n\times n}$ &$\triangleq$& Minimum eigenvalue of $A$\\
		 $\lambda_{\max}(A)$ for $A=A^T\in \mathbb{R}^{n\times n}$ &$\triangleq$& Maximum eigenvalue of $A$\\
		 $f\succeq g$, $f\succ g$ for functions $f,g$&$\triangleq$&  Elementwise inequality, $f(x)\geq g(x)$ or $f(x)>g(x) \, \forall x$ \\
		 $A\succeq B$, $A\succ B$ for $A,B\in \mathbb{R}^{n\times n}$ &$\triangleq$& Indicates $A-B$ is positive (semi)definite. \\
		\bottomrule
	\end{tabular}
\end{table}

%% file: new_app_ag_discussion.tex
\section{Discussion on concurrent work of \cite{agarwal2020optimality}}\label{sec: app_agrawal_discussion}

\jb{Here we provide some comparisons to the approach of \cite{agarwal2020optimality}, offering perspective on the strengths and shortcomings of our approach and highlighting some connections between our technical results and theirs.}

\jb{The work of \cite{agarwal2020optimality} is mostly focused on analyzing specific algorithms, with special attention given to natural-gradient actor-critic algorithms. Explicit bounds on the convergence rates of different algorithms are also given. In comparison, we take a broad algorithm-independent view, focusing primarily on the landscape of the loss function $\ell(\cdot)$ to understand when local policy search is a sensible approach for reaching a (near) optimal policy. Our choice of focus yields several results and insights that are not covered by their work:
	\begin{itemize}
		\item In Example \ref{ex: counterexample}, we provide a simple illustration of how the multi-period nature of the decision problem can lead to bad local minima. With policy closure, we show how challenges raised in that example largely disappear, and the problem is reduced to studying the single period objective function in \eqref{eq:weighted_pi_objective}. This approach of reasoning about the landscape of long-horizon objective by analyzing structure in single period problems offers new insight and greatly simplifies our analysis. 
		\item In addition to treating some examples covered by \cite{agarwal2020optimality}, like tabular and linear MDPs, our analysis seamlessly covers problems with infinite action spaces, structured cost functions, and deterministic policies -- for example linear quadratic control and optimal stopping. 
		\item Our approach extends to finite horizon problems with nonstationary policy classes as shown in Theorem \ref{thm: finite_horizon}. We instantiate this theory for an important practical problem of finite horizon inventory control. 
		\item In stating gradient dominance conditions for convergence rates, our approach to concentration coefficients implies tighter bounds for some examples as compared to the distribution mismatch coefficient in \cite{agarwal2020optimality}. 
	\end{itemize}
	Our focus on an algorithm-agnostic study of the landscape of $\ell(\cdot)$ comes at a cost. This is felt most clearly when studying the softmax parameterization which poses unique optimization challenges even in the seemingly simple single-state single-period problem with a finite number of actions. There is no optimal solution; instead optimal performance is attained only as certain components of the parameter vector tend to infinity. Although our theory applies easily to problems with entropy regularization (see e.g. Example \ref{ex:regularized softmax}), a more specialized analysis is required for the case without regularization. \cite{agarwal2020optimality} do this with a detailed study of specific algorithms and precisely characterize the convergence behavior for both the cases of with and without regularization. In addition, their work also shows how natural policy gradients can be particularly useful in alleviating conditioning issues arising with softmax policies and result in faster convergence.
}

\jb{The next subsection tries to clarify some connections between our theory and that appearing in \cite{agarwal2020optimality}. 
\paragraph{Connecting closure conditions to conditions on  value function approximation}
Compared to our work, a novel theory appears in \cite{agarwal2020optimality} on function approximaiton. Specifically, they analyze a natural gradient actor critic method working with a class of softmax linear policies and compatible function approximation \citep{sutton2000policy, konda2000actor}, i.e. deriving features for approximating the Q-function from score function of a stochastic policy. Although we do not treat this case in detail, we briefly remark on an interesting connection which shows that in one important setting (linearly parameterized Q-functions and the corresponding softmax linear policies), accuracy of value function approximation error implies approximate closure as shown in Condition \ref{cond:approx_closure}. 
}

\jb{This observation follows from a natural duality between a parametric class of policies and a parametric class of value functions. In Example \ref{ex:linear_mdp}, for instance, we leveraged the fact that if a parametric class of value functions $\{Q^{\theta}: \theta \in \Theta\}$ can approximate each $Q_{\pi}$, then the class of greedy policies induced by this class of value functions is closed under policy improvement. A similar observation holds for approximate closure as shown below in Example \ref{ex:linear_softmax}. 
}

\jb{ 
	\begin{example}\label{ex:linear_softmax} As in Example \ref{ex:tabularMDPs}, we assume the set of feasible actions $\Ac_s$ is the same for every state $s$ and denote this by $\Ac$. We also assume there is a finite set of $k$ deterministic actions to choose from and take $\Ac=\Delta^{k-1}$ to be the set of all probability distributions over these actions. Suppose $g(s,a)$ and $P(s' | s,a)$ are linear in $a$, as in \eqref{eq: tabular costs and transitions}. Fix a feature mapping $\phi:\Sc\times \{1,\cdots, k\} \to \mathbb{R}^d$. Consider the class of policies, $\Pi_{\Theta} = \{\pi_{\theta} : \theta \in \mathbb{R}^d\}$ where $\pi_{\theta}(s) = (\pi_{\theta}(s,1), \cdots \pi_{ \theta}(s,k)) \in \Delta^{k-1}$ has components $\pi_{\theta}(s,i) \propto  \exp\{ \theta^\top \phi(s,i) \}$.
	\end{example}
	\begin{restatable}[Accurate function approximation implies approximate closure]{lem}{score-function-approx}\label{lemma:score_function_approx}
		In the setting of Example \ref{ex:linear_softmax}, suppose that for each $\pi \in \Pi_{\Theta}$,
		\begin{equation}\label{eq:accurate_Q}
			\min_{w \in \mathbb{R}^d}  \E_{(s,i)\sim \eta_{ \pi}\otimes {\rm Unif}}\left[ \left(Q_{\pi}(s,e_i) - w^\top \phi(s,i)  \right)^2\right] < \epsilon^2/k^2. 
		\end{equation}
		Then $\Pi_{\Theta}$ satisfies Condition \ref{cond:approx_closure}. 
\end{restatable}}
\begin{proof}
	\jb{Given $\pi$, let $\hat{w}$ satisfy  $\E_{(s,i)\sim \eta_{ \pi}\otimes {\rm Unif}}\left[ \left(Q_{\pi}(s,e_i) - w^\top \phi(s,i)  \right)^2\right] =  \epsilon_0^2/k^2$ where $\epsilon_0<\epsilon/2$. Set 
		\[
		\hat{Q}_{\pi}(s,a) = \sum_{i=1}^k a_i \hat{Q}_{\pi}(s,e_i) = \sum_{i=1}^{k} a_i (\hat{w}^\top \phi(s,i)). 
		\]
		to be the resulting approximate $Q$-function. For a scalar $c>0$, consider the policy $\pi_{c \hat{w}}$ which assigns probability $\pi_{ c \hat{w}}(s,i) \propto \exp\{c \cdot \hat{Q}_{\pi}(s,e_i) \}$ to base action $i$ in state $s$. We show that $\pi_{c\hat{w}}$ approximately optimizes the policy iteration objective when $c$ is large. Note, 
		\begin{align*}
			\Bc(\pi_{c \hat{w}} \mid \eta_{ \pi}, J_{\pi}) &=  \intop \left(Q_{\pi}(s, \pi_{c \hat{w}(s)}) \right)  \eta_{ \pi}(ds)\\
			&\leq \intop \left(\hat{Q}_{\pi}(s, \pi_{c \hat{w}}(s)) \right)  \eta_{ \pi}(ds) + \intop \max_{a\in \Ac}\left| \hat{Q}_{\pi}(s, a) - Q_{\pi}(s,a) \right|\eta_{ \pi}(ds) \\
			&\leq \intop \left(\hat{Q}_{\pi}(s, \pi_{c \hat{w}}(s)) \right)  \eta_{ \pi}(ds) + \epsilon_0
		\end{align*}
		where the final inequality follows from using the identity $\E[X^2] \geq \left(\E[|X|]\right)^2$ for any random variable $X$ and observing that
		\begin{align*}
			 \intop \max_{a\in \Ac}\left| \hat{Q}_{\pi}(s, a) - Q_{\pi}(s,a) \right|\eta_{ \pi}(ds) 
			&\leq \intop \sum_{i=1}^{k}\left| \hat{Q}_{\pi}(s, e_i) - Q_{\pi}(s,e_i) \right|\eta_{ \pi}(ds) \\
			&= k \intop k^{-1} \sum_{i=1}^{k}\left| \hat{Q}_{\pi}(s, e_i) - Q_{\pi}(s,e_i) \right|\eta_{ \pi}(ds) \\
			&\leq k \sqrt{\intop k^{-1}\sum_{i=1}^{k}\left( \hat{Q}_{\pi}(s, e_i) - Q_{\pi}(s,e_i) \right)^2\eta_{ \pi}(ds)} \leq \epsilon_0.
		\end{align*}
		This implies,
		\begin{align*}
			\lim_{c\to \infty} \Bc(\pi_{c \hat{w}} \mid \eta_{ \pi}, J_{\pi}) &\leq \lim_{c\to \infty} \intop \left(\hat{Q}_{\pi}(s, \pi_{c \hat{w}}(s)) \right)  \eta_{ \pi}(ds) + \epsilon_0 \\
			&= \intop \lim_{c\to \infty} \left(\hat{Q}_{\pi}(s, \pi_{c \hat{w}}(s)) \right)  \eta_{ \pi}(ds) +  \epsilon_0 \\
			&= \intop \min_{a\in \Ac} \left(\hat{Q}_{\pi}(s, a) \right)  \eta_{ \pi}(ds) + \epsilon_0  \\
			&\leq  \intop \min_{a\in \Ac} \left(Q_{\pi}(s, a) \right)  \eta_{ \pi}(ds) +  \intop \max_{a\in \Ac}\left| \hat{Q}_{\pi}(s, a) - Q_{\pi}(s,a) \right|\eta_{ \pi}(ds) + \epsilon_0 \\
			&\leq \min_{\pi^+ \in \Pi} \Bc( \pi^+ \mid \eta_{ \pi}, J_{\pi} ) + 2\epsilon_0. 
		\end{align*}
		The first and second equality use the monotone convergence theorem together with the fact that, for each $s$, $\hat{Q}_{\pi}(s, \pi_{c \hat{w}}(s))\uparrow \max_{a\in \Ac}\hat{Q}_{\pi}(s, a)$. Since $\epsilon_0 <\epsilon/2$ we can choose a  $\hat{c}\in \mathbb{R}$ such that $\Bc(\pi_{\hat{c} \hat{w}} \mid \eta_{ \pi}, J_{\pi}) \leq \min_{\pi^+ \in \Pi}\Bc( \pi^+ \mid \eta_{ \pi}, J_{\pi} ) + \epsilon$.}
\end{proof}

\jb{The theory in \cite{agarwal2020optimality} is a tour-de-force, containing many ideas which are not reflected in the result above. Let us remark on two of them:
	\begin{enumerate}
		\item \cite{agarwal2020optimality} treat generalizations of Example \ref{ex:linear_softmax} where policies are still stochastic but are not log-linear. If $\pi_{ \theta}(s,i) \propto \exp\{f_{\theta}(s,i)\}$, then the theory of compatible function  approximation \citep{sutton2000policy}  		
		implies modeling $Q_{\pi}(s,e_i) \approx w^\top \nabla_{\theta} f_{\theta}(s,i)$ for some choice of weights $w\in \mathbb{R}^d$.  When this approximation is accurate, natural gradient updates to the parameter vector will tilt the policies' action probabilities towards selecting actions with higher $Q$-values. Because it is local in nature, such a property appears to no longer imply closure of the policy class under policy improvement steps, which involves global properties of the policy class. The choice to focus on natural gradient methods helps \cite{agarwal2020optimality} to analyze this example, since there is an explicit formula for the parameter updates which has a very close connection to compatible function approximation of the Q--function. 
		\item \cite{agarwal2020optimality} state their results in terms of a notion called \emph{transfer error.} This measures implicitly depends on both issues of approximation, as in \eqref{eq:accurate_Q}, and issues of distribution shift, as in our term $\kappa_{\rho}$. It is a very nice insight that this term is what is really needed in the analysis. 
		%
\end{enumerate}}

%% file: app_initial_distribution.tex
\section{On the necessity of an exploratory initial distribution}\label{sec: app_iniital_distribution}

Our results critically rely on using an exploratory initial distribution (see Assumption \ref{ass: exploratory rho}). This is not an artifact of the proof techniques and it is well known that, in the absence of strong assumptions on the transition kernel, policy gradient methods have poor convergence properties if applied without some form of sophisticated exploration. While this aspect of policy gradient methods is not always highlighted in the literature, many applied papers assume access to a diverse set of starting states using either explicit restarts \citep{fu2018variational, haarnoja2018soft} or some form of continual learning that aims to increase the support of a training distribution \citep{lesort2020continual, zhu2020ingredients}.

The following example, which is commonly known as a ``chain'' MDP \citep{thrun1992cient,kakade2002approximately} or the ``river swim'' problem \citep{strehl2008analysis, osband2013more}, illustrates the challenges for policy gradient in the absence of sufficient exploration. Many other examples in the reinforcement learning literature, like  the ``combination lock'' problem \citep{koenig1993complexity} and the ``grid world'' problem \citep{azar2011reinforcement} highlight the same issue. While these examples are typically used to highlight a statistical challenge, here we focus on the  optimization landscape. This example is partly inspired by one in \cite{kakade2002approximately}. A similar discussion appears also in \cite{agarwal2020optimality}. We include this section to keep the paper self contained. In addition,  it does not seem that past work has shown clearly that $\ell(\cdot)$ may have suboptimal local minima in the absence of an exploratory initial distribution, instead showing the existence of suboptimal polices with small but nonzero gradient norm.

\begin{example}\label{ex: vanishing gradient} Consider the MDP shown in Figure \ref{fig:chainMDP}. There are $N$ states and the agent can move either left (L) or right (R) from each state. The agent always begins in the leftmost state (i.e. $\rho(s_1)=1$). She incurs a cost of 2 per-period when in any state other than the leftmost or rightmost state, a cost $g(s_1)=1$ from the leftmost state and a cost of $g(s_N)=0$ per period in the rightmost state. A stationary policy $\pi \in [0,1]^{N}$ is a vector\footnote{Note that unlike Example \ref{ex: counterexample}, this policy class is closed under policy improvement.} where $\pi(s)$ specifies the probability of choosing the action R in state $s$. When the horizon is sufficiently long, the optimal policy moves right in each period. 
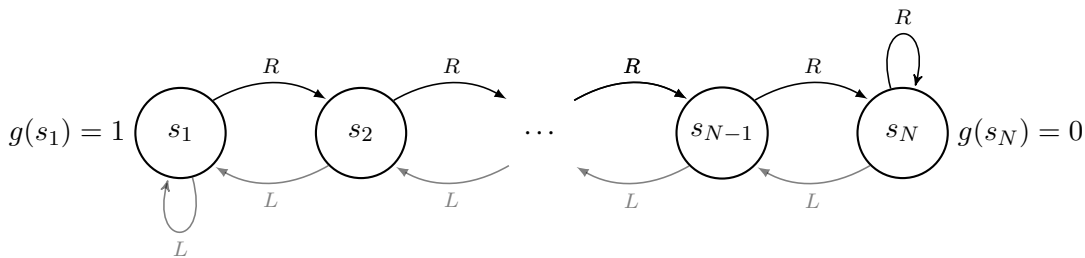
\begin{figure}
\begin{center}
	\begin{tikzpicture}[->,>=stealth',shorten >=1pt, auto,node distance=2.4cm,semithick]
	\tikzstyle{every state}=[fill=white,draw=black,thick,text=black,scale=1]
	\node[state, minimum size=1.2cm]        (A) {$s_1$};
	\node[state, minimum size=1.2cm]        (B)     [right of=A] {$s_2$};
	\node[state, minimum size=1.2cm, draw=white]        (C)     [right of=B] {\dots};
	\node[state, minimum size=1.2cm]        (E)     [right of=C] {$s_{N-1}$};
	\node[state, minimum size=1.2cm]        (F)     [right of=E] {$s_N$};
	
	\path[dashed] [-latex]  node[left] {$g(s_1)=1$ \hspace{5mm} } (A);
	\path[opacity=0.5] [-latex]  (A) edge [loop below] node[below] {\scriptsize $L$} (A);
	\path[opacity=0.5] [-latex]  (B.south west) edge [bend left] node[below] {\scriptsize $L$} (A.south east);
	\path[opacity=0.5] [-latex]  (C.south west) edge [bend left] node[below] {\scriptsize $L$} (B.south east);
	\path[opacity=0.5] [-latex]  (E.south west) edge [bend left] node[below] {\scriptsize $L$} (C.south east);
	\path[opacity=0.5] [-latex]  (F.south west) edge [bend left] node[below] {\scriptsize $L$} (E.south east);
	
	\path[-latex]  (A.north east) edge [bend left] node[above] {\scriptsize $R$} (B.north west);
	
	\path[-latex]  (B.north east) edge [bend left] node[above] {\scriptsize $R$} (C.north west);
	
	\path[-latex]  (C.north east) edge [bend left] node[above] {\scriptsize $R$} (E.north west);
	
	\path[-latex]  (C.north east) edge [bend left] node[above] {\scriptsize $R$} (E.north west);
	
	\path[-latex]  (E.north east) edge [bend left] node[above] {\scriptsize $R$} (F.north west);
	
	\path[-latex]  (F) edge [loop above] node[above] {\scriptsize $R$} (F);
	\path[-latex]  (F) node[right] {\hspace{5mm} $g(s_N)=0$} ;
	\end{tikzpicture}
\end{center}
\caption{A simple chain MDP example to illustrate how policy gradient methods face suboptimal local minima in the absence of an exploratory initial distribution.}
\label{fig:chainMDP}
\end{figure}
From Lemma \ref{lem: PG theorem}, one can calculate the policy gradient as 
\[ 
\frac{\partial  \ell(\pi)}{\partial \pi(s) } = \eta_{\pi}(s) \left( Q_{\pi}(s, R) - Q_{\pi}(s, L) \right). 
\]
We argue that a suboptimal policy $\pi$ that always moves left, i.e. $\pi(s_i)=0 \,\, \forall \, i \in \{1,\ldots,N\}$, is a local minimum of $\ell(\cdot)$. To see this, first note that the agent will always start and stay in the leftmost state, so $\eta_{\pi}(s_i) = 0$ when $i\geq 2$. The only possible nonzero component of $\nabla \ell(\pi)$ is the first term corresponding to state $s_1$. Therefore, for any policy $\pi' \in [0,1]^N$, 
\[
\langle \nabla \ell(\pi), \pi' - \pi \rangle =  \eta_{\pi}(s_1) \left( Q_{\pi}(s_1, R) - Q_{\pi}(s_1, L) \right) (\pi'(s_1) - \pi(s_1)) \geq 0,
\]
which follows as $Q_{\pi}(s_1, R) > Q_{\pi}(s_1, L)$, given that moving to $s_2$ for a single period is more costly than staying in $s_1$ and the fact that $\pi(s_1)=0$, so $\pi'(s_1) - \pi(s_1) \geq 0$ for any feasible policy $\pi'$. 

Similar issues arise under a (non-degenerate) stochastic policy. The main idea is that policies which are more likely to move left from \emph{every} state are expected to require exponentially (in the number of states) many periods to reach the rightmost state. An explicit bound confirming that the policy gradient can be exponentially small in $N$ is shown in \cite{agarwal2020optimality}. 
\end{example} 

%% file: omitted_proofs.tex

\section{Omitted proofs.}
\label{appendix B}
In this section, we provide proofs for some of the main results along with the supporting lemmas.

\subsection{General results.}\label{subsec:app_general_results}
We prove some key lemmas which are used to show our general results in Theorems \ref{thm: no bad stationary points general} and \ref{thm: convergence_rates}. We start with some useful background on Bellman operators.

\paragraph{Bellman operators.} 

For bounded cost-to-go functions, 
Bellman operators are monotone, meaning that $J \preceq J'$ implies $TJ\preceq TJ'$ and $T_{\pi}J \preceq T_{\pi}J'$, and contractive in $\|\cdot\|_{\infty}$ with modulus $\gamma$. A useful consequence of contractivity relates optimality gap to errors in the cost-to-go functions.
\begin{align}\label{eq: simple near solutions to Bellmans equation}
	\| J_{\pi} - J^* \|_{\infty}\leq \frac{1}{1-\gamma} \| J_{\pi} - TJ_{\pi}  \|_{\infty} 
\end{align} 
where $J^*$ is the optimal cost-to-go function. A simple argument \citep{bertsekas1995dynamic} shows \eqref{eq: simple near solutions to Bellmans equation}. 
\begin{align*}
	\norm{J_{\pi}-J^*}_{\infty} = \norm{T_{\pi}J_{\pi} - TJ_{\pi} + TJ_{\pi} - J^*}_{\infty} &\leq \norm{T_{\pi}J_{\pi} - TJ_{\pi}}_{\infty} + \norm{ TJ_{\pi} - TJ^*}_{\infty} \\
	&\leq \norm{T_{\pi}J_{\pi} - TJ_{\pi}}_{\infty} + \gamma \norm{J_{\pi} - J^*}_{\infty}.
\end{align*}

\paragraph{Balance equation}
The best way to understand equation \eqref{eq:balance_eq_general} below is by analogy to an equivalent undiscounted problem: $\eta_{ \pi}$ is the steady state distribution in a problem in which the next state is drawn from the restart distribution $\rho(\cdot)$ with probability $1-\gamma$ and otherwise is drawn according to the transition kernel under the policy $\pi$. This result is only used explicitly in one place, but may provide helpful intuition throughout. 

\begin{lem}\label{lem: balance equation} 
	The discounted state occupancy measure satisfies the balance equation,
		\begin{equation*}\label{eq:balance_eq_general}
		\eta_{ \pi}( \Mc) =   \intop_{\Sc}\left[ (1-\gamma) \rho(\Mc)+  \gamma P(\Mc | s, \pi(s)) \right]\eta_{\pi}(ds)    \qquad \forall \Mc \subset \Sc. 
	\end{equation*}
\end{lem}
\begin{proof}
	This can be directly verified by using the tower property of conditional expectation as follows:
	\begin{align*}
		\eta_{\pi}(\Mc) &= (1-\gamma)\sum_{t=0}^{\infty} \gamma^t \E_{\rho}^{\pi}\left[\mathbbm{1}\left( s_t \in \Mc \right)\right] \\
		&= (1-\gamma)\rho(\Mc)+ (1-\gamma)\E_{\rho}^{\pi}\left[   \sum_{t=1}^{\infty} \gamma^t  \Prob_{\rho}^{\pi}\left[ s_t \in \Mc  \mid s_{t-1}  \right]   \right] \\
		&= (1-\gamma)\rho(\Mc)+ (1-\gamma)\E_{\rho}^{\pi}\left[   \sum_{t=1}^{\infty} \gamma^t  P(\Mc | s_{t-1}, \pi(s_{t-1}))  \right] \\
		&=(1-\gamma)\rho(\Mc)+ \gamma (1-\gamma)\E_{\rho}^{\pi}\left[   \sum_{t=0}^{\infty} \gamma^t  P(\Mc | s_{t}, \pi(s_{t}))\right]\\
		&= (1-\gamma)\rho(\Mc)  + \gamma \intop_{\Sc}  P(\Mc | s, \pi(s))  \eta_{\pi}(ds).  
	\end{align*}
\end{proof}

\paragraph{Optimal policies and minimizers of the policy gradient loss.}
For the reader's convenience, we recall Lemma \ref{lem:minimizers_of_ell}, which relates minimizers of $\ell(\cdot)$ to the classic definition of optimal policies in dynamic programming. 
\optimalPolicies* 
\begin{proof}
	Recall $\ell(\pi)=(1-\gamma)\intop J_{\pi}(s)\rho(ds)$. An optimal policy $\pi^*$ satisfies $J_{\pi^*}(s)=J^*(s)$ for every state $s\in \Sc$. Since $J_{\pi}(s) \geq J^*(s)$ for each $s\in \Sc$, we have
	\begin{equation}\label{eq:diff_in_l}
	\ell(\pi) - \ell(\pi^*) = (1-\gamma)\intop  \left( J_{\pi}(s) - J^*(s)\right) \rho(ds)\geq 0.
	\end{equation}
	Since this holds for every policy $\pi$, it is clear that $\ell(\pi^*) = \min_{ \pi \in \Pi} \ell(\pi)$. 	
	
	A basic fact in measure theory states that, for a non-negative function $J:\Sc \to \mathbb{R}_+$, $\intop J d\rho =0$ if and only if $J=0$ $\rho$-almost surely. Since $J_{\pi}(s) \geq J^*(s)$ for each $s\in \Sc$, applying this fact with a choice of $J=J_{\pi}-J^*$ implies 
	equality holds in \eqref{eq:diff_in_l} if and only if $J_{\pi}-J^* = 0$  $\rho$-almost-surely.   
\end{proof} 

\paragraph{Performance difference and telescoping sums.} 
Throughout the analysis, we use a basic result which relates the difference in cost-to-go functions to the gap in Bellman's equation at future states. For any two cost-to-go functions, $J_{\pi}, J \in \Jc$, and any starting state $s_0 \in \Sc$, we have
\begin{align*}
J_{\pi}(s_0) - J(s_0) &= T_{\pi}J(s_0) - J(s_0) + T_{\pi} J_{\pi}(s_0) - T_{\pi} J(s_0) \\
&= T_{\pi}J(s_0) - J(s_0) + \gamma P_{\pi} \left( J_{\pi}(s_0) - J(s_0) \right)
\end{align*}
where we use the notation $(P_{\pi}J)(s) = \intop J(s') P(ds'| s, \pi(s))$. Unrolling this recursion, taking expectation over some initial distribution $\nu$, and noting that $(P_{\pi}J)(s_t) = \E^{\pi}\left[ J(s_{t+1}) | s_t \right]$, we have
\begin{equation}
	\label{eq: perf diff 1}
	\E^{\pi}_{\nu} \left[ J_{\pi}(s_0) - J(s_0) \right] = \E^{\pi}_{\nu} \left[ \sum_{t=0}^{\infty} \gamma^t \left[ T_{\pi}J(s_t) - J(s_t)  \right] \right],
\end{equation}
where we use the tower property of conditional expectation to simplify the telescoping sum. \cite{kakade2002approximately} use this to give a particularly convenient form, which is commonly known as the \emph{performance difference lemma}. Choosing $J = J_{\pibar}, \nu = \rho$ in \eqref{eq: perf diff 1} and recalling that $\ell(\pi)=(1-\gamma)\E_{\rho}[J_{\pi}(s_0)]$ gives
\begin{equation}\label{eq: perf diff 2}
	\ell(\pi) - \ell(\pibar) = (1-\gamma) \E_{\rho}^{\pi}\left[  \sum_{t=0}^{\infty} \gamma^t \left( T_{\pi}J_{\pibar}(s_t) - J_{\pibar}(s_t) \right)  \right] = \intop \left[ T_{\pi}J_{\pibar} -J_{\pibar} \right] d\eta_{\pi}.
\end{equation}
The second equality follows using the definition of the discounted state occupancy measure, $\eta_{\pi}(\Mc) = (1-\gamma)\E_{\rho}^{\pi} \left[\sum_{t=0}^{\infty} \gamma^t \mathbf{1}(s_t\in \Mc )\right]$ for any measurable set $\Mc \subset \Sc$.

\paragraph{A policy gradient formula.}
We give a short of the policy gradient theorem in Lemma \ref{lem: PG theorem} assuming the differentiability conditions hold. 
\pgTheorem*
\begin{proof}
	Recall that $\Bc(\theta | \eta_{ \thetabar},J_{\theta} ) = \intop J_{\pi_{\theta}} d\eta_{ \pi_{\thetabar}}$. From the performance difference lemma in \eqref{eq: perf diff 2}, we have 
	\[
	\ell(\thetabar) - \ell(\theta) =  \intop  \left[ T_\thetabar J_\theta - J_\theta \right] d\eta_{\thetabar} = \Bc(\thetabar | \eta_{\thetabar},J_{\theta} ) -  \Bc(\theta | \eta_{ \thetabar},J_{\theta} ).
	\]
	Expanding the total derivative in terms of partial derivatives gives,
	\begin{align*}
		\nabla \ell(\thetabar) &=  \nabla_{\thetabar} \, \Bc(\thetabar | \eta_{\thetabar},J_{\theta} ) \bigg\vert_{\thetabar=\theta} - \nabla_{\thetabar} \, \Bc(\theta | \eta_{\thetabar},  J_{\theta} ) \bigg\vert_{\thetabar=\theta} =\nabla_{\thetabar} \, \Bc(\thetabar | \eta_{\theta} , J_{\theta} ) \bigg\vert_{\thetabar=\theta}.
	\end{align*}
\end{proof}

\paragraph{On average Bellman equation.}
We prove the on average Bellman equation which is a key lemma we used for proving Theorem 1.
\label{app: on average Bellman}
\onAvgBellman*
\begin{proof}
	Recall that a non-negative function $f$ satisfies $\intop f d\mu = 0$ if and only if $f=0$ almost surely under the probability distribution $\mu$. We also use the fact that $J_{\pi} \succeq J^*$, by definition of the optimal cost-to-go, and $J_{\pi} \succeq TJ_{\pi}$, as shown in \eqref{eq: bellman update reduces cost}.
	
	 Take $\pi^+$ to be the policy iteration update at $\pi$, i.e. $T_{\pi^+} J_{\pi} = TJ_{\pi}$. 
	To show the left hand side, recall that by definition, $\ell(\pi) -\ell(\pi')= (1-\gamma) \intop \left(J_{\pi} -J_{\pi'} \right) d\rho$.Therefore, $\ell(\pi) = \ell(\pi^*)$ implies that,
	\begin{align*}
		0 =  \intop \left(J_{\pi} -J^*\right)d\rho  \geq \intop \left(J_{\pi} -J_{\pi^+}\right) d\rho
		&\overset{(a)}{=}  (1-\gamma)^{-1}\intop\left(  J_{\pi} - T_{\pi^+}J_{\pi} \right)  d\eta_{\pi^{+}} \\ &=(1-\gamma)^{-1}\intop\left(  J_{\pi} - TJ_{\pi} \right)  d\eta_{\pi^{+}} \\
		&\geq \intop \left(J_{\pi} -TJ_{\pi}\right) d\rho \geq 0. 
	\end{align*}
	where (a) follows by using the performance difference lemma in \eqref{eq: perf diff 2} with $\pibar=\pi^+$. The penultimate inequality uses that $\eta_{ \pi^+} \succeq (1-\gamma)\rho$ while the final inequality follows by using that $J_{\pi} \succeq TJ_{\pi}$.
	
	To show the other side, suppose $\intop (J_{\pi}-TJ_{\pi}) d\rho = 0$. Let $S_{0}=\{s:  J_{\pi}(s)-TJ_{\pi}(s) =0 \}$ and $S_0^c$ denote its complement. Since $J_{\pi}-TJ_{\pi}\succeq 0$, we must have $\rho(S_0^c)=0$. But as we assumed $\eta_{ \pi^*}$ to be absolutely continuous with respect to $\rho$, we have that $\rho(S_0^c) = 0 \implies \eta_{ \pi^*}(S_0^c)=0$. Therefore, 
	\[
	\intop (J_{\pi}-TJ_{\pi}) d\eta_{\pi^*} =0. 
	\]
	As $J_{\pi} \succeq J^*$, we have $\ell(\pi) - \ell(\pi^*) = (1-\gamma )\intop \left(J_{\pi} -J^*\right) d\rho \geq 0$. Then, we get our result by noting
	\begin{align*}
	0 \leq \ell(\pi) - \ell(\pi^*) \overset{(b)}{=} \intop  \left(  J_{\pi} - T_{\pi^*}J_{\pi} \right)  d\eta_{\pi^*} \leq \intop   \left(  J_{\pi} - TJ_{\pi} \right)  d\eta_{\pi^*} =0
	\end{align*}
	where (b) follows from the performance difference lemma in \eqref{eq: perf diff 2}.
\end{proof}

\subsection{Non-stationary policy classes: Proof of Theorem \ref{thm: finite_horizon}} 
\label{subsec: finite horizon proof}

For the reader's convenience, we restate Theorem \ref{thm: finite_horizon}.

\finiteHorizon*
\begin{proof}
To give a more transparent proof, it is helpful to develop some notation that highlights a (limited) sense in which the problem decomposes across time periods. With some abuse of notation\footnote{Technically, for this to be appropriate we should imagine $\Theta_1, \cdots, \Theta_H$ are disjoint, which we could assume without loss of generality.}, for any parameter vector $\theta=(\theta_1, \cdots, \theta_H)$, define  $\pi_{\theta_h}: \Sc_h \to \Ac$ to be the restriction of the policy $\pi_\theta$ to $\Sc_{h}$, i.e.   $\pi_{\theta}(s) = \pi_{\theta_h}(s)$ for all $s\in \Sc_{h}$. 

\paragraph{Single period PI objectives.} Similarly, define 
\[ 
\Bc_{h}(\theta_h \mid \eta, J_{\pi}) = \intop_{\Sc_h}  Q_{\pi}(s, \pi_{\theta_h}(s)) \eta(ds) 
\]
so that 
\[ 
\Bc(\theta \mid \eta, J_{\pi} ) = \sum_{h=1}^{H} \Bc_{h}(\theta_h \mid \eta, J_{\pi}). 
\]
Because the parameter space factorizes as $\Theta = \Theta_{1} \times \cdots \times \Theta_H$, this separability of the weighted Bellman objective implies,
\begin{equation}\label{eq:finite_horizon_optimality_characterization} 
\theta \in \argmin_{ \thetabar \in \Theta } \Bc(\thetabar \mid \eta, J_{\pi} ) \iff \theta_h \in \argmin_{ \thetabar_h \in \Theta_h } \Bc_{h}(\thetabar_h \mid \eta, J_{\pi}) \,\,\,\, \forall h.
\end{equation}

\paragraph{Single period characterization of stationary points.}
The policy gradient formula in Lemma \ref{lem: PG theorem} states $\nabla_{\theta} \ell(\theta) = \nabla_{\thetabar} \Bc (\thetabar  \mid \eta_{\pi_{\theta} }, J_{\pi_{\theta} } ) \vert_{\thetabar = \theta}$. Therefore, $\theta$ is a stationary point of $\min_{\thetabar \in \Theta} \ell(\thetabar)$ if and only if it is a the stationary points of the optimization problem $\min_{ \thetabar \in \Theta}  \Bc (\thetabar  \mid \eta_{\pi_{\theta} }, J_{\pi_{\theta} } )$. Because $\Theta=\Theta_1 \times \cdots \times \Theta_H$, this problem separates across time periods, and we find $\theta$ is a stationary point of $\min_{\thetabar \in \Theta} \ell(\thetabar)$ if and only if $\theta_h$ is a stationary point of $\min_{\thetabar_h \in \Theta_h}  \Bc_h(\thetabar_h  \mid \eta_{\pi_{\theta} }, J_{\pi_{\theta} } )  \,\, \forall h$. That is,
\begin{equation}\label{eq:finite_horizon_stationary_characterization} 
	\langle \nabla \ell(\theta), \theta'-\theta \rangle \geq 0 \quad \forall \theta' \in \Theta \iff \left[\frac{\partial}{\partial \thetabar_h} \Bc_h(\thetabar_h \mid \eta_{\pi_{\theta}}, J_{\pi_{\theta}}) \bigg\vert_{\thetabar_h=\theta_{h}} \right] (\theta'_h - \theta_h)  \geq 0 \quad \forall \theta'_h \in \Theta_h, \forall h
\end{equation}

\paragraph{Inductive proof.} We argue  that any stationary point $\theta$ of $\ell(\cdot)$ must satisfy $J_{\pi_{\theta}}=J^*$ $\rho$-almost surely. By Lemma \ref{lem:minimizers_of_ell}, this implies it is a minimizer of $\ell(\cdot)$. By assumption \ref{ass: exploratory rho finite horizon}, $\eta_{ \pi} \ll \rho$ for any $\pi \in \Pi_{\Theta}$. In the reverse direction, $\eta_{\pi} \succeq (1-\gamma) \rho$ by definition, which implies $\rho \ll \eta_{ \pi}$ for each $\pi \in \Pi_{\Theta}$. Therefore, we can throughout claim that certain events hold ``almost surely'' without reference to whether the base measure is $\rho$ or $\eta_{ \pi}$. 

Let $\theta$ be a stationary point of $\ell(\cdot)$. We proceed by backward induction, showing $J_{\pi_{\theta}}(s)=J^*(s)$ almost surely for $s\in \Sc_h$. As a base case, consider $h=H+1$.
By definition, $J_{\pi_{\theta}}(s)=J^*(s)=0$ for all $s \in \Sc_{H+1}$ as $\Sc_{H+1}=\{\tau\}$ contains a single costless absorbing state. 

Now, for any $h\leq H$, suppose $J_{\pi_{\theta}}(s)=J^*(s)$ almost surely for $s\in \Sc_{h+1}$. We first claim that $\Bc_{h}(\thetabar_h | \eta_{ \pi_{\theta}}, J_{\pi_{\theta}}) = \Bc_{h}(\thetabar_h | \eta_{ \pi_{\theta}}, J^*)$ for all $\thetabar_{h}\in \Theta_h$. This is a consequence of our induction hypothesis and Assumption \ref{ass: exploratory rho finite horizon}. In particular, 
\begin{align*}
0  \leq \Bc_{h}(\thetabar_h | \eta_{ \pi_{\theta}}, J_{\pi_{\theta}}) - \Bc_{h}(\thetabar_h | \eta_{ \pi_{\theta}}, J^*)  &= \intop_{\Sc_h} \left[Q_{\pi_{\theta}}(s, \pi_{\thetabar_h}(s)) - Q^*(s,\pi_{\thetabar_h}(s))\right] \eta_{ \pi_{\theta}}(ds) \\
&= \gamma \intop_{\Sc_h} \intop_{s' \in \Sc_{h+1}} [J_{\pi_{\theta}}(s') - J^*(s')] \, P(ds'|s,\pi_{\thetabar_h}(s)) \, \eta_{ \pi_{\theta}}(ds) \\
&\overset{(a)}{\leq}  \intop_{s' \in \Sc_{h+1}} [J_{\pi_{\theta}}(s') - J^*(s')] \,  \eta_{\pi_{\thetabar}}(ds') \\
&\overset{(b)}{=} 0,
\end{align*}
where we use throughout that $Q_{\pi_{\theta}} \succeq Q^*$ or $J_{\pi_{\theta}} \succeq J^*$. Inequality (a) is justified by the following balance equation that holds for the discounted state occupancy measure: for any $h\in \{1,\cdots, H\}$ and and $\Mc\subset \Sc_{h+1}$,
\[ 
\eta_{ \pi}( \Mc) = (1-\gamma)\rho(\Mc)  +\gamma \intop_{\Sc_h } P(\Mc | s, \pi(s)) \eta_{\pi}(ds). 
\] 
This follows from the general balance equation of Lemma \ref{lem: balance equation}, which states  $\eta_{ \pi}(\Mc) = (1-\gamma) \rho(\Mc)+ \intop P(\Mc | s, \pi(s)) \eta_{\pi}(ds)$ holds for all $\Mc\subset \Sc$, and the fact that $P(\Sc_{h+1}|s, \pi(s))=0$ for $s\notin \Sc_{h}$ due to Condition \ref{cond: finite_horizon}. 
Inequality (b) uses the induction hypothesis that $J_{\pi_{\theta}}(s)=J^*(s)$ for $s \in \Sc_{h+1}$ almost surely.

Having shown $\Bc_{h}(\thetabar_h | \eta_{ \pi_{\theta}}, J_{\pi_{\theta}}) = \Bc_{h}(\thetabar_h | \eta_{ \pi_{\theta}}, J^*)$ for all $\thetabar_{h}\in \Theta_h$, we now complete the induction step. 
 As $\theta$ is a stationary point of $\ell(\cdot)$,   the characterization of stationary points in \eqref{eq:finite_horizon_stationary_characterization} implies that $\theta_h$ is a stationary point of the optimization problem $\min_{\thetabar_h\in \Theta_h }  \Bc_{h}(\thetabar_h | \eta_{ \pi_{\theta}}, J_{\pi_{\theta}})$, or equivalently, of $\min_{\thetabar_h\in \Theta_h }  \Bc_{h}(\thetabar_h | \eta_{ \pi_{\theta}}, J^*)$. But in Condition \ref{cond:finite_horizon_stationarity} we assumed that $\thetabar \mapsto \Bc(\thetabar \mid \eta_{ \pi_{\theta} }, J^*)$ has no suboptimal stationary points. By the separability structure highlighted in \eqref{eq:finite_horizon_optimality_characterization}, this implies that $\thetabar_h \mapsto \Bc_h(\thetabar_h \mid \eta_{ \pi_{\theta} }, J^*)$ also has no suboptimal stationary points, so we have shown  $\theta_h \in \argmin_{\thetabar_h \in \Theta_h} \Bc_h\left( \thetabar_h \mid \eta_{ \pi_{\theta} }, J^* \right)$. Putting it all together, we get, 

\begin{align*}
     \intop_{\Sc_h} J_{\pi_{\theta}}(s)  \eta_{ \pi_{\theta} }(ds) = \intop_{\Sc_h} Q_{\pi_{\theta}}(s, \pi_{\theta_h}(s))  \eta_{ \pi_{\theta} }(ds)=   \Bc_h(\theta_h \mid \eta_{ \pi_{\theta} }, J_{\pi_{\theta}}) &= 
     \Bc_h(\theta_h \mid \eta_{ \pi_{\theta} }, J^*)\\
      &= \min_{\thetabar_h \in \Theta_h} \Bc_h\left( \thetabar_h \mid \eta_{ \pi_{\theta} }, J^* \right) \\
     &= \min_{ \thetabar_{h} \in \Theta_h}  \intop_{\Sc_h} Q^*( s , \pi_{\thetabar_h}(s) )\eta_{ \pi_{\theta} }(ds)\\
     &\overset{(c)}{=} \intop_{\Sc_{h}} J^*(s)  \eta_{ \pi_{\theta}}(ds),
\end{align*}    
where equality (c) applies our assumption that the policy class contains an optimal policy, i.e. there exists $\theta_h \in \Theta_h$ such that $Q^*(s, \pi_{\theta_h}(s) ) = \min_{a  \in \Ac_s}   Q^*(s, a )=J^*(s)$ for all $s\in \Sc_h$. 
Since $J_{\pi_{\theta}} \succeq J^*$, we conclude that $J_{\pi_{\theta}}(s)=J^*(s)$ for $s \in \Sc_{h}$ almost surely, completing the induction step. The statement in Theorem \ref{thm: finite_horizon} follows by invoking Lemma \ref{lem:minimizers_of_ell}.


\end{proof}

\subsection{Concentrability coefficients}
\label{subsec: app_concentrability}

\concentrabilityGeneral*
\begin{proof} The proof of part (a) follows as a simple corollary of the result in part (b).
	\paragraph{Proof of part (b).} Recall that $\pi^*$ denotes an optimal policy. Using that $J_{\pi} \succeq J^*$ and the performance difference lemma in \eqref{eq: perf diff 2}, we get
	\begin{align*} 
		(1 - \gamma)\intop \left( J_{\pi} - J^* \right) d\rho = (1 - \gamma)\| J_{\pi} - J^* \|_{1,\rho} = \ell(\pi) - \ell(\pi^*) &= \intop \left( J_{\pi} - T_{\pi^*} J_{\pi} \right) d\eta_{\pi^*} \\
		&\overset{(a)}{\leq} \intop \left( J_{\pi} - TJ_{\pi} \right) d\eta_{\pi^*} \\
		&\overset{(b)}{=}\intop \left( J_{\pi} - TJ_{\pi} \right) \left(\frac{d\eta_{\pi^*}}{d\rho} \right) d\rho \\
		&\leq  \left\|  \frac{d\eta_{\pi^*}}{d\rho}  \right\|_{\infty} \intop \left( J_{\pi} - TJ_{\pi} \right) d\rho \\
		&\overset{(c)}{=} \left\|  \frac{d\eta_{\pi^*}}{d\rho}  \right\|_{\infty} \| J_{\pi} - TJ_{\pi} \|_{1,\rho}
	\end{align*}
	where (a) follows by using that $T_{\pi'}J \succeq T J$ for any policy $\pi'$ and each $J\in \Jc$, (b) uses definition of the Radon-Nikodym derivative and (c) follows as $J_{\pi} \succeq TJ_{\pi}$.
	
	\paragraph{Proof of part (c).}
	From \eqref{eq: simple near solutions to Bellmans equation}, if $T$ is contraction with modulus $\gamma$ in $\|\cdot\|$, then $\|J - J^*\| \leq \frac{1}{(1-\gamma)}\|J - TJ^*\|$. Our result follows by noting,  	
	\[
	\| J- J^*\|_{1,\rho}  \leq C  \| J- J^*\| \leq \frac{C}{(1-\gamma)}\| J - TJ\| \leq \frac{C}{c(1-\gamma)} \| J-TJ\|_{1,\rho}
	\] 

\end{proof}
Lemmas \ref{lemma: concentrability Stopping} and \ref{lemma: concentrability LQ} which state the concentrability coefficients for optimal stopping and LQ control are proved in Appendix \ref{appendix: example details} where both these examples are treated in detail.

%% file: app_example_details.tex
\section{Example details.}
\label{appendix: example details}
\jb{
	\subsection{Finite state and action MDPs with natural parameterization.}\label{app:tabular MDPs}
	We restate and prove the convergence rate result for tabular MDPs given in Lemma \ref{lemma: convergence rate tabular}. We use a result in \citep{agarwal2020optimality} which shows that for the policy gradient objective with natural parameterization, $\nabla \ell$ is Lipschitz continuous with constant $L=\frac{2\gamma |\Ac|}{(1-\gamma)^2}$.
	\convergencerateresult*
	\begin{proof}
		We assumed that the per-step costs are normalized, i.e. $\sup_{s \in\Sc, a \in \Ac} |g(s,a)| \leq 1$. As the weighted PI objective $\pi' \mapsto \Bc(\pi'|\eta_{\pi}, J_{\pi})$ in \eqref{eq:finiteMDP_weightedPI} is (1,0)-gradient dominated, Theorem \ref{thm: convergence_rates} implies that $\ell(\cdot)$ is $(\frac{\kappa_{\rho}}{(1-\gamma)}, 0)$-gradient dominated. For finite state-actions MDPs, Theorem \ref{thm: concentrability general} implies $\kappa_{\rho} \leq \sup_{s \in \Sc} \frac{\eta_{\pi^*}(s)}{\rho(s)}$. Application of Lemma \ref{lemma:second_order_convergence_rate} also requires smoothness conditions. For this we appeal to Lemma D3 in \citep{agarwal2020optimality} which shows that $\nabla \ell$ is Lipschitz continuous with constant\footnote{Note that there is a difference of a factor of $(1-\gamma)$ in the denominator as compared to the result in Lemma D3 of \cite{agarwal2020optimality} due to our definition of the policy gradient objective. While we consider the discounted average cost, $\ell(\pi) = (1-\gamma) \sum_{s \in \Sc} J_{\pi}(s) \rho(s)$, \cite{agarwal2020optimality} consider the undiscounted objective, $\ell(\pi) = \sum_{s \in \Sc} J_{\pi}(s) \rho(s)$.} $L=\frac{2\gamma |\Ac|}{(1-\gamma)^2}$. Finally, it is easy to compute the constants $R$ and $k$ in part (1) of Lemma 3 for tabular MDPs with natural parameterization as:
		\begin{align}
			R &= \sup_{\pi, \pi' \in \Pi} \, \|\pi - \pi'\|_2 \leq \sqrt{2|\Sc|}, \label{eq: boundR}\\
			k &= \sup_{\pi \in \Pi} \, \|\nabla \ell(\pi)\|_2 \leq \frac{\sqrt{|\Ac|}}{(1-\gamma)} \label{eq: boundk}
		\end{align}
		where \eqref{eq: boundR} follows from the fact that $\sum_{j=1}^k (\pi(s, e_j) - \pi'(s, e_j))^2 \leq 2$ for any $s\in \Sc$ and $\pi \in \Pi$. For \eqref{eq: boundk}, we use that $\frac{\partial \ell(\pi)}{\partial \pi(s, e_j)} = \eta_{\pi}(s) Q_{\pi}(s, e_j)$, where $|Q_{\pi}(s, e_j)| \leq \frac{1}{(1-\gamma)}$ since we assume per-step costs are normalized and $0 \leq \eta_{\pi}(s) \leq 1$ by our definition of the discounted state occupancy measure. As $\frac{1}{k} > \frac{1}{L}$ for $\gamma \geq 1/3$, the claim follows by letting $\alpha=\frac{1}{L}$ and $c=\frac{\kappa_{\rho}}{(1-\gamma)}$ in the statement of Lemma \ref{lemma:second_order_convergence_rate} part (1).
	\end{proof}
}

\subsection{Regularized finite state and action MDPs with natural parameterization.}\label{app:regularized}

\jb{
	\regularizationimpact*}
\begin{proof}
\jb{To make the dependence on $\lambda$ explicit, we  consider the MDP $M_{\lambda}= (\Sc, \Ac, g_{\lambda}, P, \gamma, \rho)$ where 
	\[
	g_{\lambda}(s,a) = g_s^{\top}a +  \lambda \cdot D_{\rm KL}(U || a) = \sum_{i=1}^k g_0(s,e_i)a_i +  \lambda \cdot D_{\rm KL}(U || a)
	\]
	For the regularized problem, we write the appropriate terms as $T_{\lambda}^{\pi}, T_{\lambda},  \ell_{\lambda}(\pi), Q_{\lambda}^{\pi}$, $Q^*_{\lambda}$, $J^*_{\lambda}$, $J_{ \lambda}^{\pi}$. For ease of notation, we write $D(\cdot || \cdot)$ to denote $D_{\rm KL} (\cdot || \cdot)$. We prove this result in a sequence of steps.}
\\

\noindent \jb{\textit{\underline{Step 1: Cost-function decomposition}}: Using the relationship between the per-step cost function and the average cost function, we have that for any $\pi \in \Pi$,
\begin{equation}
	\label{eq: cost decomp}
	\ell_{\lambda}(\pi) = \sum_{s\in \Sc} \eta_{ \pi}(s) g_{\lambda}(s, \pi(s) ) =  \sum_{s\in \Sc} \eta_{ \pi}(s) \left( g_0(s, \pi(s) ) + \lambda \cdot D(U || \pi(s)) \right)  = \ell_{0}(\pi) + \lambda \sum_{s\in \Sc} \eta_{\pi}(s) D(U|| \pi(s)). 
\end{equation}
} 
	
\noindent \jb{\textit{\underline{Step 2: We construct a policy $\pi_{\lambda}$ with bounded suboptimality gap for the unregularized objective}}: \\
	Essentially, we show that for $\pi_{\lambda}$, 
	\begin{equation}
		\label{eq: suboptim gap}
		\ell_0(\pi_{\lambda}) \leq \min_{\pi \in \Pi} \ell_0(\pi) + \lambda
	\end{equation}
To do this, define $\pi_{\lambda}$ as the solution to the following regularized optimization problem, 
\begin{equation}
	\label{eq:pi lambda}
	\pi_{\lambda}(s) = 	\argmin_{a^{\top}e = 1}  \, \sum_{i=1}^{k} Q^*_{0}(s,e_i) a_i + \lambda \cdot D(U || a). 
\end{equation}
where $Q_0^*$ denotes the optimal state-action cost-to-go function of the unregularized MDP. We can interpret $ D(U || a  ) = -\frac{1}{k} \sum_{i=1}^{k} \log(a_i) - \log(k)$ as a log-barrier penalty function for the probability simplex with effective regularization parameter $\lambda/k$. The log-barrier regularization plays a key role in the theory of interior point methods \cite[Chaper 11]{boyd2004convex}. In particular, we use a result \cite[page 566]{boyd2004convex} which shows that optimal solutions to log-barrier regularized problems are near-optimal for un-regularized ones. For our construction in \eqref{eq:pi lambda}, this implies
	\[
	\sum_{i=1}^{k} Q^*_0(s, e_i) \pi_{\lambda}(i |s) \leq \min_{a\in \Delta^{k-1}} \sum_{i=1}^{k} Q^*_0(s, e_i) a_i  + \lambda 
	\]
	where $\pi_{\lambda}(s) =\left(\pi_{\lambda}(1 |s), \cdots, \pi_{\lambda}(k |s) \right)$. In terms of Bellman operators, this means 
	\begin{equation} 
		T_0^{\pi_\lambda} J^*_0 \leq T_{0} J^*_0 + \lambda e = J^*_0 + \lambda e
	\end{equation} 
	where $e$ is a vector of 1's. Repeatedly applying $T_0^{\pi_\lambda}$ to each side of this expression and using the monotonicity and constant shift properties of the Bellman operator \citep{bertsekas1995dynamic} shows $J_0^{\pi_{\lambda}} \preceq J^*_0 + \frac{\lambda}{1-\gamma} e$. Combining this with definition of the average cost, $\ell_{0}(\pi) = (1-\gamma)\sum_{s} \rho(s) J_{0, \pi}(s)$, shows our result in \eqref{eq: suboptim gap}. \\
}

\noindent \jb{ \underline{\textit{Step 3: Bound} $D(U||\pi_{\lambda}(s))$}: 
Combining the results in \eqref{eq: suboptim gap} with the cost decomposition in \eqref{eq: cost decomp} implies,
\begin{equation}
	\label{eq: suboptim gap 2}
	\ell_{\lambda}(\pi_{\lambda}) \leq \min_{\pi \in \Pi} \ell_0(\pi_{\lambda}) + \lambda \sum_{s \in \Sc} \eta_{\pi}(s) D(U||\pi_{\lambda}(s))
\end{equation}
We  now show an upper bound on $D(U||\pi_{\lambda}(s))$ which shows our result. For simplicity, focus on a single fixed state $s$ and take 
	$$
	a^*= \pi_{\lambda}(s) =  \argmin_{a^{\top}e=1} \,\, \sum_{i=1}^{k} Q^*_{0}(s,e_i) a_i - \frac{\lambda}{k} \sum_{i=1}^{k} \log(a_i).
	$$
	Without loss of generality, assume actions are ordered such that $Q^*_0(s,e_1) \geq Q^*_0(s, e_j)$ for all $j$. In this case we must have that $a^*_1 \leq a^*_j$ for each $j$, i.e. $a^*_1 \leq 1/k$. From the first order optimality conditions, we have
	\[
	Q^*_{0}(s, e_1) - \frac{(\lambda/k)}{a^*_1} =  Q^*_{0}(s, e_j) - \frac{(\lambda/k)}{a^*_j} \,\,\,\, \forall j\geq 2. 
	\]
	This implies,
	\begin{align*}
		Q^*_{0}(s, e_1) - \frac{(\lambda/k)}{a^*_1}  = \frac{1}{1- a^*_1} \left[\sum_{j=2}^{k}  \left( Q^*_{0}(s, e_1) - \frac{(\lambda/k)}{a^*_1}\right) a_j^*\right] &=  \frac{1}{1- a^*_1} \left[\sum_{j=2}^{k}  \left( Q^*_{0}(s, e_j) - \frac{(\lambda/k)}{a^*_j}\right) a_j^*\right]\\
		&=   \frac{1}{1- a^*_1} \sum_{j=2}^{k} Q^*_{0}(s, e_j) a^*_j  - \frac{\lambda\cdot (k-1)/k }{1-a^*_1}
	\end{align*}
	Rearranging terms gives
	\[
	\frac{\lambda/k}{a^*_1} - \frac{\lambda\cdot (k-1)/k  }{1-a^*_1} =   Q^*_{0}(s, e_1) -  \frac{1}{1- a^*_1} \sum_{j=2}^{k} Q^*_{0}(s, e_j) a^*_j := \Delta. 
	\]
	Here, $\Delta$ is roughly interpreted as the excess cost of action 1 over the weighted average of other actions. We use the uniform bound $\Delta \leq c$.  Multiplying each side by $k/\lambda$ and using that $a_1^* \leq 1/k \implies 1-a_1^* \geq (k-1)/k$, 
	\[ 
	\frac{1}{a^*_1} = \frac{k}{\lambda} \cdot \Delta + \frac{k-1}{1-a_1^*} \leq \frac{k}{\lambda} \cdot \Delta + k  \leq k\left( 1+ \frac{c }{\lambda}\right).
	\]
	Since $D(U || a^*) = \frac{1}{k} \sum_{j=1}^k \log\left( \frac{1}{a^*_j} \right) - \log(k) \leq \log(1/a^*_1)  - \log(k)\leq \log\left(1+ \frac{c}{\lambda}\right)$, we have that
	\[ 
	\lambda \cdot D(U || \pi_{\lambda}(s)) \leq  \lambda \log\left(1+ \frac{c }{\lambda}\right).
	\]
	Since this holds for every $s$, combining it with \eqref{eq: suboptim gap 2} gives the desired bound
	\[ 
	\min_{\pi\in \Pi} \ell_{\lambda}(\pi) \leq  \ell_{\lambda}(\pi_{\lambda} ) \leq \min_{\pi \in \Pi} \ell_0(\pi) +    \lambda+  \lambda \log\left(1+ \frac{c }{\lambda}\right).
	\]}
\end{proof}

\subsection{Regularized finite state and action MDPs with nonlinear parameterization.}\label{app:softmax}
We prove Lemma \ref{lem:softmax_diffeomorphism} 
as discussed in Example \ref{ex:vanishin_grads_softmax}. Given a feasible descent direction $D$, which by definition means  $\pi_{ \theta}+\alpha D \in \Pi$ is a feasible policy for sufficiently small $\alpha$, our goal is to verify that there exists $N$ solving the linear system $\left[ \frac{\partial \pi_{ \theta}}{\partial \theta} \right] N = D.$  For simplicity, imagine there is only a single state, so that a policy is described by the vector $(\pi_{\theta}(1),\cdots, \pi_{\theta}(k))$.
If there were multiple states, the same argument  could be repeated for each block of parameters corresponding to each distinct state. 	

Since we have effectively fixed a choice of $\theta_1=0$ in Example \ref{ex:vanishin_grads_softmax},  $\left[ \frac{\partial \pi_{ \theta}}{\partial \theta} \right] N = D$ is a system of $k$ equations with $k-1$ variables, denoted $N=(N_2, \cdots, N_k)$. We first temporarily ignore the first linear equality and show we can solve the remaining $k-1$ equations: 
$\frac{\partial \pi_{ \theta}(i) }{\partial \theta_2}N_2 + \cdots + \frac{\partial \pi_{ \theta}(i) }{\partial \theta_k}N_k = D_i$ for each $i\geq 2$. To see this, consider the  $(k-1)\times (k-1)$ submatrix of the Jacobian that comes from dropping the first row:
\[
\left[ \frac{\partial \pi_{ \theta}(i)}{\partial\theta_j } \right]_{i,j\geq 2} = \begin{bmatrix}
	\pi_{ \theta}(2)(1-\pi_{ \theta}(2)) & -\pi_{ \theta}(2)\pi_{ \theta}(3) & \cdots & -\pi_{ \theta}(2)\pi_{ \theta}(k)\\
	-\pi_{ \theta}(2)\pi_{ \theta}(3)  & \pi_{ \theta}(3)(1-\pi_{ \theta}(3)) & \cdots & -\pi_{ \theta}(3)\pi_{ \theta}(k) \\   
	\vdots & \vdots & \vdots & \vdots \\
	-\pi_{ \theta}(2)\pi_{ \theta}(k) & -\pi_{ \theta}(3)\pi_{ \theta}(k) & \cdots & \pi_{ \theta}(k)(1-\pi_{ \theta}(k))
\end{bmatrix}.
\]
This matrix is diagonally dominant and therefore non-singular, implying a solution to these $k-1$ linear equations exist.

We now show that the first linear equation is redundant and follows from the other $k-1$. In particular, we have 
\[
\sum_{j=2}^{k} \frac{\partial \pi_{ \theta}(1) }{\partial \theta_j}N_j = \sum_{j=2}^{k} \left( - \sum_{i=2}^{k} \frac{\partial \pi_{ \theta}(i) }{\partial \theta_j}N_j \right) =   \sum_{i=2}^{k}\left( - \sum_{j=2}^{k} \frac{\partial \pi_{ \theta}(i) }{\partial \theta_j}N_j \right) = -\sum_{i=2}^{k} D_i = D_1,
\]
where the first equality uses that $\pi_{\theta}(1)+\cdots+\pi_{ \theta}(k)=1$ and the final equality uses that $D_1 + \cdots + D_k = 0$ for any feasible descent direction.

\subsection{LQ control}
\label{app:LQ}
\paragraph{Preliminaries.} We consider the LQ control problem as described in Example \ref{ex: lq control} with all the notations and assumptions introduced there. Even though this example doesn't fit our general formulation as the per period costs are not uniformly bounded, the important properties of Bellman operators that are used in our proofs hold when restricting attention to stable linear policies and quadratic value functions. Define the set of strictly convex quadratic cost to go functions as 
\[
\Jc_q = \{J : s\in \mathbb{R}^n \mapsto s^\top K s   \mid K \in \mathbb{R}^{n\times n},  K\succ 0  \}. 
\]


\begin{restatable}[Bellman operators for LQ control]{lem}{LQBellman}
	\label{lem: bellman operators for LQ}
	Consider the LQ control problem formulated in Example \ref{ex: lq control}. For $J, \bar{J}\in \Jc_{q}$ and a stable linear policy $\pi\in \{ \pi_{ \theta} : \theta \in \Theta_S \} $, the following hold: 
	\begin{enumerate}
		\item (Closure on the set of quadratic cost-to-go functions)   $T_{\pi} J \in \Jc_{q}$ and $TJ \in \Jc_{q}$.
		\item (Monotonicity) If $J\preceq \bar{J}$, then $T_{\pi} J \preceq T_{\pi} \bar{J}$ and $TJ \preceq T \bar{J}$. 
		\item (Bellman equation) $J_{\pi} = T_{\pi}J_{\pi}$ and $J_{\pi}= \lim_{k\to \infty} T_{\pi}^k J$. Moreover, $J=TJ$ if and only if $J=J^*$. 
	\end{enumerate}
\end{restatable}
We use these properties extensively for our analysis but omit the proofs as these results can be found\footnote{It is worth mentioning that many references state such results in terms of the cost matrices instead of the functions $J\in \Jc_{q}$. For example, the uniqueness of solutions to the Bellman optimality equation within $\Jc_q$ is identical to the more common statement that the algebraic Riccatti equation has a unique positive definite solution.} in standard textbooks \cite[e.g.][]{bertsekas1995dynamic}. In addition, we use the following standard property of the trace operator repeatedly in our analysis. This can be found in \citep{fang1994inequalities}, for example. Let $\lambda_{\min}(A)$ and $\lambda_{\max}(A)$ denote the minimum and maximum eigenvalues of a symmetric matrix respectively. 
\begin{lem}\label{lem:trace}
	For any two symmetric and positive semi-definite symmetric matrices $A, B \in \mathbb{R}^{n\times n}$,
	\[
	\lambda_{\min}(A) {\rm Trace}(B) \leq {\rm Trace} (AB) \leq \lambda_{\max}(A) {\rm Trace}(B)
	\]
\end{lem} 

Finally, we use several times the following observation. Recall that the second moment matrix of the initial state, denoted $\Sigma_{\rho} = \E_\rho\left[ s_0 s_0^\top\right]$ is assumed to be finite and positive definite.
\begin{lem}\label{lem:lq_trace_argument}
	For a square matrix $M$ with Frobenius norm $\| M\|_{F} = \sqrt{{\rm Trace}(M^\top M)}$,  
	\[ 
	\lambda_{\min}\left( \Sigma_\rho  \right)  \| M\|_F^2  \leq \E_{\rho}\left[\| Ms_0\|_2^2 \right]\leq \lambda_{\max}\left( \Sigma_\rho  \right)  \| M\|_F^2. 
	\]
\end{lem}
\begin{proof}
	Using Lemma \ref{lem:trace} gives
	\begin{align*}
		\E_{\rho}\left[\| Ms_0\|_2^2 \right] = \E_{\rho}\left[s_0^\top M^\top M s_0 \right] = \E\left[ {\rm Trace}\left(s_0^\top M^\top M s_0   \right)\right]   &= {\rm Trace}\left( M^\top M  \Sigma_\rho \right) \\
		&\geq  \lambda_{\min}\left( \Sigma_\rho  \right) {\rm Trace}(M^\top M)\\
		&= \lambda_{\min}\left( \Sigma_\rho  \right)  \| M\|_F^2.
	\end{align*}
	An identical argument yields a corresponding upper bound.
\end{proof}

\paragraph{Stability in LQ control.}
The following lemma is a straightforward adaption of classical understanding of stability in linear dynamical systems to our setting, which considers the cumulative discounted cost incurred from a random initial state. To ensure reproducibility, a proof of the following result is given in the supplementary Technical report \citep{technicalReport}.   
\begin{restatable}[]{lem}{LQStable}
\label{lem:lq_stability}  In the LQ control problem formulated in Example \ref{ex: lq control}, 
	$\ell(\theta)<\infty$ if and only if $\theta \in \Theta_S$. 
\end{restatable}

\paragraph{Smoothness properties for LQ control.} 
Recall the cost cost function in LQ control is  \[J_{\pi_{ \theta}}(s_0)=\lim_{T\to \infty} \sum_{t=0}^{T} \gamma^t s_t^\top (\theta^\top R \theta + C)s_t   \quad \text{where} \quad s_t=[A+B\theta]^t s_0.
\]       
Since $s_t$ is differentiable in $\theta$, differentiability properties for finite $T$ follow almost immediately. For stable policies, $(\sqrt{\gamma})^t s_t \to 0$ at a geometric rate, so later terms in the sum have a negligible contribution to the total cost. It is then natural that smoothness properties hold for stable policies, which is shown carefully by \cite{rautert1997computational} (though derivative calculations seem to date back even to \cite{kalman1960contributions}). The next lemma states this finding in a form that is sufficient 
to apply results on the convergence of first-order algorithms like Lemma \ref{lem: pgd reaches global optimum}. The idea is that, beginning with a stable linear policy $\theta_0$, iterates produced by first order methods with appropriate step-sizes are assured to stay in the sublevel set $C_{\ell(\theta_0)}$, which only contains stable policies. 


\LQloss*
\begin{proof}
	That any sublevel set only contains stable policies, i.e. $C_{\alpha} \subseteq \Theta_{\text{S}}$, is an immediate consequence of Lemma \ref{lem:lq_stability}. With a little bit of algebraic simplification (see \cite{rautert1997computational} in continuous time and \cite{bu2019lqr} in discrete time), one can show that $\ell(\theta)$ is twice continuously differentiable for any $\theta \in \Theta_{\text{S}}$ and hence over sublevel sets (as $C_{\alpha} \subseteq \Theta_{\text{S}})$. 
	
	We show that sublevel sets are compact by showing that they are closed and bounded. As $\ell(\cdot)$ is continuous over $\Theta_{\rm S}$, by definition its sublevel sets are closed. We show $\ell(\theta)$ is a coercive function, meaning  $\lim_{\|\theta\| \to \infty} \ell(\theta) = \infty$. (By the equivalence of norms in finite dimensional spaces, the definition does not depend on the choice of matrix norm.)  By definition, sublevel sets of a coercive function are bounded, (see for example \citep{peressini1988mathematics}) so this completes our argument.
	To show $\ell(\cdot)$ is coercive, we lower bound it using Lemma \ref{lem:lq_trace_argument} as 
	\begin{align*}
		(1-\gamma)^{-1}\ell(\theta) = \E_{\rho} \left[ \sum_{t=0}^\infty \gamma^t s_t^\top (\theta^\top R \theta + C)s_t \right] \geq \E_{\rho}\left[ (\theta s_0)^\top R (\theta s_0) \right] &\geq \lambda_{\min}(R) \E_{\rho}\left[ \| \theta s_0\|_2^2\right] \\
		&\geq \lambda_{\min}(R) \lambda_{\min}(\Sigma_\rho) \| \theta \|_F^2,  
	\end{align*}
	which clearly tends to infinity as $ \| \theta \|_F^2 \to \infty$. Recall that $\Sigma_{\rho} = \E_\rho\left[ s_0 s_0^\top\right]$. 
	
	To prove the final claim, observe that as $\ell(\cdot)$ is twice continuously differentiable, $\|\nabla^2 \ell(\theta) \|_2$ is a continuous function. Because any sublevel set $C_{\alpha}$ of $\ell(\cdot)$ is compact, the Extreme Value Theorem implies $\|\nabla^2 \ell(\theta) \|_2$ is bounded on any sublevel set, i.e. $\max_{\theta \in C_{\alpha}} \|\nabla^2 \ell(\theta) \|_2 < \infty$.

\end{proof}

\paragraph{Optimality of stationary points for LQ control: Proof of Lemma \ref{lem: lqr stationary}.}

\LQstationary*
\begin{proof}
	If $\theta$ defines an optimal policy, then the first order necessary conditions imply $\nabla \ell(\theta)=0$. Now consider a sub-optimal stable linear policy $\pi_{\theta}$ and let $\pi_{\thetabar}$ be the policy iteration update to $\pi_{\theta}$. That is, $\thetabar$ satisfies 
	$T_{\pi_{\thetabar}} J_{\pi_{\theta}} = T J_{\pi_{\theta}}$. Set $\theta^{\alpha} = (1-\alpha)\theta + \alpha \thetabar$ for some $\alpha \in [0,1]$. As both $\pi_{\theta}$ and $\pi_{\thetabar}$ are linear policies, this implies, $\pi_{\theta^\alpha}(s)=(1-\alpha)\theta s + \alpha \thetabar s$ . For every $s\in \mathbb{R}^n$,
	\begin{align}
		T_{\pi_{\theta^{\alpha}}} J_{\pi_{\theta}}(s) = Q_{\pi_{\theta}}(s,\pi_{\theta^{\alpha}}(s) ) &=Q_{\pi_{\theta}}(s, (1-\alpha)\theta s+ \alpha \thetabar s ) \nonumber \\
		&\leq (1-\alpha)Q_{\pi_{\theta}}(s, \theta s) + \alpha  Q_{\pi_{\theta}}(s, \thetabar s) \nonumber \\
		& = (1-\alpha) 	T_{\pi_{\theta}} J_{\pi_{\theta}}(s) + \alpha	T_{\pi_{\thetabar}} J_{\pi_{\theta}}(s) \nonumber \\
		&= (1-\alpha) J_{\pi_{\theta}}(s) + \alpha	T J_{\pi_{\theta}}(s) \nonumber \\
		&=J_{\pi_{\theta}}(s) - \alpha \left(J_{\pi_{\theta}}(s)- T J_{\pi_{\theta}}(s)\right) \label{eq: LQ cost reduction descent direction}
	\end{align}
	where the first inequality uses that $a\mapsto Q_{\pi_{\theta}}(s,a)$ is convex, as noted in Section \ref{sec:Motivation_LQ}. As $J_{\pi_{\theta}} \succeq TJ_{\pi_{\theta}}$, we conclude from \eqref{eq: LQ cost reduction descent direction} that $J_{\pi_{\theta}} \succeq T_{\pi_{\theta^{\alpha}}} J_{\pi_{\theta}}$. Repeatedly applying the Bellman operator and using the monotonicity property gives,
	\begin{equation}
		\label{eq: LQ Appendix monotonicity}
		J_{\pi_{\theta}} \succeq T_{\pi_{\theta^{\alpha}}} J_{\pi_{\theta}} \succeq T_{\pi_{\theta^{\alpha}}}^2 J_{\pi_{\theta}} \succeq \cdots \succeq \lim_{k \to \infty} T^k_{\pi_{\theta^{\alpha}}}J_{\pi_{\theta^\alpha}} = J_{\pi_{\theta^\alpha}}.
	\end{equation}
	As, $J_{\pi_{\theta^\alpha}} \preceq J_{\pi_{\theta}}$, the interpolated policy $\pi_{\theta^{\alpha}}$ is stable. Then from \eqref{eq: LQ cost reduction descent direction} and \eqref{eq: LQ Appendix monotonicity}, we have 
	\[ 
	\frac{J_{\pi_{\theta^\alpha}}- J_{\pi_{\theta}} }{\alpha} \preceq \frac{ T_{\pi_{\theta^{\alpha}}}J_{\pi_{\theta}}- J_{\pi_{\theta}} }{\alpha} \preceq \left[ TJ_{\pi_{\theta}}-J_{\pi_{\theta}} \right].
	\]
	Multiplying each side by $(1-\gamma)$, taking the expectation over $s_0$ drawn from the initial distribution $\rho$, and then taking $\alpha \to 0$ gives 
	\begin{equation}
		\label{eq:LQ descent}
		\frac{d}{d\alpha} \ell(\theta^\alpha) \bigg\vert_{ \alpha=0}  \leq  (1-\gamma) \, \E_{\rho} \left[ TJ_{\pi_{\theta}}(s_0)- J_{\pi_{\theta}}(s_0) \right] = -(1-\gamma) \E_{\rho}\left[ E(s_0) \right]    
	\end{equation}
	where we have denoted the error in Bellman's equation by $E(s) \triangleq J_{\pi_{\theta}}(s) - TJ_{\pi_{\theta}}(s)$. 
	
	We know $E(s)\geq 0$ for all $s$ because $J_{\pi_{\theta}} \succeq T J_{\pi_{\theta}}$ and the inequality is strict at some $s$ because the policy $\pi_{ \theta}$ is sub-optimal by assumption and therefore  $J_{\pi_{\theta}} \neq T J_{\pi_{\theta}}$. We argue that $\E_{\rho}[E(s_0) ]>0$, showing that the right hand side of \eqref{eq:LQ descent} is negative and hence $\theta$ cannot be a stationary point. To show this, we first observe that, since $J_{\pi_{\theta}}  \in \Jc_{q}$ and $T J_{\pi_{\theta}} \in \Jc_{q}$ are both quadratic functions (See Lemma \ref{lem: bellman operators for LQ}), $E(s)$ is a quadratic function. We can then write $E(s) = s^\top K s$ for some symmetric matrix $K$ satisfying $K\succeq 0$ and $K\neq 0$. Applying Lemma \ref{lem:lq_trace_argument} gives,
	\[
	\E_{\rho}\left[ E(s_0) \right] =   \E_{\rho}\left[ s_0^\top K s_0 \right] = \E_{\rho}\left[ \|  K^{1/2} s_0\|_2^2 \right] \geq \lambda_{\min}(\Sigma_\rho) \| K^{1/2}\|_F^2 =\lambda_{\min}(\Sigma_\rho) \cdot {\rm Trace}(K) >0.   
	\] 

\end{proof}

\paragraph{Concentrability coefficient for LQ control.}
We first recall the lemma statement from Section \ref{sec:concentrability}. 
\concentrabilityLQ*

The proof leverages the performance difference lemma, described in Subsection \ref{subsec:app_general_results} and equation \eqref{eq: perf diff 2} in particular. Here, to make precise the result used in the proof, we state an analogue of that formula in linear quadratic control, which restricts to  stable policies to rule out divergent sums. The proof is essentially identical to that leading to \eqref{eq: perf diff 2} and is therefore omitted.  Observe that, since the dynamics are deterministic with states evolving according to $s_{t+1} = (A+B\thetabar) s_t$, all randomness is due to the initial state $s_0$ drawn from $\rho$.  
\begin{lem}[Performance difference lemma for LQ control] Consider the LQ control problem formulated in Example \ref{ex: lq control}. For any $\theta, \thetabar\in  \Theta_S$,
	\begin{equation*}
		\ell(\thetabar) - \ell(\theta) = (1-\gamma) \E_{\rho}^{\pi_{\thetabar}}\left[  \sum_{t=0}^{\infty} \gamma^t \left( T_{\pi_\thetabar}J_{\pi_\theta}(s_t) - J_{\pi_\theta}(s_t) \right)  \right] = \intop \left[ T_{\pi_{\thetabar}}J_{\pi_\theta} -J_{\pi_{\theta}} \right] d\eta_{\pi_{\thetabar}}.
	\end{equation*}
\end{lem}

%
%

\begin{proof}[Proof of Lemma \ref{lemma: concentrability LQ}]
	For any $\pi_{\theta} \in \Theta_{\rm S}$, by Lemma \ref{lem: bellman operators for LQ}, $J_{\pi_{\theta}} \in \Jc_{q}$ and $TJ_{\pi_{\theta}} \in \Jc_{q}$. Therefore $J_{\pi_{\theta}} - TJ_{\pi_{\theta}}\in \Jc_{q}$, which means there exists symmetric $K\in \mathbb{R}^{n\times n}$ such that $J_{\pi_{\theta}}(s) - TJ_{\pi_{\theta}}(s)=s^\top K s$ for all $s$.  Since $J_{\pi_{\theta}} \succeq TJ_{\pi_{\theta}}$, we have that $K\succeq 0$. We get 
	\begin{equation}\label{eq: lq control rhs concentrability}
		\| J_{\pi_{\theta}}- T J_{\pi_{\theta}} \|_{1,\rho} = \E_{\rho} \left[ J_{\pi_{\theta}}(s_0) - T J_{\pi_{\theta}}(s_0) \right] = \E_{\rho} \left[ s_0^\top K s_0 \right] = {\rm Trace}(K \Sigma_\rho).
	\end{equation}
	This simplifies the right hand side in the definition of $\kappa_{\rho}$. To simplify the left hand side, we use the performance difference lemma above (with a choice $\thetabar=\theta^*$), 
	\begin{align*}
		(1-\gamma )\|  J_{\pi_{ \theta}} - J^*\|_{1,\rho}  = -  \left( \ell(\theta^*) - \ell(\theta) \right) =   \intop \left[ J_{\pi_{\theta}} - T_{\pi_{\thetabar}}J_{\pi_\theta} \right] d\eta_{\pi_{\theta^*}} &\leq  \intop \left[ J_{\pi_{\theta}} - TJ_{\pi_\theta} \right] d\eta_{\pi_{\theta^*}}\\
		& = \E_{\eta_{\theta^*}}\left[s_0^\top K s_0 \right] \\
		&= {\rm Trace}\left( K \Sigma_{\eta_{ \pi_{\theta^*}}} \right), 
	\end{align*}
	where the inequality used that $TJ \preceq T_{\pi_{ \theta^*}} J$ holds for all cost-to-go functions $J$.  Combining this with \eqref{eq: lq control rhs concentrability} and applying Lemma \ref{lem:trace} gives, 
	\[
	\frac{	\|  J_{\pi_\theta}  - J^* \|_{1,\rho} }{\|  J_{\pi_\theta}  - TJ\|_{1, \rho}} \leq \frac{1}{1-\gamma} \cdot \frac{ {\rm Trace}(K \Sigma_{\eta_{ \pi_{\theta^*}}}) }{{\rm Trace}(K \Sigma_\rho)  } \leq \frac{1}{1-\gamma} \cdot \frac{ \lambda_{\max} \left( \Sigma_{\eta_{ \pi_{\theta^*}}} \right)  }{ \lambda_{\min}(\Sigma_\rho) }. 
	\]
\end{proof}

\subsection{Optimal Stopping}
\label{subsec:app_optimalstopping}
We now consider the optimal stopping problem as described in Example \ref{ex:stopping}, continuing with the notation and assumptions introduced there. Recall that in our formulation, for each context $x \in \Xc$, the offer distribution is assumed to have a density, $q_x(\cdot)$, with continuous derivative and support $\{ y \in \mathbb{R}  : q_x(y)>0 \}=(y_{\min}, y_{\max})$ where $y_{\min} >0$. We set $\Yc = [y_{\min}, y_{\max}]$. We also assume the initial distribution places zero probability on trivial instances that begin in the terminal state (i.e $\rho(\tau)=0$) and factorizes over continuation states as $\rho(x,dy)=\nu(x) q_x(y)dy$ where $\nu(x) > 0$ for all $x \in \Xc$. 

\paragraph{Preliminaries and notation.} 
Technically, the state at time $t$ consists of both the context $x_t$ and the offer $y_t$. But since $y_t$ depends only on $x_t$ and not on the previous state, it is helpful to work directly with the state variable $x_t$ and then separately take expectations over offers drawn from $q_{x_t}(\cdot)$. To this end, we develop some specialized notation. Let $\eta'_{\pi}$ denote the marginal distribution over $\Xc \cup \{\tau\}$ under the discounted state occupancy measure $\eta_{ \pi}$, and note that
\begin{equation}
	\label{eq:eta_factorization optimal stopping}
	\eta_{\pi}\Big( \{ x\} \times (y_1, y_2) \Big) = \eta'_{\pi}(x)  \intop_{y_1}^{y_2} q_{x}(y) dy. 
\end{equation}

We find it convenient to directly work with $\eta'_{\pi}$ and $q_{x}(y)$. We now state several formulas in terms of $\eta'_{\pi}$ and $q_{x}(y)$ that will be used throughout the analysis. For an action $a\in [0,1]$, indicating a probability of stopping, 
\[
Q_{\pi}((x,y),a) = \left[ ay + (1-a) \gamma\sum_{x' \in \Xc} p(x'|x) \int_{\Yc} J_{\pi}((x', y')) q_{x'}(y') dy' \right] = ay + (1-a) c_{\pi}(x)
\]
where
\begin{equation}
	c_{\pi}(x) := \gamma \sum_{x'\in \Xc} p(x'|x) \int_{\Yc} J_{\pi}(x', y') q_{x'}(y') dy' \in [0, \gamma y_{\max}] 
\end{equation}
is called the ``continuation value'' from context $x$ under policy $\pi$. That $c_{\pi}(x) \leq \gamma y_{\max}$ is due the basic fact that $ J_{\pi}(x', y')\leq y_{\max}$, i.e. no policy can accrue expected reward exceeding the maximum possible offer.  Similarly, the weighted policy iteration objective becomes,
\begin{align}\nonumber
	\Bc\left( \theta \mid \eta_{ \pi}, J_{\pi}  \right) &= \sum_{x \in \Xc} \eta'_{ \pi}(x) \intop_{y_{\min}}^{y_{\max}} Q_{\pi}( (x,y)\, ,\,    \mathbbm{1}(y > \theta_x)  ) q_{x}(y)dy   \\
	&= \sum_{x\in \Xc} \eta'_{ \pi}(x) \intop_{y_{\min}}^{y_{\max} } \left[y\mathbbm{1}(y > \theta_x) + c_{\pi}(x)\mathbbm{1}(y < \theta_x)    \right] q_{x}(y)dy. \label{eq:stopping_formula_for_B}
\end{align}
Here we use that there is zero value-to-go from the terminal state $\tau$ to restrict the sum to continuation states. We now proceed to verify all the conditions needed to apply our general results.

\paragraph{Condition \ref{cond: closure}: Closure under policy improvement.}
It is easy to verify that the class of threshold policies is closed under policy improvement. For any $\pi \in \Pi_{\Theta}$, the policy iteration update for any state $s=(x,y) \in \Sc_{\rm C}$ is given by
\[
\pi^+(x,y) = \argmax_{a \in \{0,1\}} \, Q_{\pi}((x,y),a) = \mathbbm{1}(y > c_{\pi}(x)). 
\]
This result can be seen immediately from the formula for $Q_{\pi}(\cdot)$ given above. Clearly the policy iteration update is another threshold policy. In particular, the policy iteration update to $\pi$ is a new threshold policy $\pi_{ \theta^+}$ with parameters
\begin{equation}\label{eq:stopping_pi_update}
	\theta_x^+  = \max\{y_{\min} \, , \, c_{\pi}(x) \}.
\end{equation}
for each $x\in \Xc$. This works because a policy with stopping threshold $y_{\min}$ accepts the next offer with probability one when $c_{\pi}(x)<y_{\min}$ and otherwise and we are assured that $c_{\pi}(x)\in [y_{\min}, \gamma y_{\max}]$ is a feasible choice for $\theta_x^+$.

\paragraph{Condition \ref{cond: stationary}: No suboptimal stationary points for the weighted PI objective.} \emph{This result here is implied by the gradient dominance result that follows, but it provides a warmup for that analysis.}

We now show that $\theta \mapsto \Bc(\theta|\eta_{\pi}, J_{\pi})$ has no suboptimal stationary points for each $\pi \in \Pi_{\Theta}$. As we formulate the optimal stopping example as a maximization problem, any stationary point $\theta$ satisfies,
\begin{equation*}
	\frac{\partial }{\partial \theta_x} \Bc(\theta|\eta, J_\pi) \cdot (\theta'_x - \theta_x) \leq 0 \qquad \forall\theta'_x \in \Yc , 
\end{equation*}
for every $x\in \Xc$. It is possible to separate the stationarity condition into a componentwise inequality in this manner because the parameters space is the Cartesian product $\Theta= \Yc^{|\Xc|}$.

From \eqref{eq:stopping_formula_for_B}, we have the following formula for the derivative:
\begin{equation}\label{eq:policy_grad_opt_stopping}
	\frac{\partial}{\partial \theta_x} \Bc(\theta \mid \eta_{ \pi}, J_\pi) =  \left(c_{\pi}(x) - \theta_x  \right)  \eta'_{\pi}(x)q_x(\theta_x) 
\end{equation}
Therefore, $\theta$ is stationary point when $\left(c_{\pi}(x) - \theta_x  \right)  \cdot (\theta'_x  - \theta_x) \leq 0$ for all $\theta'_x \in  [y_{\min}, y_{\max}]$. If $c_{\pi}(x) \in (y_{\min}, y_{\max})$ is in the interior of the feasible region, this implies $\theta_x = c_{\pi}(x)$.  Otherwise (since it is impossible to have $c_{\pi}(x) \geq y_{\max}$), we have  $c_{\pi}(x)\leq y_{\min}$  and a stationary point must satisfy $\theta_x = y_{\min}$. We have found any stationary point $\theta$ satisfies, $\theta_x = \max\{ y_{\min}, c_{\pi}(x) \}$, which matches the formula \eqref{eq:stopping_pi_update} for the policy iteration update  that maximizes $\theta \mapsto \Bc(\theta \mid \eta_{ \pi}, J_{\pi})$. 

\paragraph{Concentrability coefficient for optimal stopping.} 
We first recall the claim.
\concentrabilityStopping*
\begin{proof}
	We show that the Bellman operator $T$ is a contraction with modulus $\gamma$ in $\| \cdot\|_{1,\mu}$. The proof then follows immediately using part (c) of Theorem \ref{thm: concentrability general}.
	
	For a policy that never stops, the stationary distribution over continuation states, $(x,y) \in \Sc_{\rm C}$ factorizes as $\mu(x,y) = \mu'(x) q_x(y)$ where $\mu'$ is the marginal stationary distribution over context states $\Xc$ such that $\mu'(x') = \sum_{x \in \Xc} \mu'(x) p(x'|x)$. Then, for any bounded cost-to-go functions $J,J'\in \Jc$,
	\begin{align*}
		\|TJ-TJ'\|_{1,\mu} &= \sum_{x \in \Xc} \mu'(x) \int_{\Yc} \left| TJ(x,y) - TJ'(x,y) \right| q_x(y) dy. 
	\end{align*}
	By definition, 
	\[
	T J(x,y) = \max \{y, \gamma \sum_{x' \in \Xc} p(x'|x) \int_{\Yc} J(x',y') q_{x'}(y') dy' \}.
	\]
	Note that for any scalars $(x_1, x_2, y)$, we have $|\max\{y,x_1\} - \max\{y,x_2\}| \leq |x_1 - x_2 |$. Therefore, 
	\begin{align}
		\left| TJ(x,y) - TJ'(x,y) \right| \leq \gamma \sum_{x' \in \Xc} p(x'|x) \int_{\Yc} \left| J(x',y') - J'(x',y') \right| q_{x'}(y') dy'. \label{eq:contraction_opt_stopp_appx}
	\end{align}
	As the right hand side in \eqref{eq:contraction_opt_stopp_appx} is independent of $y$, integrating \eqref{eq:contraction_opt_stopp_appx} with respect to $q_x(\cdot)$ gives
	\[
	\int_{\Yc} \left| TJ(x,y) - TJ'(x,y) \right| q_x(y) dy \leq \gamma \sum_{x' \in \Xc} p(x'|x) \int_{\Yc} \left| J(x',y') - J'(x',y') \right| q_{x'}(y') dy'. 
	\]
	Therefore, 
	\begin{align*}
		\|TJ-TJ'\|_{1,\mu} &= \sum_{x \in \Xc} \mu'(x) \int_{\Yc} \left| TJ(x,y) - TJ'(x,y) \right| q_x(y) dy \\
		&\leq \gamma \sum_{x \in \X} \mu'(x) \sum_{x' \in \Xc} p(x'|x) \int_{\Yc} \left| J(x',y') - J'(x',y') \right| q_{x'}(y') dy' \\
		&\overset{(a)}{=} \gamma \sum_{x' \in \X} \mu'(x') \intop_{\Yc} \left| J(x',y') - J'(x',y') \right| q_{x'}(y') dy' \\
		&= \gamma \| J-J' \|_{1,\mu}
	\end{align*}
	where (a) follows as $\mu'$ is the stationary distribution over $\Xc$. For $\rho = \mu$, we have $C,c=1$ in part (c) of Theorem \ref{thm: concentrability general}, implying that $\kappa_\rho \leq 1$.
\end{proof}

\paragraph{Further results in the supplementary technical report.}
One expects that many smoothness results hold for this problem (since with a continuous offer distribution, infintesmal changes to a stopping threshold should have an infintesimal impact on performance), and we have already calculated one derivative in  \eqref{eq:policy_grad_opt_stopping}. For completeness, we give a detailed verification of the differentiability properties in Condition \ref{cond: diff} in the \cite{technicalReport}.  

For a reader who is interested in this specific optimal stopping problem, we note that it possible to prove stronger results that yield \emph{convergence rates} rather than just convergence to an optimal policy. For example, the electronic companion proves the following gradient dominance and smoothness results, which are sufficient to guarantee the convergence rates in Lemma \ref{lemma:second_order_convergence_rate}.  
Notice that the gradient dominance constant depends on a measure of the degree of uniformity in the offer distribution $q_{x}(\cdot)$. We know $\beta$ is finite because $\Xc$ is finite, $q_{x}(y)>0$ for each $y\in \Yc$ (by assumption), and $\Yc$ is compact. 
\begin{restatable}[Gradient dominance for optimal stopping]{lem}{graddomoptstopp}
	\label{lem:grad_dom_of_optstopp}
	Consider the optimal stopping problem formulated in Example \ref{ex:stopping}. For any $\pi \in \Pi_{\Theta}$,  the function $\theta \mapsto \Bc(\theta | \eta_{\pi}, J_{\pi} )$ is $(\beta, 0)$--gradient-dominated where $\beta = \max_{x \in \Xc, y\in \Yc} q_{x}(y) / \min_{x \in \Xc, y\in \Yc} q_{x}(y)$. 
\end{restatable}
\begin{restatable}[]{lem}{optstoppcostsmooth}
	\label{lem:smoothness_opt_stopp}
	For the optimal stopping problem in Example \ref{ex:stopping}, $\max_{\theta \in \Theta}  \| \nabla^2 \ell(\theta) \| < \infty$.  
\end{restatable}

\subsection{Finite horizon inventory control} 

\paragraph{Differentiability and derivative calculations.} In inventory control, an attractive approach for computing derivatives with respect to policy parameters is described in detail in \cite{glasserman1995sensitivity}. One can compute derivatives of expected costs with respect to the base-stock levels by calculating the derivative for many simulated sample paths (i.e. realizations of the initial state and demands) and averaging the result. Making such an argument rigorous requires justifying the exchange of an integral and derivative. This is essentially treated in past work like \cite{glasserman1995sensitivity}, but for reproducibility we carefully verify the partial differentiability requirements in Condition \ref{cond: diff} in the technical report \citep{technicalReport}. 
The proof also shows that under any base-stock policy,  the distribution of the inventory levels $(x_0, x_1,\cdots)$ has a density, which is a basic consequence of the assumption that the demands follow a continuous distribution.

\paragraph{Verifying Condition \ref{cond:finite_horizon_stationarity}.}
The following lemma shows how Condition \ref{cond:finite_horizon_stationarity} holds for the finite horizon inventory control problem.
\begin{lem}
	\label{lemma:no_stationary_points_inventory}
	Consider the finite horizon inventory control problem in Example \ref{ex:inventorycontrol}. Let $J^*$ be the cost-to-go function corresponding to the optimal policy. Then, for any $\pi, \pi_{\theta} \in \Pi_{\Theta}$, the weighted policy iteration objective $\Bc(\theta|\eta_{\pi}, J^*)$ has no suboptimal stationary points.
\end{lem}

\begin{proof}
	Recall $Q^*(s,a) = Q_{\pi^*}(s,a)$ denotes the Q-function corresponding to an optimal policy. We follow a classical approach to rewriting costs as a function of the target inventory level $x+a$ rather than the state and action. We find
	\begin{align*}
		Q^*((x,h), a) &= c \cdot a +  \E_{w}\left[ b\max\{x+a-w, 0\} + p\max\{-x-a+w, 0\} + J^*((x+a- w, h+1) ) \right] \\
		&= c \cdot x + G_{h}(x+a)
	\end{align*}
	where the expectation is taken over the demand distribution and  $G_{h}(y) :=  \E_{w}[ b\max\{y-w, 0\} + p\max\{-y+w, 0\} + J^*((y- w, h+1) )].$ Here $y$ is thought of as a target inventory level. The function $G_{h}(\cdot)$ is well known to be convex \citep[see e.g][]{bertsekas1995dynamic}.  
	
	Recall that $\pi_{\theta}((x,h)) = \max\{  \theta_h -x \, , \, 0  \}$.  We can then calculate the derivative of the weighted PI costs in terms of $G_{h}(\cdot)$ as  
	\begin{align*}
		\frac{\partial}{\partial \theta_h} \Bc(\theta|\eta_{\pi}, J^*) &= \frac{\partial}{\partial \theta_h} \int Q^*\left( (x,h),\pi_{\theta}(x,h) \right) \, \eta_{\pi}(dx,h) \\
		&= \frac{\partial}{\partial \theta_h} \left[  \int_{x < \theta_h} Q^*((x,h),\theta_h - x) \, \eta_{\pi}(dx,h) + \int_{x > \theta_h} Q^*((x,h),0) \, \eta_{\pi}(dx,h) \right]\\
		&\overset{(a)}{=} \int_{x < \theta_h} \frac{\partial}{\partial \theta_h} Q^*((x,h),\theta_h - x) \, \eta_{\pi}(dx,h) \\
		&= \int_{x < \theta_h}   G'_{h}(\theta_h)  \, \eta_{\pi}(dx,h)\\
		&= G'_{h}(\theta_h) \int_{x < \theta_h} \eta_{\pi}(dx,h) 
		\end{align*}
		where (a) uses the Leibniz rule. Since $\theta_h$ is constrained to lie above 0, and negative (i.e. back-ordered) inventory levels are possible under the initial distribution, we know $ \int_{x < \theta_h} \eta_{\pi}(dx,h)>0$. Let $\theta^*$ an optimal vector of base-stock levels, so $\theta_h^* \in \argmin_{y} G_{h}(y)$. If $\theta_h$ is not a minimizer of $G_{h}(\cdot)$ (so it is a suboptimal base-stock level), then by convexity  $G'_{h}(\theta_h) \cdot (\theta^*_{h}-\theta_h) <0$, implying $\theta_h$ cannot be a stationary point of $\Bc(\cdot | \eta_{ \pi}, J^*)$. 
\end{proof}
	
\subsection{Linear MDPs: proof of Lemma \ref{lem:linear_mdps}}\label{subsec:app_linear_mdp_proof}
\jb{
The proof sketch in the body of the paper already established Condition \ref{cond: closure} and that
\[
\frac{d}{d\alpha}  \Bc(\pi_{\theta^{\alpha}} \mid \eta_{\pi}, J_{\pi}) = \intop   \frac{d}{d\alpha} Q_{\pi}(s, \pi_{\theta^{\alpha}}(s))  \eta_{\pi}(ds) = \intop   \frac{d}{d\alpha} Q^{\theta^+}(s, \pi_{\theta^{\alpha}}(s))  \eta_{\pi}(ds). 
\]
 Our remaining goal is to show that $\frac{d}{d\alpha} Q^{\theta^+}(s, \pi_{\theta^{\alpha}}(s))<0$ for any state at which $Q^{\theta^\alpha}(s, \cdot) \neq Q^{\theta^+}(s, \cdot)$. To see this, note that one can rewrite the problem  $\max_{a\in \Ac} Q_{\pi}(s, a)$  as $\max_{a_{1:k-1} \in \Xc} f(a_{1:k-1}, y)$  where $f$ is as in Lemma \ref{lem:linear_mdp_helper},  $a_{1:k-1}$ are components of the action except $a_k$ (which is redundant, as $a_k=1-a_1- \cdots - a_{k-1}$) and $y_j =  e_j^\top \Phi(s)\theta - e_k^\top\Phi(s)\theta$. Applying Lemma \ref{lem:linear_mdp_helper} then establishes the claim.  
\begin{lem}\label{lem:linear_mdp_helper}
Define $\Xc = \{ x\in (0,1)^{k-1} : \sum_{i=1}^{k-1} x_i <1 \}$ and $f:\Xc \times \mathbb{R}^{k-1} \to \mathbb{R}$ by
\[
f(x, \theta) = \theta^\top x + \lambda H(x)
\]
where $H(x) = \sum_{i=1}^{k} \frac{1}{k} \log\left( \frac{1/k}{x_i} \right)$ with $x_k\equiv 1-\sum_{i=1}^{k-1} x_i$.   
For fixed $\theta_0 \in \mathbb{R}^{k-1}$, define
$$
h(\theta) =  f( x^*(\theta),  \theta_0) \quad \text{where}\quad x^*(\theta) \equiv \argmin_{x\in \Xc} f(x, \theta).
$$ 
Then $h(\cdot)$ is differentiable,  $\theta_0$ is its unique minimizer, and $(\theta_0-\theta)^\top \nabla h(\theta)= - \| \theta_{0}-\theta\|_{(\nabla^2 H(x^*(\theta)))^{-1}}^2 / \lambda <0$. 
\end{lem}
\begin{proof}
The function $H$ satisfies $\nabla^2 H(x)\succ 0$. 	
It is well known that $x^*(\theta)$ satisfies $x^*_i(\theta)= C e^{-\theta_i /\lambda}$ where $C= 1+ \sum_{j=1}^{k} e^{-\theta_j /\lambda}$. Instead of using this formula, we use implicit differentiation to derive a formula for $\nabla x^*(\theta)$. Since $x^*(\theta)$ is an interior solution (i.e. $0<\sum_{i=2}^{k} x^*(\theta)_i<1$), the first order condition implies $\theta +  \lambda \nabla H(x^*(\theta) ) = 0$. Differentiating this expression again and re-arranging terms yields the formula $\nabla x^*(\theta) = - \frac{1}{\lambda} \left( \nabla^2 H( x^*(\theta))\right)^{-1} \in \mathbb{R}^{k\times k}$.   
Then 
\[
\nabla h(\theta) =  (\nabla x^*(\theta))  \theta_0 + \lambda  (\nabla x^*(\theta)) \nabla H(x^*(\theta)) =  (\nabla x^*(\theta))  (\theta_0 -\theta) = - \lambda^{-1}  \left( \nabla^2 H( x^*(\theta))\right)^{-1} (\theta_0 -\theta)
\]
where the second equality uses the first order optimality condition. Taking the dot product of both sides of the equation above with $\theta_0 -\theta$ yields the result. 
\end{proof}
}

%% file: policy_grad_optimality.bbl
\begin{thebibliography}{115}
\providecommand{\natexlab}[1]{#1}
\providecommand{\url}[1]{\texttt{#1}}
\expandafter\ifx\csname urlstyle\endcsname\relax
  \providecommand{\doi}[1]{doi: #1}\else
  \providecommand{\doi}{doi: \begingroup \urlstyle{rm}\Url}\fi

\bibitem[Agarwal et~al.(2020)Agarwal, Kakade, Lee, and
  Mahajan]{agarwal2020optimality}
Alekh Agarwal, Sham~M Kakade, Jason~D Lee, and Gaurav Mahajan.
\newblock Optimality and approximation with policy gradient methods in markov
  decision processes.
\newblock In J.~Abernethy and S.~Agarwal, editors, \emph{Proceedings of Thirty
  Third Conference on Learning Theory}, volume 125, pages 64--66. (PMLR), 2020.

\bibitem[Agarwal et~al.(2017)Agarwal, Allen-Zhu, Bullins, Hazan, and
  Ma]{agarwal2017finding}
Naman Agarwal, Zeyuan Allen-Zhu, Brian Bullins, Elad Hazan, and Tengyu Ma.
\newblock Finding approximate local minima faster than gradient descent.
\newblock In P.~McKenzie H.~Hatami and V.~King, editors, \emph{Proceedings of
  the 49th Annual ACM SIGACT Symposium on Theory of Computing}, pages
  1195--1199. (Association for Computing Machinery, NY, USA), 2017.

\bibitem[Antos et~al.(2008)Antos, Szepesv{\'a}ri, and Munos]{antos2008learning}
Andr{\'a}s Antos, Csaba Szepesv{\'a}ri, and R{\'e}mi Munos.
\newblock Learning near-optimal policies with bellman-residual minimization
  based fitted policy iteration and a single sample path.
\newblock \emph{Machine Learning}, 71\penalty0 (1):\penalty0 89--129, 2008.

\bibitem[Asmussen and Glynn(2007)]{asmussen2007stochastic}
S{\o}ren Asmussen and Peter~W Glynn.
\newblock \emph{Stochastic simulation: algorithms and analysis}, volume~57.
\newblock (Springer Science \& Business Media, Berlin, Germany), 2007.

\bibitem[Azar et~al.(2011)Azar, Munos, Ghavamzadeh, and
  Kappen]{azar2011reinforcement}
Mohammad~Gheshlaghi Azar, R{\'e}mi Munos, Mohammad Ghavamzadeh, and Hilbert
  Kappen.
\newblock Reinforcement learning with a near optimal rate of convergence.
\newblock Technical report, INRIA Lille, 2011.
\newblock HAL Id: inria-00636615v2.

\bibitem[Bagnell et~al.(2004)Bagnell, Kakade, Schneider, and
  Ng]{bagnell2004policy}
Andrew Bagnell, Sham~M Kakade, Jeff~G Schneider, and Andrew~Y Ng.
\newblock Policy search by dynamic programming.
\newblock In S.~Thrun, L.~Saul, and B.~Sch\"{o}lkopf, editors, \emph{Advances
  in neural information processing systems}, volume~16, pages 831--838. (MIT
  Press, MA, USA), 2004.

\bibitem[Baxter and Bartlett(2001)]{baxter2001infinite}
Jonathan Baxter and Peter~L Bartlett.
\newblock Infinite-horizon policy-gradient estimation.
\newblock \emph{Journal of Artificial Intelligence Research}, 15\penalty0
  (1):\penalty0 319--350, 2001.

\bibitem[Beck(2002)]{beck2002convergence}
Amir Beck.
\newblock \emph{Convergence rate analysis of gradient based algorithms}.
\newblock PhD thesis, Tel-Aviv University, 2002.

\bibitem[Beck(2017)]{beck2017first}
Amir Beck.
\newblock \emph{First-order methods in optimization}, volume~25.
\newblock (SIAM, PA, USA), 2017.

\bibitem[Beck and Teboulle(2003)]{beck2003mirror}
Amir Beck and Marc Teboulle.
\newblock Mirror descent and nonlinear projected subgradient methods for convex
  optimization.
\newblock \emph{Operations Research Letters}, 31\penalty0 (3):\penalty0
  167--175, 2003.

\bibitem[Bertsekas and Shreve(1978)]{bertsekas1978stochastic}
Dimitir~P Bertsekas and Steven Shreve.
\newblock \emph{Stochastic optimal control: the discrete-time case}.
\newblock (Athena Scientific, MA, USA), 1978.

\bibitem[Bertsekas(1995)]{bertsekas1995dynamic}
Dimitri~P Bertsekas.
\newblock \emph{Dynamic programming and optimal control}.
\newblock (Athena scientific, MA, USA), 1995.

\bibitem[Bertsekas(1997)]{bertsekas1997nonlinear}
Dimitri~P Bertsekas.
\newblock Nonlinear programming.
\newblock \emph{Journal of the Operational Research Society}, 48\penalty0
  (3):\penalty0 334--334, 1997.

\bibitem[Bertsekas(2011)]{bertsekas2011approximate}
Dimitri~P Bertsekas.
\newblock Approximate policy iteration: A survey and some new methods.
\newblock \emph{Journal of Control Theory and Applications}, 9\penalty0
  (3):\penalty0 310--335, 2011.

\bibitem[Bertsekas(2019)]{bertsekas2019feature}
Dimitri~P Bertsekas.
\newblock Feature-based aggregation and deep reinforcement learning: A survey
  and some new implementations.
\newblock \emph{IEEE/CAA Journal of Automatica Sinica}, 6\penalty0
  (1):\penalty0 1--31, 2019.

\bibitem[Bertsekas and Tsitsiklis(1996)]{bertsekas1996neuro}
Dimitri~P Bertsekas and John~N Tsitsiklis.
\newblock \emph{Neuro-dynamic programming}, volume~5.
\newblock (Athena Scientific, MA, USA), 1996.

\bibitem[Bhandari and Russo(2021)]{technicalReport}
Jalaj Bhandari and Daniel Russo.
\newblock Global optimiality guarantees for policy gradient methods: Technical
  report with supplementary materials.
\newblock 2021.
\newblock URL
  \url{https://djrusso.github.io/docs/policy_grad_optimality_EC.pdf}.

\bibitem[Bhojanapalli et~al.(2016)Bhojanapalli, Neyshabur, and
  Srebro]{bhojanapalli2016global}
Srinadh Bhojanapalli, Behnam Neyshabur, and Nati Srebro.
\newblock Global optimality of local search for low rank matrix recovery.
\newblock In D.~Lee, M.~Sugiyama, U.~Luxburg, I.~Guyon, and R.~Garnett,
  editors, \emph{Advances in Neural Information Processing Systems}, volume~29,
  pages 3873--3881. (Curran Associates, NY, USA), 2016.

\bibitem[Blackwell(1965)]{blackwell1965discounted}
David Blackwell.
\newblock Discounted dynamic programming.
\newblock \emph{The Annals of Mathematical Statistics}, 36\penalty0
  (1):\penalty0 226--235, 1965.

\bibitem[Borkar(2009)]{borkar2009stochastic}
Vivek~S Borkar.
\newblock \emph{Stochastic approximation: a dynamical systems viewpoint},
  volume~48.
\newblock (Springer, Berlin, Germany), 2009.

\bibitem[Boyd and Vandenberghe(2004)]{boyd2004convex}
Stephen Boyd and Lieven Vandenberghe.
\newblock \emph{Convex optimization}.
\newblock Cambridge university press, 2004.

\bibitem[Bu et~al.(2019)Bu, Mesbahi, Fazel, and Mesbahi]{bu2019lqr}
Jingjing Bu, Afshin Mesbahi, Maryam Fazel, and Mehran Mesbahi.
\newblock Lqr through the lens of first order methods: Discrete-time case.
\newblock \emph{arXiv preprint arXiv:1907.08921}, 2019.

\bibitem[Caramanis and Liberopoulos(1992)]{caramanis1992perturbation}
Michael Caramanis and George Liberopoulos.
\newblock Perturbation analysis for the design of flexible manufacturing system
  flow controllers.
\newblock \emph{Operations Research}, 40\penalty0 (6):\penalty0 1107--1125,
  1992.

\bibitem[Carmon et~al.(2018)Carmon, Duchi, Hinder, and
  Sidford]{carmon2018accelerated}
Yair Carmon, John~C Duchi, Oliver Hinder, and Aaron Sidford.
\newblock Accelerated methods for nonconvex optimization.
\newblock \emph{SIAM Journal on Optimization}, 28\penalty0 (2):\penalty0
  1751--1772, 2018.

\bibitem[Chen and Jiang(2019)]{chen2019information}
Jinglin Chen and Nan Jiang.
\newblock Information-theoretic considerations in batch reinforcement learning.
\newblock In \emph{International Conference on Machine Learning}, pages
  1042--1051. PMLR, 2019.

\bibitem[Davis and Grimmer(2019)]{davis2019proximally}
Damek Davis and Benjamin Grimmer.
\newblock Proximally guided stochastic subgradient method for nonsmooth,
  nonconvex problems.
\newblock \emph{SIAM Journal on Optimization}, 29\penalty0 (3):\penalty0
  1908--1930, 2019.

\bibitem[Davis et~al.(2020)Davis, Drusvyatskiy, Kakade, and
  Lee]{davis2020stochastic}
Damek Davis, Dmitriy Drusvyatskiy, Sham Kakade, and Jason~D Lee.
\newblock Stochastic subgradient method converges on tame functions.
\newblock \emph{Foundations of computational mathematics}, 20\penalty0
  (1):\penalty0 119--154, 2020.

\bibitem[Defazio et~al.(2014)Defazio, Bach, and
  Lacoste-Julien]{defazio2014saga}
Aaron Defazio, Francis Bach, and Simon Lacoste-Julien.
\newblock {SAGA}: A fast incremental gradient method with support for
  non-strongly convex composite objectives.
\newblock In Z.~Ghahramani, M.~Welling, C.~Cortes, N.~Lawrence, and K.~Q.
  Weinberger, editors, \emph{Advances in neural information processing
  systems}, volume~27, pages 1646--1654. (Curran Associates, NY, USA), 2014.

\bibitem[Du and Lee(2018)]{du2018power}
Simon~S Du and Jason~D Lee.
\newblock On the power of over-parametrization in neural networks with
  quadratic activation.
\newblock In Jennifer Dy and Andreas Krause, editors, \emph{Proceedings of the
  35th International Conference on Machine Learning}, volume~80, pages
  1329--1338. (PMLR), 2018.

\bibitem[Du et~al.(2020)Du, Kakade, Wang, and Yang]{Du2020Is}
Simon~S. Du, Sham~M. Kakade, Ruosong Wang, and Lin~F. Yang.
\newblock Is a good representation sufficient for sample efficient
  reinforcement learning?
\newblock In \emph{International Conference on Learning Representations}, 2020.
\newblock URL \url{https://openreview.net/forum?id=r1genAVKPB}.

\bibitem[Evans(2005)]{evans2005introduction}
Lawrence~C Evans.
\newblock An introduction to mathematical optimal control theory.
\newblock \emph{Lecture Notes, University of California, Department of
  Mathematics, Berkeley}, 2005.

\bibitem[Fang et~al.(1994)Fang, Loparo, and Feng]{fang1994inequalities}
Yuguang Fang, Kenneth~A Loparo, and Xiangbo Feng.
\newblock Inequalities for the trace of matrix product.
\newblock \emph{IEEE Transactions on Automatic Control}, 39\penalty0
  (12):\penalty0 2489--2490, 1994.

\bibitem[Farahmand et~al.(2010)Farahmand, Szepesv{\'a}ri, and
  Munos]{farahmand2010error}
Amir-massoud Farahmand, Csaba Szepesv{\'a}ri, and R{\'e}mi Munos.
\newblock Error propagation for approximate policy and value iteration.
\newblock In J.~Lafferty, C.~Williams, J.~Shawe-Taylor, R.~Zemel, and
  A.~Culotta, editors, \emph{Advances in Neural Information Processing
  Systems}, volume~23, pages 568--576. (Curran Associates, NY, USA), 2010.

\bibitem[Fazel et~al.(2018)Fazel, Ge, Kakade, and Mesbahi]{fazel2018global}
Maryam Fazel, Rong Ge, Sham Kakade, and Mehran Mesbahi.
\newblock Global convergence of policy gradient methods for the linear
  quadratic regulator.
\newblock In Jennifer Dy and Andreas Krause, editors, \emph{Proceedings of the
  35th International Conference on Machine Learning}, volume~80, pages
  1467--1476. (PMLR), 2018.

\bibitem[Ferns et~al.(2004)Ferns, Panangaden, and Precup]{ferns2004metrics}
Norm Ferns, Prakash Panangaden, and Doina Precup.
\newblock Metrics for finite markov decision processes.
\newblock In C.~Meek, M.~Chickering, and J.~Halpern, editors, \emph{Proceedings
  of the 20th conference on Uncertainty in artificial intelligence}, pages
  162--169. (AUAI Press, Virginia, USA), 2004.

\bibitem[Fu et~al.(2018)Fu, Singh, Ghosh, Yang, and Levine]{fu2018variational}
Justin Fu, Avi Singh, Dibya Ghosh, Larry Yang, and Sergey Levine.
\newblock Variational inverse control with events: A general framework for
  data-driven reward definition.
\newblock In S.~Bengio, H.~Wallach, H.~Larochelle, K.~Grauman, N.~Cesa-Bianchi,
  and R.~Garnett, editors, \emph{Advances in Neural Information Processing
  Systems}, volume~31, pages 8538--8547. (Curran Associates, NY, USA), 2018.

\bibitem[Fu(2006)]{fu2006gradient}
Michael~C Fu.
\newblock Gradient estimation.
\newblock \emph{Handbooks in operations research and management science},
  13:\penalty0 575--616, 2006.

\bibitem[Ge et~al.(2015)Ge, Huang, Jin, and Yuan]{ge2015escaping}
Rong Ge, Furong Huang, Chi Jin, and Yang Yuan.
\newblock Escaping from saddle points—online stochastic gradient for tensor
  decomposition.
\newblock In P.~Grünwald, E.~Hazan, and S.~Kale, editors, \emph{Proceedings of
  The 28th Conference on Learning Theory}, volume~40, pages 797--842. (PMLR),
  2015.

\bibitem[Ge et~al.(2016)Ge, Lee, and Ma]{ge2016matrix}
Rong Ge, Jason~D Lee, and Tengyu Ma.
\newblock Matrix completion has no spurious local minimum.
\newblock In D.~Lee, M.~Sugiyama, U.~Luxburg, I.~Guyon, and R.~Garnett,
  editors, \emph{Advances in Neural Information Processing Systems}, volume~29,
  pages 2973--2981. (Curran Associates, NY, USA), 2016.

\bibitem[Geist et~al.(2017)Geist, Piot, and Pietquin]{geist2017bellman}
Matthieu Geist, Bilal Piot, and Olivier Pietquin.
\newblock Is the bellman residual a bad proxy?
\newblock In I.~Guyon, U.V. Luxburg, S.~Bengio, H.~Wallach, R.~Fergus,
  S.~Vishwanathan, and R.~Garnett, editors, \emph{Advances in Neural
  Information Processing Systems}, volume~30, pages 3208--3217. (Curran
  Associates, NY, USA), 2017.

\bibitem[Geist et~al.(2019)Geist, Scherrer, and Pietquin]{geist2019theory}
Matthieu Geist, Bruno Scherrer, and Olivier Pietquin.
\newblock A theory of regularized markov decision processes.
\newblock In \emph{ICML 2019-Thirty-sixth International Conference on Machine
  Learning}, 2019.

\bibitem[Ghadimi and Lan(2013)]{ghadimi2013stochastic}
Saeed Ghadimi and Guanghui Lan.
\newblock Stochastic first-and zeroth-order methods for nonconvex stochastic
  programming.
\newblock \emph{SIAM Journal on Optimization}, 23\penalty0 (4):\penalty0
  2341--2368, 2013.

\bibitem[Ghadimi and Lan(2016)]{ghadimi2016accelerated}
Saeed Ghadimi and Guanghui Lan.
\newblock Accelerated gradient methods for nonconvex nonlinear and stochastic
  programming.
\newblock \emph{Mathematical Programming}, 156\penalty0 (1):\penalty0 59--99,
  2016.

\bibitem[Glasserman and Ho(1991)]{glasserman1991gradient}
Paul Glasserman and Yu-Chi Ho.
\newblock \emph{Gradient estimation via perturbation analysis}, volume 116.
\newblock (Springer Science \& Business Media, Berlin, Germany), 1991.

\bibitem[Glasserman and Tayur(1995)]{glasserman1995sensitivity}
Paul Glasserman and Sridhar Tayur.
\newblock Sensitivity analysis for base-stock levels in multiechelon
  production-inventory systems.
\newblock \emph{Management Science}, 41\penalty0 (2):\penalty0 263--281, 1995.

\bibitem[Grondman et~al.(2012)Grondman, Busoniu, Lopes, and
  Babuska]{grondman2012survey}
Ivo Grondman, Lucian Busoniu, Gabriel~AD Lopes, and Robert Babuska.
\newblock A survey of actor-critic reinforcement learning: Standard and natural
  policy gradients.
\newblock \emph{IEEE Transactions on Systems, Man, and Cybernetics, Part C
  (Applications and Reviews)}, 42\penalty0 (6):\penalty0 1291--1307, 2012.

\bibitem[Haarnoja et~al.(2018)Haarnoja, Zhou, Hartikainen, Tucker, Ha, Tan,
  Kumar, Zhu, Gupta, Abbeel, et~al.]{haarnoja2018soft}
Tuomas Haarnoja, Aurick Zhou, Kristian Hartikainen, George Tucker, Sehoon Ha,
  Jie Tan, Vikash Kumar, Henry Zhu, Abhishek Gupta, Pieter Abbeel, et~al.
\newblock Soft actor-critic algorithms and applications.
\newblock \emph{arXiv preprint arXiv:1812.05905}, 2018.

\bibitem[Hern{\'a}ndez-Lerma and Lasserre(2012)]{hernandez2012discrete}
On{\'e}simo Hern{\'a}ndez-Lerma and Jean~B Lasserre.
\newblock \emph{Discrete-time Markov control processes: basic optimality
  criteria}, volume~30.
\newblock (Springer Science \& Business Media, Berlin, Germany), 2012.

\bibitem[Hewer(1971)]{hewer1971iterative}
G~Hewer.
\newblock An iterative technique for the computation of the steady state gains
  for the discrete optimal regulator.
\newblock \emph{IEEE Transactions on Automatic Control}, 16\penalty0
  (4):\penalty0 382--384, 1971.

\bibitem[Huh and Rusmevichientong(2013)]{huh2013online}
Woonghee~Tim Huh and Paat Rusmevichientong.
\newblock Online sequential optimization with biased gradients: theory and
  applications to censored demand.
\newblock \emph{INFORMS Journal on Computing}, 26\penalty0 (1):\penalty0
  150--159, 2013.

\bibitem[Jacot et~al.(2018)Jacot, Gabriel, and Hongler]{jacot2018neural}
Arthur Jacot, Franck Gabriel, and Cl{\'e}ment Hongler.
\newblock Neural tangent kernel: Convergence and generalization in neural
  networks.
\newblock In \emph{Advances in neural information processing systems}, pages
  8571--8580, 2018.

\bibitem[Jin et~al.(2017)Jin, Ge, Netrapalli, Kakade, and
  Jordan]{jin2017escape}
Chi Jin, Rong Ge, Praneeth Netrapalli, Sham~M Kakade, and Michael~I Jordan.
\newblock How to escape saddle points efficiently.
\newblock In D.~Precup and Y.~W. Teh, editors, \emph{Proceedings of the 34th
  International Conference on Machine Learning-Volume 70}, pages 1724--1732.
  (JMLR. org), 2017.

\bibitem[Jin et~al.(2020)Jin, Yang, Wang, and Jordan]{jin2020provably}
Chi Jin, Zhuoran Yang, Zhaoran Wang, and Michael~I Jordan.
\newblock Provably efficient reinforcement learning with linear function
  approximation.
\newblock In \emph{Conference on Learning Theory}, pages 2137--2143. PMLR,
  2020.

\bibitem[Kakade and Langford(2002)]{kakade2002approximately}
Sham Kakade and John Langford.
\newblock Approximately optimal approximate reinforcement learning.
\newblock In C.~Sammut and A.~G. Hoffmann, editors, \emph{Proceedings of the
  Nineteenth International Conference on Machine Learning}, volume~2, pages
  267--274. (Morgan Kaufmann Publishers, CA, USA), 2002.

\bibitem[Kakade(2002)]{kakade2002natural}
Sham~M Kakade.
\newblock A natural policy gradient.
\newblock In T.~G. Dietterich, S.~Becker, and Z.~Ghahramani, editors,
  \emph{Advances in neural information processing systems}, volume~14, pages
  1531--1538. (MIT Press, MA, USA), 2002.

\bibitem[Kallus and Uehara(2020{\natexlab{a}})]{Masatoshi-OPPG}
Nathan Kallus and Masatoshi Uehara.
\newblock Statistically efficient off-policy policy gradients.
\newblock In Hal~Daumé III and Aarti Singh, editors, \emph{Proceedings of the
  37th International Conference on Machine Learning}, volume 119 of
  \emph{Proceedings of Machine Learning Research}, pages 5089--5100. PMLR,
  13--18 Jul 2020{\natexlab{a}}.

\bibitem[Kallus and Uehara(2020{\natexlab{b}})]{Masatoshi_doubly-robust}
Nathan Kallus and Masatoshi Uehara.
\newblock Doubly robust off-policy value and gradient estimation for
  deterministic policies.
\newblock In H.~Larochelle, M.~Ranzato, R.~Hadsell, M.~F. Balcan, and H.~Lin,
  editors, \emph{Advances in Neural Information Processing Systems}, volume~33,
  pages 10420--10430. Curran Associates, Inc., 2020{\natexlab{b}}.

\bibitem[Kalman et~al.(1960)]{kalman1960contributions}
Rudolf~Emil Kalman et~al.
\newblock Contributions to the theory of optimal control.
\newblock \emph{Bol. soc. mat. mexicana}, 5\penalty0 (2):\penalty0 102--119,
  1960.

\bibitem[Karimi et~al.(2016)Karimi, Nutini, and Schmidt]{karimi2016linear}
Hamed Karimi, Julie Nutini, and Mark Schmidt.
\newblock Linear convergence of gradient and proximal-gradient methods under
  the polyak-{\l}ojasiewicz condition.
\newblock In \emph{Joint European Conference on Machine Learning and Knowledge
  Discovery in Databases}, pages 795--811. (Springer, Berlin, Germany), 2016.

\bibitem[Kawaguchi(2016)]{kawaguchi2016deep}
Kenji Kawaguchi.
\newblock Deep learning without poor local minima.
\newblock In D.~Lee, M.~Sugiyama, U.~Luxburg, I.~Guyon, and R.~Garnett,
  editors, \emph{Advances in neural information processing systems}, volume~29,
  pages 586--594. (Curran Associates, NY, USA), 2016.

\bibitem[Kleinman(1968)]{kleinman1968}
D.~Kleinman.
\newblock On an iterative technique for riccati equation computations.
\newblock \emph{IEEE Transactions on Automatic Control}, 13\penalty0
  (1):\penalty0 114 -- 115, 1968.

\bibitem[Koenig and Simmons(1993)]{koenig1993complexity}
Sven Koenig and Reid~G Simmons.
\newblock Complexity analysis of real-time reinforcement learning.
\newblock In R.~Fikes and W.~Lehnert, editors, \emph{Proceedings of the
  Eleventh National Conference on Artificial Intelligence}, pages 99--107.
  (AAAI Press, CA, USA), 1993.

\bibitem[Konda and Tsitsiklis(2000)]{konda2000actor}
Vijay~R Konda and John~N Tsitsiklis.
\newblock Actor-critic algorithms.
\newblock In \emph{Advances in neural information processing systems}, pages
  1008--1014, 2000.

\bibitem[Kunnumkal and Topaloglu(2008)]{kunnumkal2008using}
Sumit Kunnumkal and Huseyin Topaloglu.
\newblock Using stochastic approximation methods to compute optimal base-stock
  levels in inventory control problems.
\newblock \emph{Operations Research}, 56\penalty0 (3):\penalty0 646--664, 2008.

\bibitem[L'Ecuyer and Glynn(1994)]{l1994stochastic1}
Pierre L'Ecuyer and Peter~W Glynn.
\newblock Stochastic optimization by simulation: Convergence proofs for the
  gi/g/1 queue in steady-state.
\newblock \emph{Management Science}, 40\penalty0 (11):\penalty0 1562--1578,
  1994.

\bibitem[L'Ecuyer et~al.(1994)L'Ecuyer, Giroux, and Glynn]{l1994stochastic2}
Pierre L'Ecuyer, Nataly Giroux, and Peter~W Glynn.
\newblock Stochastic optimization by simulation: numerical experiments with the
  m/m/1 queue in steady-state.
\newblock \emph{Management science}, 40\penalty0 (10):\penalty0 1245--1261,
  1994.

\bibitem[Lee et~al.(2016)Lee, Simchowitz, Jordan, and Recht]{lee2016gradient}
Jason~D Lee, Max Simchowitz, Michael~I Jordan, and Benjamin Recht.
\newblock Gradient descent only converges to minimizers.
\newblock In V.~Feldman, A.~Rakhlin, and O.~Shamir, editors, \emph{29th Annual
  Conference on Learning Theory}, volume~49, pages 1246--1257. (PMLR), 2016.

\bibitem[Lesort et~al.(2020)Lesort, Lomonaco, Stoian, Maltoni, Filliat, and
  D{\'\i}az-Rodr{\'\i}guez]{lesort2020continual}
Timoth{\'e}e Lesort, Vincenzo Lomonaco, Andrei Stoian, Davide Maltoni, David
  Filliat, and Natalia D{\'\i}az-Rodr{\'\i}guez.
\newblock Continual learning for robotics: Definition, framework, learning
  strategies, opportunities and challenges.
\newblock \emph{Information Fusion}, 58:\penalty0 52--68, 2020.

\bibitem[Livni et~al.(2014)Livni, Shalev-Shwartz, and
  Shamir]{livni2014computational}
Roi Livni, Shai Shalev-Shwartz, and Ohad Shamir.
\newblock On the computational efficiency of training neural networks.
\newblock In Z.~Ghahramani, M.~Welling, C.~Cortes, N.~Lawrence, and K.~Q.
  Weinberger, editors, \emph{Advances in neural information processing
  systems}, volume~27, pages 855--863. (Curran Associates, NY, USA), 2014.

\bibitem[Marbach and Tsitsiklis(2001)]{marbach2001simulation}
Peter Marbach and John~N Tsitsiklis.
\newblock Simulation-based optimization of markov reward processes.
\newblock \emph{IEEE Transactions on Automatic Control}, 46\penalty0
  (2):\penalty0 191--209, 2001.

\bibitem[Maxwell et~al.(2013)Maxwell, Henderson, and
  Topaloglu]{maxwell2013tuning}
Matthew~S Maxwell, Shane~G Henderson, and Huseyin Topaloglu.
\newblock Tuning approximate dynamic programming policies for ambulance
  redeployment via direct search.
\newblock \emph{Stochastic Systems}, 3\penalty0 (2):\penalty0 322--361, 2013.

\bibitem[Mohamed et~al.(2020)Mohamed, Rosca, Figurnov, and
  Mnih]{mohamed2020monte}
Shakir Mohamed, Mihaela Rosca, Michael Figurnov, and Andriy Mnih.
\newblock Monte carlo gradient estimation in machine learning.
\newblock \emph{Journal of Machine Learning Research}, 21\penalty0
  (132):\penalty0 1--62, 2020.

\bibitem[Munos(2003)]{munos2003error}
R{\'e}mi Munos.
\newblock Error bounds for approximate policy iteration.
\newblock In T.~Fawcett and N.~Mishra, editors, \emph{Proceedings of the
  Twentieth International Conference on International Conference on Machine
  Learning}, pages 560--567. (AAAI Press, CA, USA), 2003.

\bibitem[Munos(2005)]{munos2005error}
R{\'e}mi Munos.
\newblock Error bounds for approximate value iteration.
\newblock In \emph{Proceedings of the National Conference on Artificial
  Intelligence}, volume~20, page 1006. Menlo Park, CA; Cambridge, MA; London;
  AAAI Press; MIT Press; 1999, 2005.

\bibitem[Munos(2007)]{munos2007performance}
R{\'e}mi Munos.
\newblock Performance bounds in {L\_p}-norm for approximate value iteration.
\newblock \emph{SIAM Journal on Control and Optimization}, 46\penalty0
  (2):\penalty0 541--561, 2007.

\bibitem[Munos and Szepesv{\'a}ri(2008)]{munos2008finite}
R{\'e}mi Munos and Csaba Szepesv{\'a}ri.
\newblock Finite-time bounds for fitted value iteration.
\newblock \emph{Journal of Machine Learning Research}, 9\penalty0
  (27):\penalty0 815--857, 2008.

\bibitem[Nesterov and Polyak(2006)]{nesterov2006cubic}
Yurii Nesterov and Boris~T Polyak.
\newblock Cubic regularization of newton method and its global performance.
\newblock \emph{Mathematical Programming}, 108\penalty0 (1):\penalty0 177--205,
  2006.

\bibitem[Osband et~al.(2013)Osband, Russo, and Van~Roy]{osband2013more}
Ian Osband, Daniel Russo, and Benjamin Van~Roy.
\newblock ({M}ore) efficient reinforcement learning via posterior sampling.
\newblock In C.~J.~C. Burges, L.~Bottou, M.~Welling, Z.~Ghahramani, and K.~Q.
  Weinberger, editors, \emph{Advances in Neural Information Processing
  Systems}, volume~26, pages 3003--3011. (Curran Associates, NY, USA), 2013.

\bibitem[Osband et~al.(2019)Osband, Van~Roy, Russo, and Wen]{osband2017deep}
Ian Osband, Benjamin Van~Roy, Daniel~J Russo, and Zheng Wen.
\newblock Deep exploration via randomized value functions.
\newblock \emph{Journal of Machine Learning Research}, 20\penalty0
  (124):\penalty0 1--62, 2019.

\bibitem[Peressini et~al.(1988)Peressini, Sullivan, and
  Uhl]{peressini1988mathematics}
Anthony~L Peressini, Francis~E Sullivan, and J~Jerry Uhl.
\newblock \emph{The mathematics of nonlinear programming}.
\newblock (Springer-Verlag, NY, USA), 1988.

\bibitem[Peters and Schaal(2006)]{peters2006policy}
Jan Peters and Stefan Schaal.
\newblock Policy gradient methods for robotics.
\newblock In \emph{2006 IEEE/RSJ International Conference on Intelligent Robots
  and Systems}, pages 2219--2225. (IEEE, NY, USA), 2006.

\bibitem[Pflug(1988)]{pflug1988derivatives}
G.~Ch. Pflug.
\newblock Derivatives of probability measures-concepts and applications to the
  optimization of stochastic systems.
\newblock In P.~Varaiya and A.B. Kurzhanski, editors, \emph{Discrete Event
  Systems: Models and Applications. Lecture Notes in Control and Information
  Sciences}, volume 103, pages 252--274. (Springer, Berlin, Germany), 1988.

\bibitem[Pflug(1990)]{pflug1990line}
G.~Ch. Pflug.
\newblock On-line optimization of simulated markovian processes.
\newblock \emph{Mathematics of Operations Research}, 15\penalty0 (3):\penalty0
  381--395, 1990.

\bibitem[Polyak(1963)]{polyak1963gradient}
Boris~T Polyak.
\newblock Gradient methods for the minimisation of functionals.
\newblock \emph{USSR Computational Mathematics and Mathematical Physics},
  3\penalty0 (4):\penalty0 864--878, 1963.

\bibitem[Puterman(2014)]{puterman2014markov}
Martin~L Puterman.
\newblock \emph{Markov decision processes: discrete stochastic dynamic
  programming}.
\newblock (John Wiley \& Sons, NJ, USA), 2014.

\bibitem[Rajeswaran et~al.(2017)Rajeswaran, Lowrey, Todorov, and
  Kakade]{rajeswaran2017towards}
Aravind Rajeswaran, Kendall Lowrey, Emanuel~V Todorov, and Sham~M Kakade.
\newblock Towards generalization and simplicity in continuous control.
\newblock In I.~Guyon, U.~V. Luxburg, S.~Bengio, H.~Wallach, R.~Fergus,
  S.~Vishwanathan, and R.~Garnett, editors, \emph{Advances in Neural
  Information Processing Systems}, volume~30, pages 6550--6561. (Curran
  Associates, NY, USA), 2017.

\bibitem[Rautert and Sachs(1997)]{rautert1997computational}
Tankred Rautert and Ekkehard~W Sachs.
\newblock Computational design of optimal output feedback controllers.
\newblock \emph{SIAM Journal on Optimization}, 7\penalty0 (3):\penalty0
  837--852, 1997.

\bibitem[Reddi et~al.(2016{\natexlab{a}})Reddi, Hefny, Sra, P{\'o}czos, and
  Smola]{reddi2016stochastic}
Sashank~J Reddi, Ahmed Hefny, Suvrit Sra, Barnab{\'a}s P{\'o}czos, and Alex
  Smola.
\newblock Stochastic variance reduction for nonconvex optimization.
\newblock In M.~F. Balcan and K.~Q. Weinberger, editors, \emph{Proceedings of
  The 33rd International Conference on Machine Learning}, volume~48, pages
  314--323. (PMLR), 2016{\natexlab{a}}.

\bibitem[Reddi et~al.(2016{\natexlab{b}})Reddi, Sra, P{\'o}czos, and
  Smola]{reddi2016stochasticfrank}
Sashank~J Reddi, Suvrit Sra, Barnab{\'a}s P{\'o}czos, and Alex Smola.
\newblock Stochastic frank-wolfe methods for nonconvex optimization.
\newblock In \emph{54th Annual Allerton Conference on Communication, Control,
  and Computing (Allerton)}, pages 1244--1251. (IEEE, NY, USA),
  2016{\natexlab{b}}.

\bibitem[Reddi et~al.(2016{\natexlab{c}})Reddi, Sra, Poczos, and
  Smola]{reddi2016proximal}
Sashank~J Reddi, Suvrit Sra, Barnabas Poczos, and Alexander~J Smola.
\newblock Proximal stochastic methods for nonsmooth nonconvex finite-sum
  optimization.
\newblock In D.~Lee, M.~Sugiyama, U.~Luxburg, I.~Guyon, and R.~Garnett,
  editors, \emph{Advances in Neural Information Processing Systems}, volume~29,
  pages 1145--1153. (Curran Associates, NY, USA), 2016{\natexlab{c}}.

\bibitem[Rhee and Glynn(2017)]{rhee2017lyapunov}
Chang-Han Rhee and Peter Glynn.
\newblock Lyapunov conditions for differentiability of markov chain
  expectations: the absolutely continuous case.
\newblock \emph{arXiv preprint arXiv:1707.03870}, 2017.

\bibitem[Riedmiller et~al.(2007)Riedmiller, Peters, and
  Schaal]{riedmiller2007evaluation}
Martin Riedmiller, Jan Peters, and Stefan Schaal.
\newblock Evaluation of policy gradient methods and variants on the cart-pole
  benchmark.
\newblock In \emph{2007 IEEE International Symposium on Approximate Dynamic
  Programming and Reinforcement Learning}, pages 254--261. (IEEE, NY, USA),
  2007.

\bibitem[Salimans et~al.(2017)Salimans, Ho, Chen, Sidor, and
  Sutskever]{salimans2017evolution}
Tim Salimans, Jonathan Ho, Xi~Chen, Szymon Sidor, and Ilya Sutskever.
\newblock Evolution strategies as a scalable alternative to reinforcement
  learning.
\newblock \emph{arXiv preprint arXiv:1703.03864}, 2017.

\bibitem[Scherrer and Geist(2014)]{scherrer2014local}
Bruno Scherrer and Matthieu Geist.
\newblock Local policy search in a convex space and conservative policy
  iteration as boosted policy search.
\newblock In \emph{Joint European Conference on Machine Learning and Knowledge
  Discovery in Databases}, pages 35--50. (Springer, Berlin, Germany), 2014.

\bibitem[Schulman et~al.(2015)Schulman, Levine, Abbeel, Jordan, and
  Moritz]{schulman2015trust}
John Schulman, Sergey Levine, Pieter Abbeel, Michael Jordan, and Philipp
  Moritz.
\newblock Trust region policy optimization.
\newblock In F.~Bach and D.~Blei, editors, \emph{Proceedings of the 32nd
  International Conference on Machine Learning}, volume~37, pages 1889--1897.
  (PMLR), 2015.

\bibitem[Schulman et~al.(2016)Schulman, Moritz, Levine, Jordan, and
  Abbeel]{schulman2015high}
John Schulman, Philipp Moritz, Sergey Levine, Michael Jordan, and Pieter
  Abbeel.
\newblock High-dimensional continuous control using generalized advantage
  estimation.
\newblock In Y.~Bengio and Y.~LeCun, editors, \emph{Proceedings of the 4th
  International Conference on Learning Representations (ICLR)}, 2016.

\bibitem[Schulman et~al.(2017)Schulman, Wolski, Dhariwal, Radford, and
  Klimov]{schulman2017proximal}
John Schulman, Filip Wolski, Prafulla Dhariwal, Alec Radford, and Oleg Klimov.
\newblock Proximal policy optimization algorithms.
\newblock \emph{arXiv preprint arXiv:1707.06347}, 2017.

\bibitem[Shani et~al.(2020)Shani, Efroni, and Mannor]{shani2020adaptive}
Lior Shani, Yonathan Efroni, and Shie Mannor.
\newblock Adaptive trust region policy optimization: Global convergence and
  faster rates for regularized mdps.
\newblock In F.~Rossi, V.~Conitzer, and F.~Sha, editors, \emph{Thirty-Fourth
  AAAI Conference on Artificial Intelligence}, volume~34. (AAAI Press, CA,
  USA), 2020.

\bibitem[Silver et~al.(2014)Silver, Lever, Heess, Degris, Wierstra, and
  Riedmiller]{silver2014deterministic}
David Silver, Guy Lever, Nicolas Heess, Thomas Degris, Daan Wierstra, and
  Martin Riedmiller.
\newblock Deterministic policy gradient algorithms.
\newblock In E.~P. Xing and T.~Jebara, editors, \emph{Proceedings of the 31st
  International Conference on Machine Learning}, volume~32. (PMLR), 2014.

\bibitem[Singh et~al.(1994)Singh, Jaakkola, and Jordan]{singh1995reinforcement}
Satinder~P Singh, Tommi Jaakkola, and Michael~I Jordan.
\newblock Reinforcement learning with soft state aggregation.
\newblock In G.~Tesauro, D.~Touretzky, and T.~Leen, editors, \emph{Advances in
  neural information processing systems}, volume~7, pages 361--368. (MIT Press,
  MA, USA), 1994.

\bibitem[Strehl and Littman(2008)]{strehl2008analysis}
Alexander~L Strehl and Michael~L Littman.
\newblock An analysis of model-based interval estimation for markov decision
  processes.
\newblock \emph{Journal of Computer and System Sciences}, 74\penalty0
  (8):\penalty0 1309--1331, 2008.

\bibitem[{Sun} et~al.(2017){Sun}, {Qu}, and {Wright}]{7755794}
J.~{Sun}, Q.~{Qu}, and J.~{Wright}.
\newblock Complete dictionary recovery over the sphere {I}: Overview and the
  geometric picture.
\newblock \emph{IEEE Transactions on Information Theory}, 63\penalty0
  (2):\penalty0 853--884, 2017.

\bibitem[Sutton and Barto(2018)]{sutton2018reinforcement}
Richard~S Sutton and Andrew~G Barto.
\newblock \emph{Reinforcement learning: An introduction}.
\newblock (MIT Press, MA, USA), 2018.

\bibitem[Sutton et~al.(1999)Sutton, McAllester, Singh, and
  Mansour]{sutton2000policy}
Richard~S Sutton, David~A McAllester, Satinder~P Singh, and Yishay Mansour.
\newblock Policy gradient methods for reinforcement learning with function
  approximation.
\newblock In S.~Solla, T.~Leen, and K.~M\"{u}ller, editors, \emph{Advances in
  neural information processing systems}, volume~12, pages 1057--1063. (MIT
  Press, MA, USA), 1999.

\bibitem[Talluri and Van~Ryzin(2006)]{talluri2006theory}
Kalyan~T Talluri and Garrett~J Van~Ryzin.
\newblock \emph{The theory and practice of revenue management}, volume~68.
\newblock (Springer Science \& Business Media, Berlin, Germany), 2006.

\bibitem[Thrun(1992)]{thrun1992cient}
Sebastian~B Thrun.
\newblock Efficient exploration in reinforcement learning.
\newblock Technical report, CMU-CS-92-102, School of Computer Science, Carnegie
  Mellon University, 1992.

\bibitem[Toivonen(1985)]{toivonen1985globally}
Hannu~T Toivonen.
\newblock A globally convergent algorithm for the optimal constant output
  feedback problem.
\newblock \emph{International Journal of Control}, 41\penalty0 (6):\penalty0
  1589--1599, 1985.

\bibitem[Tsitsiklis and Van~Roy(1996)]{tsitsiklis1996feature}
John~N Tsitsiklis and Benjamin Van~Roy.
\newblock Feature-based methods for large scale dynamic programming.
\newblock \emph{Machine Learning}, 22\penalty0 (1-3):\penalty0 59--94, 1996.

\bibitem[Uehara et~al.(2021)Uehara, Imaizumi, Jiang, Kallus, Sun, and
  Xie]{uehara2021finite}
Masatoshi Uehara, Masaaki Imaizumi, Nan Jiang, Nathan Kallus, Wen Sun, and
  Tengyang Xie.
\newblock Finite sample analysis of minimax offline reinforcement learning:
  Completeness, fast rates and first-order efficiency.
\newblock \emph{arXiv preprint arXiv:2102.02981}, 2021.

\bibitem[Van~Roy(2006)]{van2006performance}
Benjamin Van~Roy.
\newblock Performance loss bounds for approximate value iteration with state
  aggregation.
\newblock \emph{Mathematics of Operations Research}, 31\penalty0 (2):\penalty0
  234--244, 2006.

\bibitem[Wang et~al.(2019)Wang, Cai, Yang, and Wang]{wang2019neural}
Lingxiao Wang, Qi~Cai, Zhuoran Yang, and Zhaoran Wang.
\newblock Neural policy gradient methods: Global optimality and rates of
  convergence.
\newblock In Y.~Bengio and Y.~LeCun, editors, \emph{Proceedings of 7th
  International Conference on Learning Representations (ICLR)}, 2019.

\bibitem[Wang et~al.(2021)Wang, Foster, and Kakade]{wang2021what}
Ruosong Wang, Dean Foster, and Sham~M. Kakade.
\newblock What are the statistical limits of offline {RL} with linear function
  approximation?
\newblock In \emph{International Conference on Learning Representations}, 2021.
\newblock URL \url{https://openreview.net/forum?id=30EvkP2aQLD}.

\bibitem[Weisz et~al.(2021)Weisz, Amortila, and Szepesv\'ari]{Weisz21ALT}
Gell\'ert Weisz, Philip Amortila, and Csaba Szepesv\'ari.
\newblock Exponential lower bounds for planning in mdps with
  linearly-realizable optimal action-value functions.
\newblock In \emph{ALT}, volume 132 of \emph{Proceedings of Machine Learning
  Research}, pages 1237--1264. PMLR, March 2021.

\bibitem[Xiao and Zhang(2014)]{xiao2014proximal}
Lin Xiao and Tong Zhang.
\newblock A proximal stochastic gradient method with progressive variance
  reduction.
\newblock \emph{SIAM Journal on Optimization}, 24\penalty0 (4):\penalty0
  2057--2075, 2014.

\bibitem[Zhu et~al.(2020)Zhu, Yu, Gupta, Shah, Hartikainen, Singh, Kumar, and
  Levine]{zhu2020ingredients}
Henry Zhu, Justin Yu, Abhishek Gupta, Dhruv Shah, Kristian Hartikainen, Avi
  Singh, Vikash Kumar, and Sergey Levine.
\newblock The ingredients of real-world robotic reinforcement learning.
\newblock In Y.~Bengio and Y.~LeCun, editors, \emph{Proceedings of 8th
  International Conference on Learning Representations (ICLR)}, 2020.

\end{thebibliography}
